\documentclass[sigconf,nonacm]{acmart}
\AtBeginDocument{%
  }

\newcommand{\clipcount}{CLIP-Count}

\usepackage{amsmath} 
\usepackage{graphicx}
\usepackage{subcaption}
\usepackage{makecell}
\usepackage{xcolor}
\usepackage{adjustbox}
\begin{document}

\title{Detection-Driven Object Count Optimization for Text-to-image Diffusion Models}

\author{Oz Zafar}
\affiliation{%
  \institution{Tel-Aviv University}
  \city{Tel Aviv}
  \country{Israel}
}
\email{ozzafar@gmail.com}

\author{Yuval Cohen}
\affiliation{%
  \institution{Bar-Ilan University}
  \city{Ramat Gan}
  \country{Israel}
}
\email{cohenyu@biu.ac.il}

\author{Lior Wolf}
\affiliation{%
  \institution{Tel-Aviv University}
  \city{Tel Aviv}
  \country{Israel}
}
\email{wolf@cs.tau.ac.il}

\author{Idan Schwartz}
\affiliation{%
  \institution{Bar-Ilan University}
  \city{Ramat Gan}
  \country{Israel}
}
\email{idanschwartz@gmail.com}

\renewcommand{\shortauthors}{Trovato et al.}

\begin{abstract}
Accurately controlling object count in text-to-image generation remains a key challenge. Supervised methods often fail, as training data rarely covers all count variations. Methods that manipulate the denoising process to add or remove objects can help; however, they still require labeled data, limit robustness and image quality, and rely on a slow, iterative process. 

Pre-trained differentiable counting models that rely on soft object density summation exist and could steer generation, but employing them presents three main challenges: (i) they are pre-trained on clean images, making them less effective during denoising steps that operate on noisy inputs; (ii) they are not robust to viewpoint changes; and (iii) optimization is computationally expensive, requiring repeated model evaluations per image. 

We propose a new framework that uses pre-trained object counting techniques and object detectors to guide generation. First, we optimize a counting token using an outer-loop loss computed on fully generated images. Second, we introduce a detection-driven scaling term that corrects errors caused by viewpoint and proportion shifts, etc., without requiring backpropagation through the detection model. Third, we show that the optimized parameters can be reused for new prompts, removing the need for repeated optimization. Our method provides efficiency through token reuse, flexibility via compatibility with various detectors, and accuracy with improved counting across diverse object categories.

\end{abstract}

\begin{teaserfigure}
    \centering
    \includegraphics[width=0.9\linewidth]{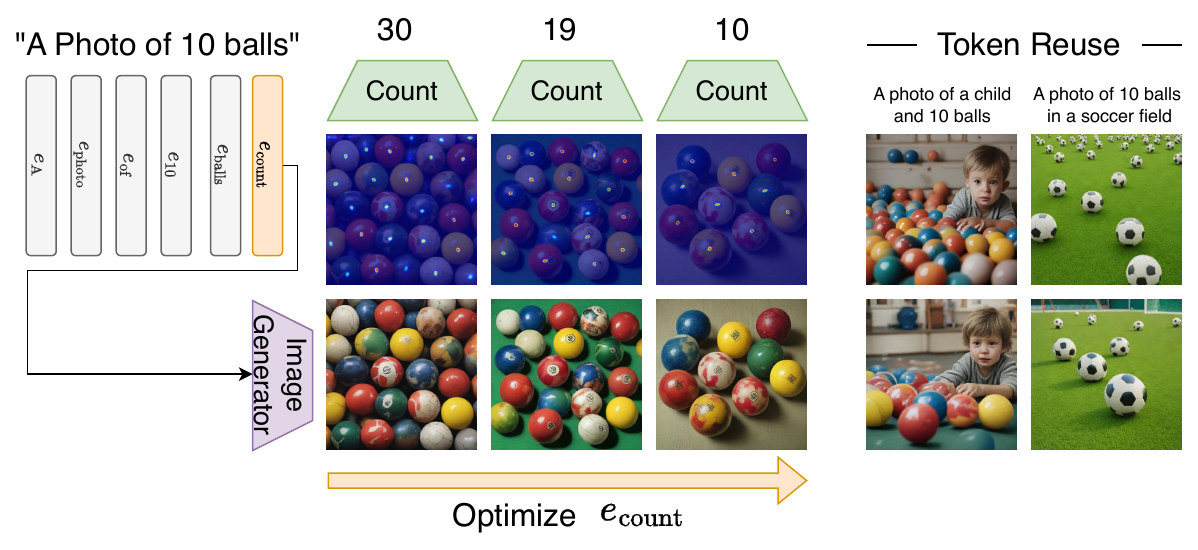}
    \caption{Our method optimizes the embeddings of a newly added token, namely the counting token. The optimization is done iteratively, where we softly calculate the number of objects from the desired class at each step using a density map and derive a loss function. Our method uses detection-driven scaling to improve the counting approximation for image geometry variations. Once converged, the optimized token can be reused in different prompts to correct the number of depicted objects, as shown on the right.}
    \Description{Overview of our optimization technique and results}
    \label{fig:teas}
\end{teaserfigure}

\maketitle

\section{Introduction}

Image generation models have advanced in generating high-quality, realistic images~\citep{rombach2022high, ramesh2022hierarchical, podell2023sdxl}. The introduction of conditioning mechanisms has made them a standard tool for content generation, 3D modeling~\citep{liu2023zero1to3}, improving classifiers~\citep{samuel2024generating}, and more. However, depicting a specific number of objects in the image with text conditioning often fails to capture the exact quantity. For example, the model usually depicts a significantly larger quantity when asked to generate an image of a specific number, such as ``ten balls'' (see Fig.~\ref{fig:teas}).

The challenge of accurate counting extends beyond text-to-image diffusion models. It has also been explored in visual question answering, where ``how many'' questions proved difficult for end-to-end training and required dedicated mechanisms~\citep{zhang2018learning}. Similarly, achieving satisfactory results in text-image matching often requires fine-tuning on specialized counting datasets~\citep{paiss2023teaching}. These efforts parallel how humans acquire counting skills, which are believed to rely on dedicated cognitive modules~\citep{starr2013number}. Inspired by this learning process, counting methods often estimate object quantities using a dedicated module that directly guides the image generation process.

Previous methods have suggested dedicated solutions for image generation with diffusion models. One approach involves training models to add or remove objects~\citep{binyamin2024make}, which requires curated supervision that limits robustness to large quantities, can harm image naturalness, and makes the counting correction a long, iterative process. Another approach proposes using density maps to approximate counting in a differentiable manner, while steering the denoising process by decoding intermediate steps, calculating a density-based loss, and updating the intermediate latent representation with gradient-based optimization. However, this approach also faces three issues: (i) the pre-trained counting density maps are trained on clean images and are often inaccurate when applied to the noisy intermediate outputs of diffusion models; (ii) calculating the density map during denoising for each image is a slow process; and (iii) the density map provides a soft, differentiable representation of object presence, but it is often sensitive to the viewpoint, scale, perspective, object arrangement of the generated image (see Fig.~\ref{fig:potential}), which can lead to systematic overestimation or underestimation of object count.

Here, we propose a novel technique that introduces several changes to how density maps are used. To leverage pre-trained density models effectively, we apply our loss on fully generated images (see Fig.~\ref{fig:teas}, left). To improve efficiency, we introduce a new \textit{counting token}, whose embedding is optimized after the image is generated, instead of optimizing the latent. After each optimization step, the image is regenerated using the updated token embedding. Once convergence is reached, the token can be reused across prompts, including those with different backgrounds and object classes, without additional optimization (see Fig.~\ref{fig:teas}, right), thereby avoiding expensive inference for each image. Our method performs backpropagation only at the final denoising step to further reduce the computational cost of full diffusion backpropagation. Additionally, we leverage distilled diffusion models, which require only a few steps to regenerate images quickly~\citep{podell2023sdxl}.

To address inaccuracies in the density maps caused by variations in scene geometry, we adaptively scale the maps using signals from object detection models during each optimization step. While detection-based methods offer strong cues for object localization and counting, they rely on non-differentiable thresholding operations, which makes them unsuitable for direct gradient-based optimization. To work around this, we do not backpropagate through the detection network. Instead, we use its output to guide the scaling of the density maps. This lightweight integration improves count accuracy while maintaining the efficiency of our pipeline, effectively bridging the gap between smooth, differentiable estimators and robust, detection-based techniques.

In summary, we make the following contributions:
\begin{itemize}
    \item We introduce an inference-time optimization scheme that operates on fully generated images by refining a \textit{counting token} embedding, leveraging a pre-trained model that computes a density map. This token can be reused to generate images for different prompts and classes.

    \item We propose a detection-driven loss that scales the density map with non-differentiable detection signals. It is plug-and-play and allows further improvements with better detectors.

    \item Our method significantly improves object count accuracy with minimal changes to the scene, even for larger quantities where previous baselines fail.
    
\end{itemize}

\section{Related Work}

The field of image generation has been studied extensively, both with GANs~\citep{goodfellow2014generative, crowson2022vqgan} and, more recently, with diffusion models~\citep{ho2020denoising}. The use of diffusion models has, in particular, enabled an unprecedented capability in generating high-quality, diverse images from natural language input with models such as DALL$\cdot$E 2~\citep{ramesh2022hierarchical}, Imagen~\citep{saharia2022photorealistic},  Parti~\citep{yu2022scaling}, Stable Diffusion (SD)~\citep{rombach_2022_cvpr, podell2023sdxl}, and CogView2~\citep{ding2022cogview2}. 

Our method is built on top of text-conditioned image generation. Below, we briefly review the related areas of controlling image generation models.  %

\paragraph{Personalized image generation.} Personalization in image generation typically relies on a small set of images representing a specific concept. Textual Inversion (TI)~\citep{gal2022textual} learns a new token to represent the concept, requiring only 3–5 images to capture its identity. DreamBooth~\citep{ruiz2022dreambooth} extends this by fine-tuning the diffusion model’s UNet parameters. However, such methods are ill-suited for counting tasks, as collecting training data for arbitrary object quantities is impractical. Moreover, it remains unclear whether these techniques can learn abstract compositional properties such as object count, rather than memorizing specific visual entities.

\paragraph{Class-conditioned image generation.} Classifier guidance conditions the diffusion process by injecting gradients from a classifier into the denoising steps~\citep{dhariwal2021diffusion}. This approach requires a classifier trained specifically on intermediate diffusion representations. Classifier-free guidance conditions the model directly on class labels without requiring an external classifier~\citep{ho2022classifier, rombach_2022_cvpr}, but typically relies on small, curated datasets, which can limit generation quality. Subsequent work has proposed leveraging pre-trained classifiers applied to fully generated images~\citep{schwartz2023discriminative}. Our work extends this line of research by introducing a detection-driven counting loss from an external model, demonstrating that classifier-based supervision can be generalized to other forms of external guidance, such as object counting.

\paragraph{Controlled image generation.} ControlNet~\citep{zhang2023adding} uses local controls, such as edge maps, depth maps, or segmentation masks, for precise diffusion control. GLIGEN~\citep{li2023gligen} is trained for open-world grounded text-to-image generation. Uni-Control~\citep{zhao2023uni} allows using local and global controls, such as CLIP image embeddings. However, while local controls can add a specific number of objects, for instance, from segmentation maps, they often generate non-realistic images as the conditioning limits the natural placement of objects.

\paragraph{Optimizing prompt correspondence.} Several works have focused on improving the alignment between text prompts and generated images. Prompt-to-Prompt~\citep{hertz2022prompt} modifies the diffusion denoising process by leveraging cross-attention maps for word-level control, enabling edits such as emphasizing specific terms or replacing objects. Null-text Inversion~\citep{mokady2023null} extends this capability to real images by inverting them into the latent space of the diffusion model. Attend-and-Excite~\citep{chefer2023attend} promotes the depiction of all mentioned objects by optimizing attention uniformity, while Init-noise~\citep{guo2024initno} takes a complementary approach by directly optimizing the initial noise. However, these methods do not address numerical quantities, as number-related words are often not faithfully represented in diffusion models, and simply adjusting their influence is insufficient for accurate control.

\paragraph{Score alignment.} Recent methods such as DAS~\cite{kim2021diffusionclip} and DDPO~\cite{black2024ddpo} align diffusion model outputs with non-differentiable score functions using reinforcement learning. These approaches rely on sample-based updates, estimating gradients through techniques like policy gradients. In contrast, our method employs direct gradient-based optimization with a differentiable, detection-driven counting objective, making it more effective and efficient for this task.

\paragraph{Object count optimization.}  
Recent work has begun addressing object counting in text-to-image generation.
AFreeCA~\citep{dalessandro2024afreeca} uses diffusion models to generate supervised counting data via unsupervised object sorting and density-based labeling, potentially enabling scalable data creation.
CountGen~\citep{binyamin2024make} learns to incrementally add or remove objects using $n$/$n{+}1$ training pairs, but it is limited to 10 objects, requires retraining for different diffusion models or higher counts, and suffers from high latency due to its iterative generation process.
A related method, Counting Guidance~\citep{kang2025counting}, guides denoising by optimizing the latent space using a density map, which imposes no strict limit on object count. The density map is derived from a pre-trained model fine-tuned to match image patches with their corresponding classes by leveraging CLIP~\citep{jiang2023clip, radford2021learning}.
However, applying it during denoising requires decoding an image at each step via a look-ahead mechanism.
This strategy introduces inference-time overhead, and the look-ahead decoding makes the pre-trained density map less reliable than when operating on fully generated images.
In contrast to these two methods, our approach applies the loss on fully generated images and incorporates detection-driven scaling, further improving accuracy. Moreover, the optimized counting token can be reused across prompts, eliminating the need for per-image optimization.

\section{Method}

\begin{figure}[t!]
    \centering
    \includegraphics[width=1\linewidth]{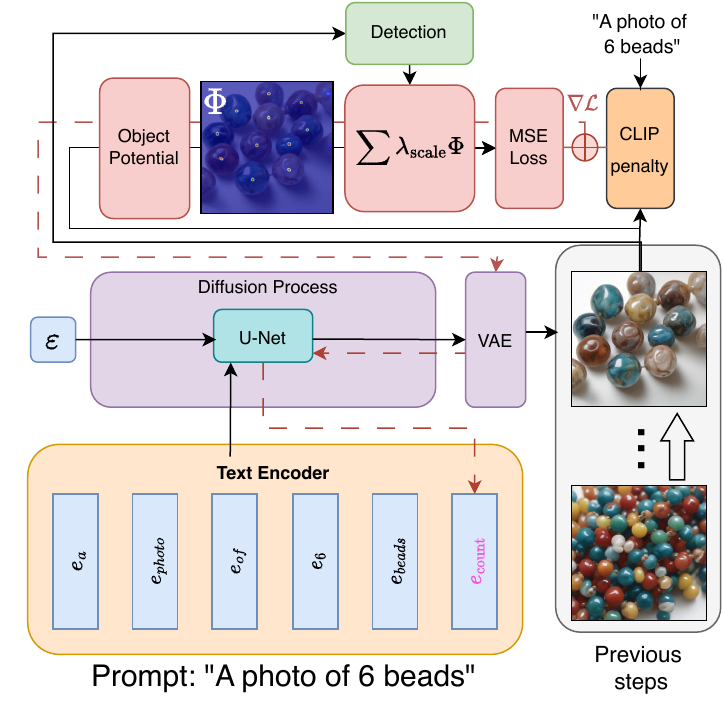}
    \caption{Overview of our method. Given the prompt ``A photo of 6 beads,'' initial generations depict too many beads (bottom left). To improve counting accuracy, we iteratively optimize a dedicated counting token embedding, $e_{\text{count}}$. The loss is computed using a density map ($\Phi$), scaled by a detection-driven factor (shown in green). Since detection is non-differentiable, it is used only to compute a fixed scaling term. Optimization continues until the generated image aligns with the desired count (bottom right).}
    \label{fig:method_diagram}
    \Description{A schematic diagram showing image generation through a diffusion process. A series of images illustrates a gradual reduction in the number of beads, starting with many and ending with six. The scheme shows how a counting token is optimized by propagating gradients through a density map, while detection results are used only for scaling and are not backpropagated.}
\end{figure}

We now outline our strategy for steering the text-to-image model. Intuitively, our method uses a simple prompt containing the object class and desired quantity.  We perform iterative optimization steps on fully generated images, updating the counting token embedding at each step until the desired object count is achieved (see Sec.~\ref{sec:optim}). We define a detection-driven loss based on a density map (see Sec.~\ref{sec:loss}). The final counting token embedding can be reused to generate novel images without further optimization (see Sec.~\ref{sec:method_reuse}). Before detailing the full method, we introduce the relevant preliminaries.

\subsection{Preliminaries: Text-Conditioned Diffusion Process}

Diffusion models generate images by progressively denoising a noise sample to recover a clean image \( x_0 \sim p(x) \). During training, a noisy version \( x_t \) is formed as:
\[
x_t = \sqrt{\alpha_t}x_0 + \sqrt{1 - \alpha_t} \cdot \epsilon_t, \quad \epsilon_t \sim \mathcal{N}(0, \text{I}),
\]
and the model learns to predict either the noise \( \epsilon_t \), or equivalently, the original image \( x_0 \). In the conditional setting, denoising is guided by a conditioning variable \( y \), such as a text prompt:
\[
\mathbb{E}_{x, \epsilon_t, t}[||\hat{\epsilon}_\theta(x_t, t, y) - \epsilon_t||^2_2].
\]

Conditioning image generation on object count typically requires large-scale datasets. An alternative, inspired by classifier guidance~\cite{dhariwal2021diffusion}, is to steer denoising using gradients from a pre-trained counting model. However, this approach requires a model that performs reliably on noisy intermediate representations. In contrast, our method operates on fully generated images, applying loss and updating the text embedding after each generation. We describe the iterative procedure and notation in the next section.

\begin{figure}[t!]
    \centering
    \includegraphics[width=0.80\linewidth]{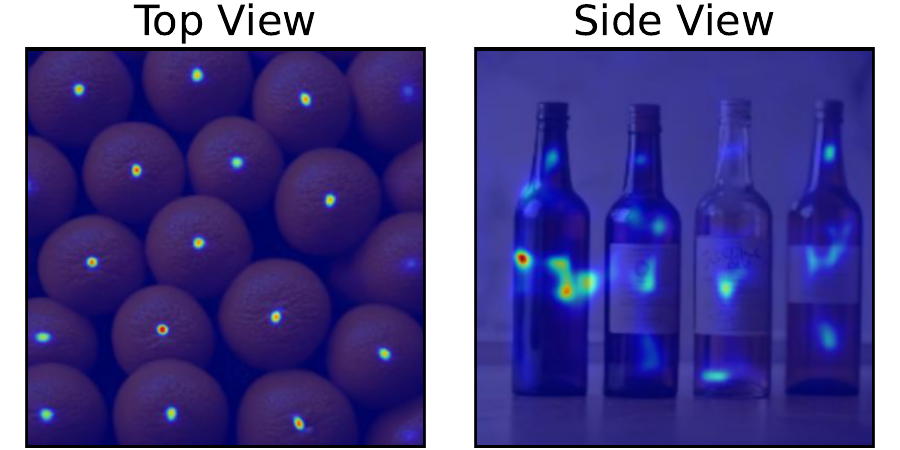}
    \caption{density maps mark objects, but viewpoint changes can distort them and reduce accuracy. We address this by dynamically scaling the map per generated image using a detection model.}
    \Description{density map issues}
    \label{fig:potential}
\end{figure}
\subsection{Inference-Time Optimization for Object Count}
\label{sec:optim}

Our goal is to optimize the number of objects in a generated image. The process begins with an initial prompt of the form ``A photo of $N$ $c$,'' where $N$ is the desired object count and $c$ is the object class. We use a relatively simple prompt during training, as it improves convergence. However, the optimized token can be reused for more complex prompt structures and object categories during inference (see Sec.~\ref{sec:method_reuse}).  

To influence generation, we concatenate a newly introduced pseudo-token embedding, denoted by ${e}_{\text{count}}$, to the text embedding. The counting token is randomly initialized and updated throughout the optimization process. While modifying existing tokens yields similar performance, it may corrupt semantically meaningful tokens with multiple roles; therefore, we prefer optimizing a new token.

At each iteration, we generate an image, compute the loss, update the embedding, and repeat the process. Formally, let $x^{(k)}$ denote the image generated at the $k$-th iteration. We update the counting token embedding as:
\begin{equation}
{e}_{\text{count}}^{(k+1)} \leftarrow {e}_{\text{count}}^{(k)} + \alpha \frac{\nabla_{{e}_{\text{count}}^{(k)}} \mathcal{L}(x^{(k)}, c, N)}{\|\mathcal{L}(x^{(k)}, c, N)\|^2},
\end{equation}
where $\mathcal{L}(x^{(k)}, c, N)$ is the loss measuring the deviation from the target count $N$ for object class $c$ in the generated image $x^{(k)}$, and $\alpha$ is the learning rate. After each update, a new image is generated using the updated token ${e}_{\text{count}}^{(k+1)}$. 

Computing gradients across many diffusion steps can lead to memory exhaustion. Notably, the recent introduction of distilled diffusion models~\cite{sauer2023adversarial} makes optimization with fully generated images significantly more practical, which we leverage in this work.

This process continues until $\mathcal{L}(x^{(k)}, c, N)$ falls below a predefined threshold. We define the loss in the following section.

\subsubsection{Counting Loss}
\label{sec:loss}

At generation step $k$, our detection loss for object class $c$ is defined as the distance from the desired count $N$:
\begin{equation}
\mathcal{L}_{\text{counting}}(x^{(k)}, c, N) = \left\| \operatorname{Count}(x^{(k)}, c) - N \right\|.
\end{equation}

The key challenge lies in defining an accurate and differentiable $\operatorname{Count}(x, c)$. We introduce two counting methods: a static variant that remains fixed throughout optimization, and a dynamic, detection-driven variant that adapts the scaling at each iteration.

\paragraph{Static Counting.}  
We refer to this loss as static, as it does not change between generation steps.  We begin by computing a density map over the image, $\Phi(x, c) \in \mathbb{R}^{h \times w \times 1}$, where $h$ and $w$ are the image dimensions. This density map is derived from a pre-trained model~\citep{jiang2023clip}. The object count is then estimated by aggregating the density map:
\begin{equation}
\operatorname{Count}(x^{(k)}, c) = \sum_{i=1}^{h \cdot w} \lambda_{\text{scale}} \cdot \Phi_i(x^{(k)}, c),
\end{equation}
where $\lambda_{\text{scale}}$ is a scaling hyperparameter used to normalize the output. Determining a general scaling factor is nontrivial, particularly for complex objects with irregular or deformed density maps (see Fig.~\ref{fig:potential}). Next, we introduce a dynamic detection-driven scaling approach to address this limitation.

\begin{figure*}[t!]
  \centering
  \setlength{\tabcolsep}{0pt}
  \resizebox{0.95\textwidth}{!}{%
    \begin{tabular}{c@{\hspace{5pt}}ccccccccc}
      & \textbf{3 birds} & \textbf{5 bowls} & \textbf{5 chairs} & \textbf{8 cups}&\textbf{12 cars} & \textbf{25 seashells} & \textbf{25 grapes} & \textbf{25 pigeons} \\
        \rotatebox{90}{\textbf{SD}}
        & \includegraphics[width=0.16\linewidth]{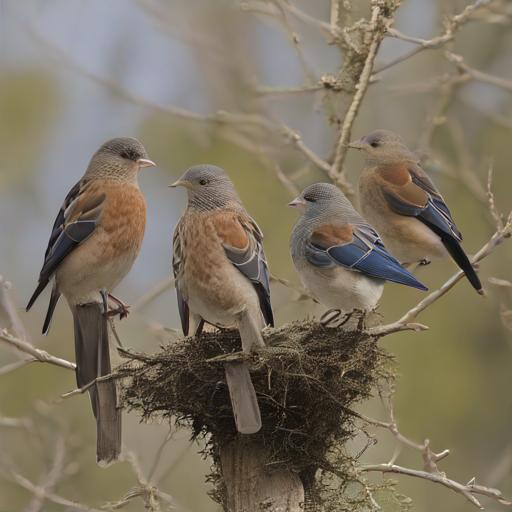}
        & \includegraphics[width=0.16\linewidth]{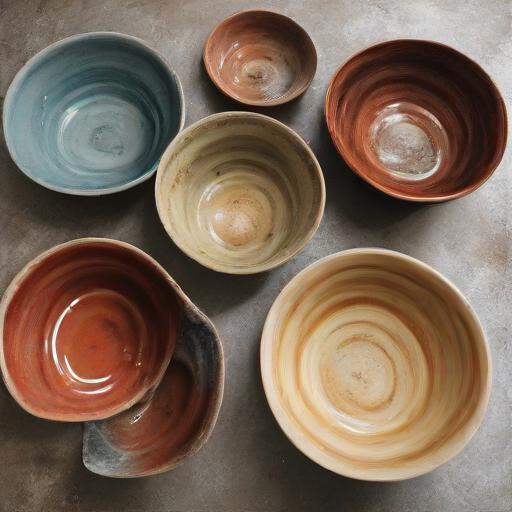}
        & \includegraphics[width=0.16\linewidth]{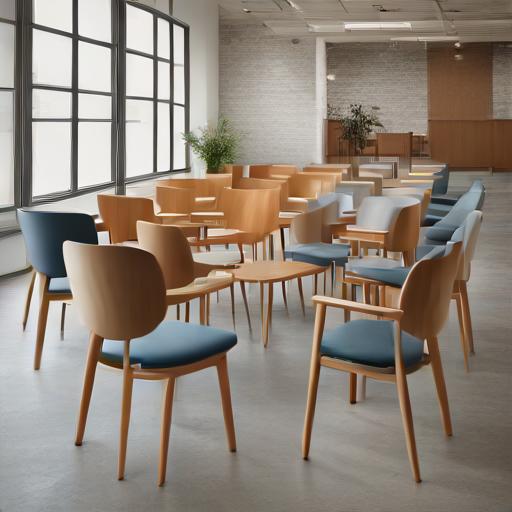}
        & \includegraphics[width=0.16\linewidth]{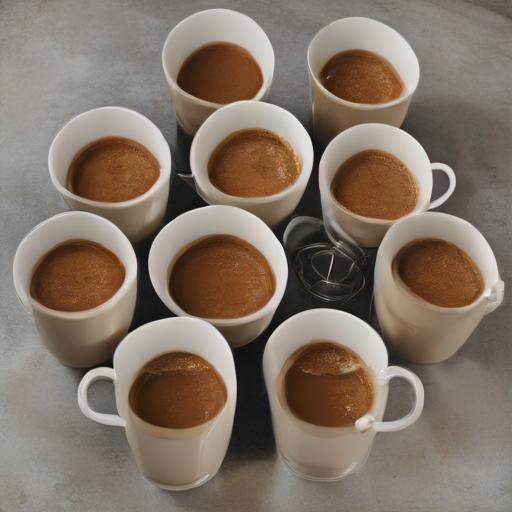}
        & \includegraphics[width=0.16\linewidth]{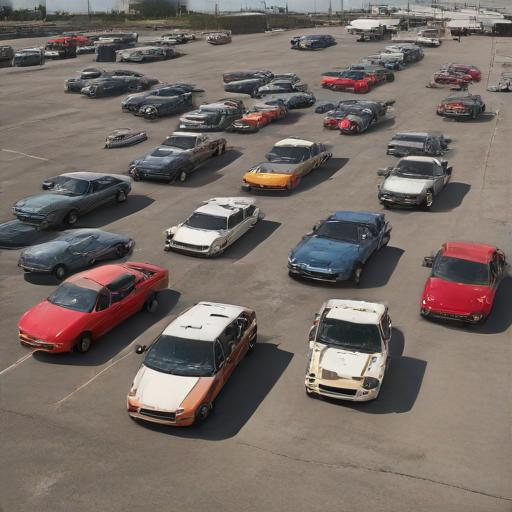}
        & \includegraphics[width=0.16\linewidth]{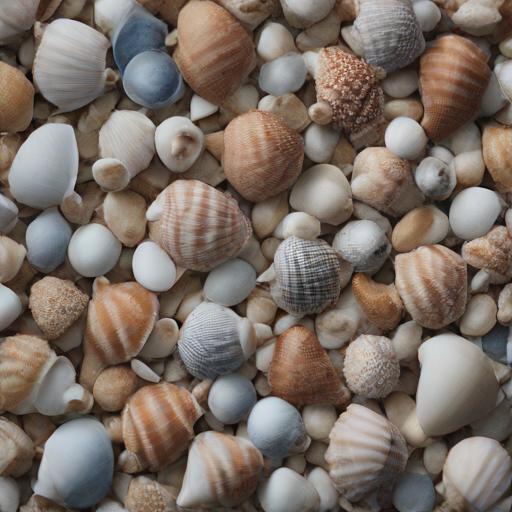}
        & \includegraphics[width=0.16\linewidth]{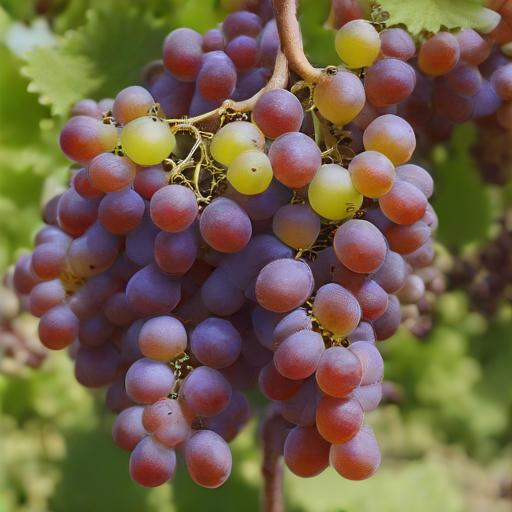}
        & \includegraphics[width=0.16\linewidth]{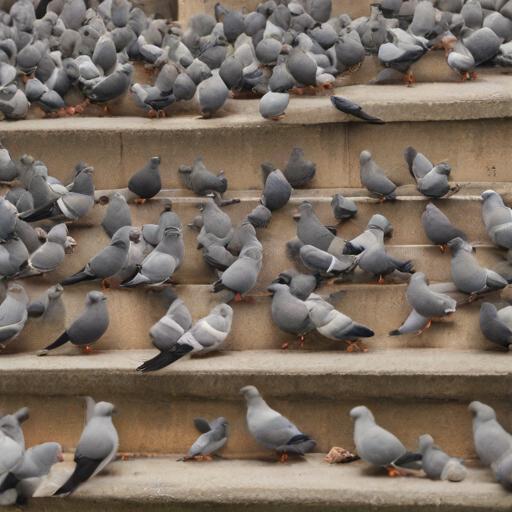} \\
      \rotatebox{90}{\textbf{Ours}} &
      \includegraphics[width=0.16\linewidth]{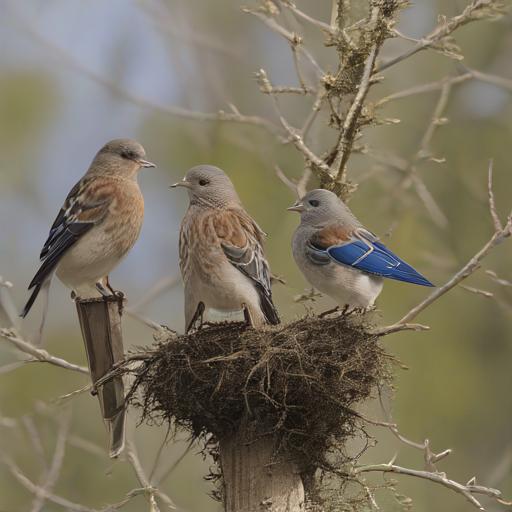}
      & \includegraphics[width=0.16\linewidth]{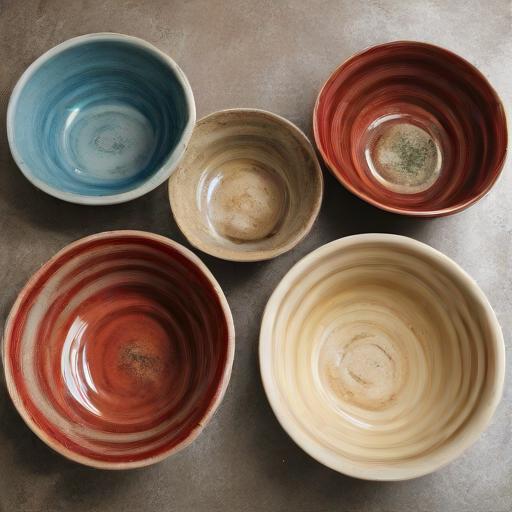}
      & \includegraphics[width=0.16\linewidth]{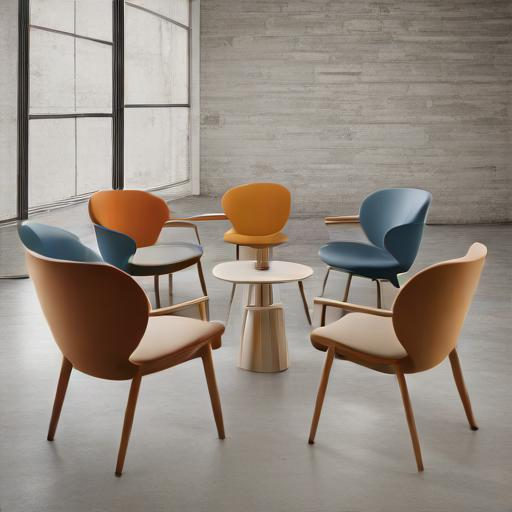}
      & \includegraphics[width=0.16\linewidth]{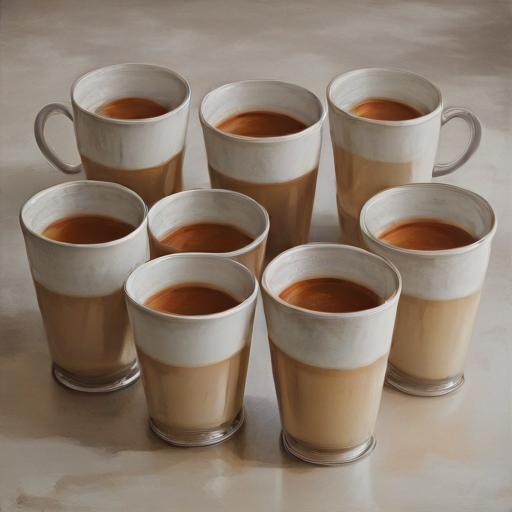}
      & \includegraphics[width=0.16\linewidth]{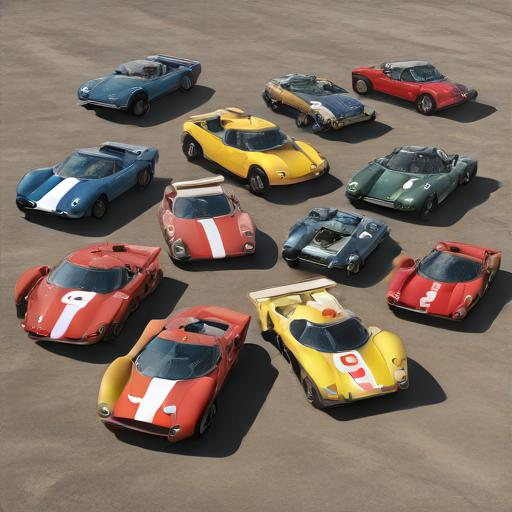}
      & \includegraphics[width=0.16\linewidth]{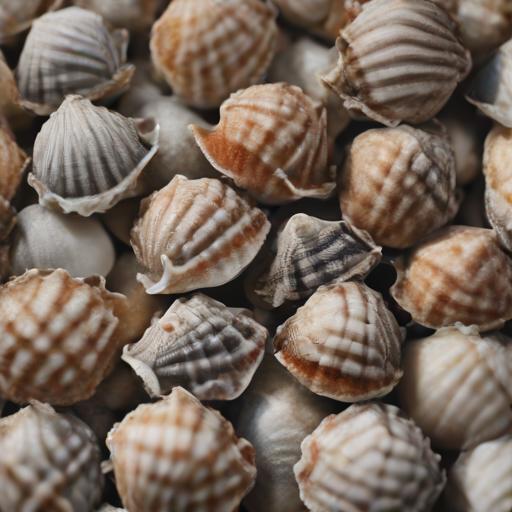}
      & \includegraphics[width=0.16\linewidth]{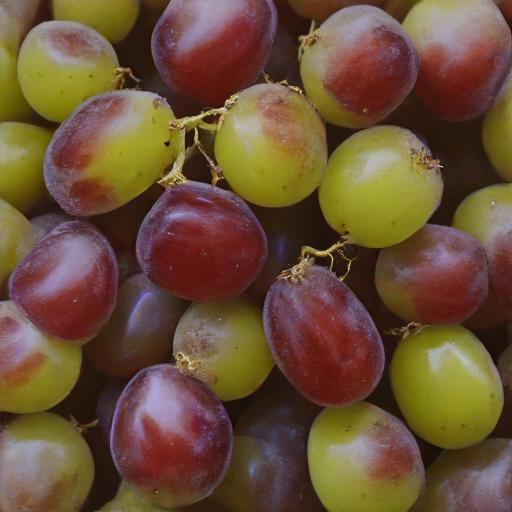}
      & \includegraphics[width=0.16\linewidth]{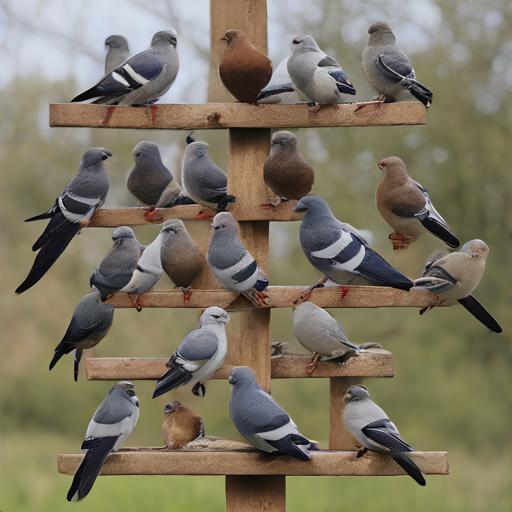} \\
    \end{tabular}}
  \caption{Illustration of images generated using SD-Turbo and our method with DETR-based dynamic scaling. We show various classes and quantities.}
  \Description{Comparison of images generated by SD-Turbo and the proposed method across different object classes and quantities. For each class, the first row shows results from SD-Turbo and the second row from our method. Categories include birds, bowls, chairs, cups, cars, seashells, grapes, and pigeons. Visual differences in object count accuracy and layout are evident.}
  \label{fig:results_basic}
\end{figure*}

\begin{figure}[t!]
  \centering
  \setlength{\tabcolsep}{0pt}
  \resizebox{\linewidth}{!}{%
    \begin{tabular}{c@{\hspace{5pt}}ccccccc}
          Steps: & \textbf{0} & \textbf{1} & \textbf{2} & \textbf{5} & \textbf{16}  \\
         & \includegraphics[width=0.2\linewidth]{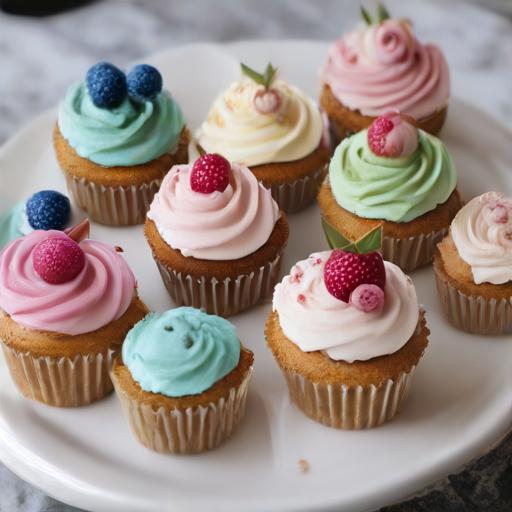}
        & \includegraphics[width=0.2\linewidth]{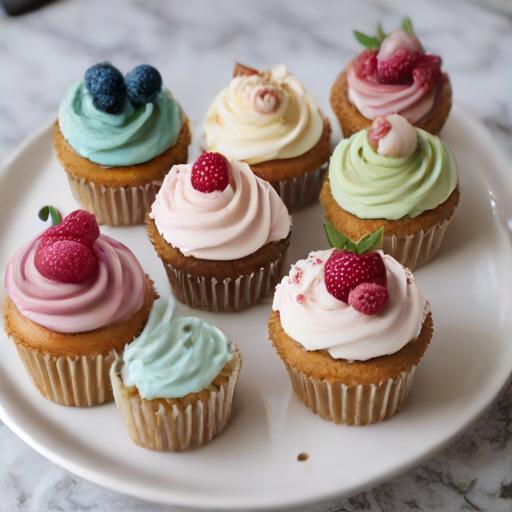}
        & \includegraphics[width=0.2\linewidth]{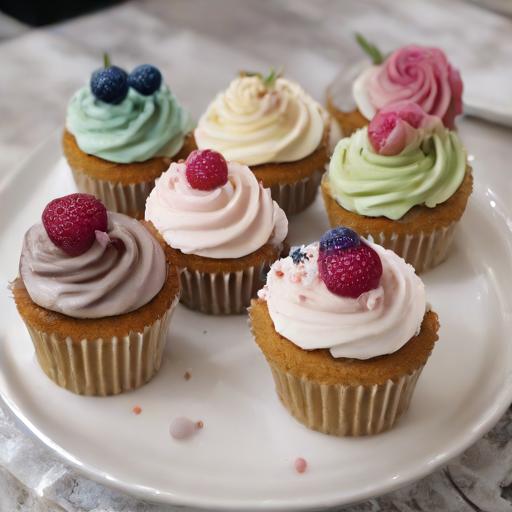}
        & \includegraphics[width=0.2\linewidth]{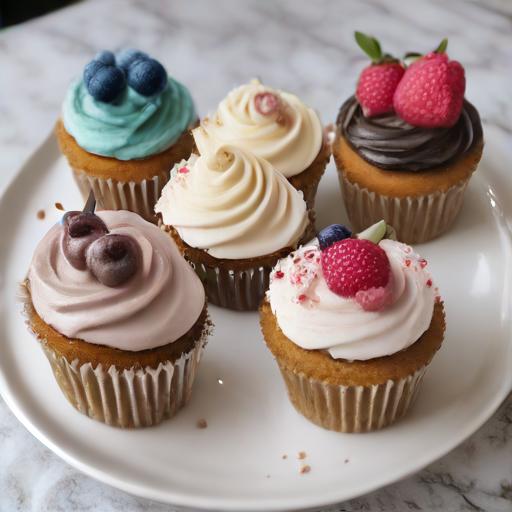}
        & \includegraphics[width=0.2\linewidth]{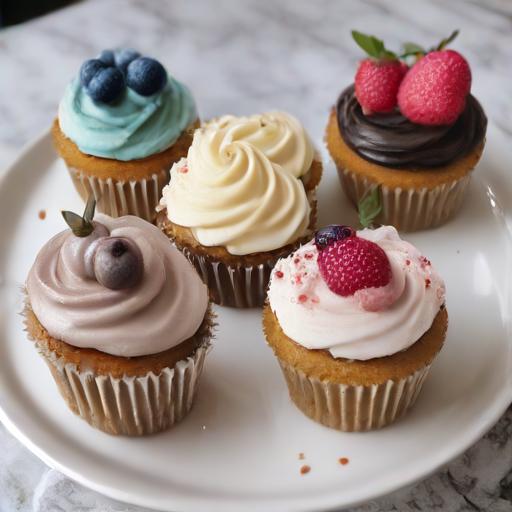} \\
    \rotatebox{90}{5 cupcakes} & \includegraphics[width=0.2\linewidth]{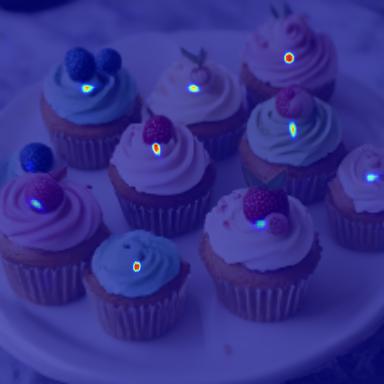}
    & \includegraphics[width=0.2\linewidth]{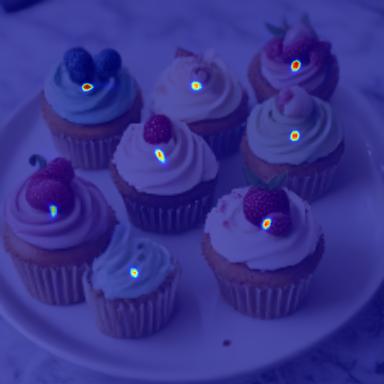}
    & \includegraphics[width=0.2\linewidth]{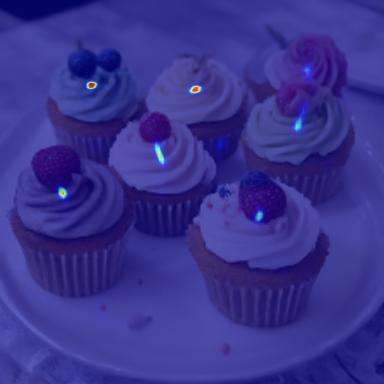}
    & \includegraphics[width=0.2\linewidth]{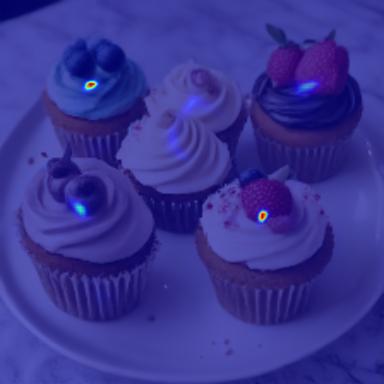}
    & \includegraphics[width=0.2\linewidth]{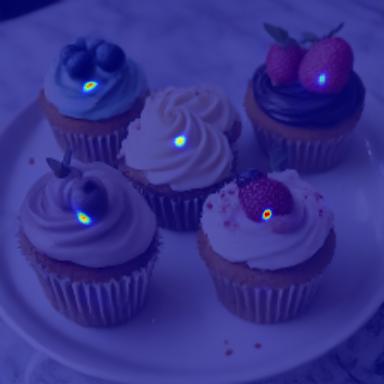} \\
    \\
    & \includegraphics[width=0.2\linewidth]{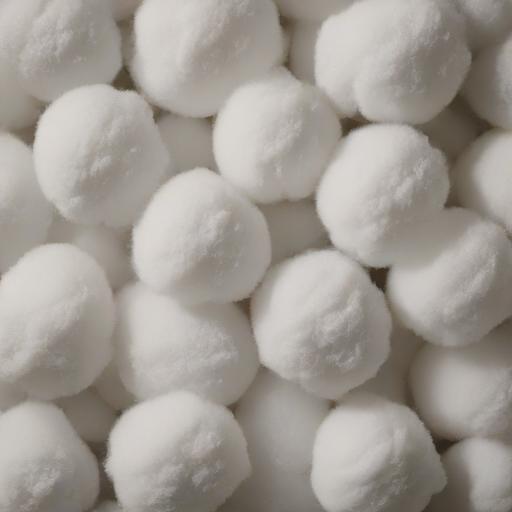}
    & \includegraphics[width=0.2\linewidth]{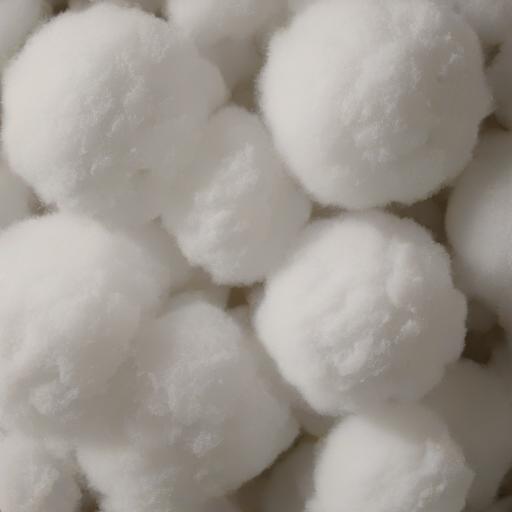}
    & \includegraphics[width=0.2\linewidth]{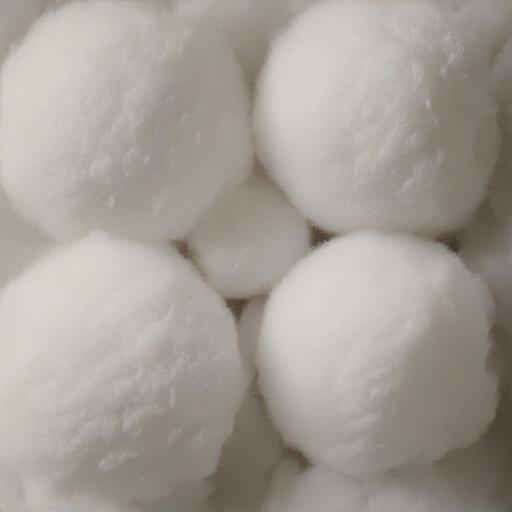}
    & \includegraphics[width=0.2\linewidth]{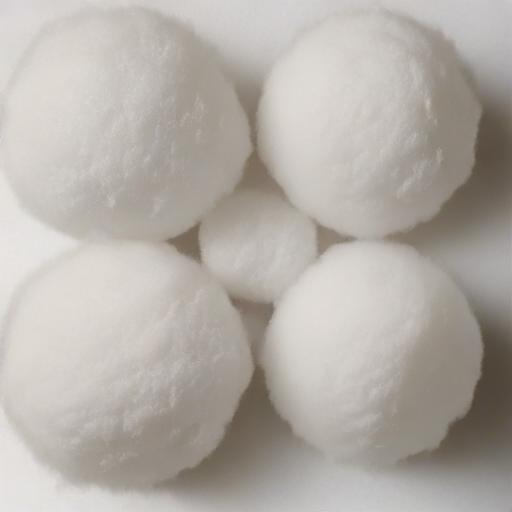}
    & \includegraphics[width=0.2\linewidth]{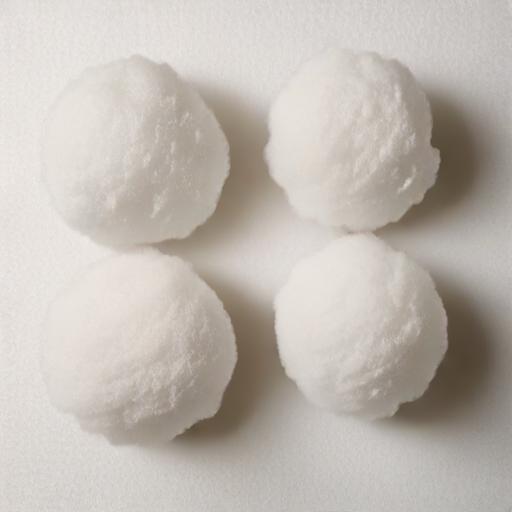} \\
    \rotatebox{90}{4 cuttons} & \includegraphics[width=0.2\linewidth]{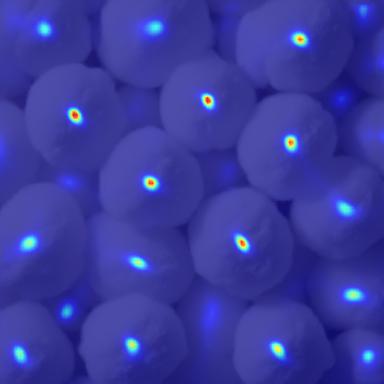}
    & \includegraphics[width=0.2\linewidth]{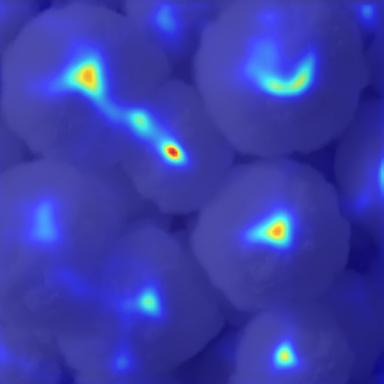}
    & \includegraphics[width=0.2\linewidth]{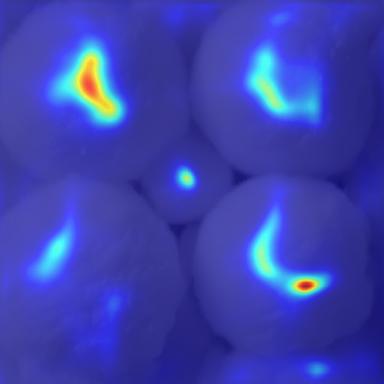}
    & \includegraphics[width=0.2\linewidth]{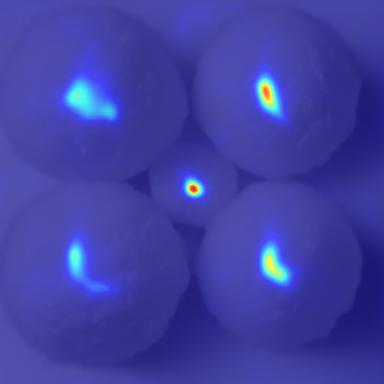}
    & \includegraphics[width=0.2\linewidth]{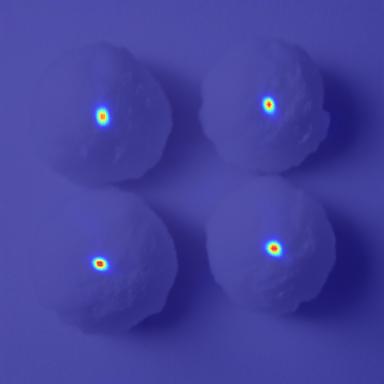} \\
    \end{tabular}%
  }
  \caption{Illustration of the optimization process over time for the prompts ``A photo of 5 cupcakes'' and ``A photo of 4 cotton balls.'' Shown are the generated images and the object density maps.}
  \label{fig:optimization_results}
\end{figure}

\paragraph{Dynamic Detection-Driven Counting.} 
Here, we dynamically update the scaling factor at each step, guided by a detection model. Counting with detection-based methods involves identifying instances of a target class and summing them. However, detection models such as YOLO~\cite{hussain2023yolo} are discrete and non-differentiable, which disrupts the continuity required for gradient-based optimization.

To incorporate detection signals while maintaining differentiability, we adjust the density map’s scaling factor at each optimization step, without backpropagating through the detection model (see Fig.~\ref{fig:method_diagram}). Specifically, at iteration $k$, we define the detection-aligned count as:
\begin{equation}
\operatorname{Count}(x^{(k)}, c) = \sum_{i=1}^{h \cdot w} \hat{\lambda}_{\text{scale}}(x^{(k)}, c) \cdot \Phi_i(x^{(k)}, c),
\end{equation}
where the dynamic scaling factor is computed as:
\begin{equation}
\hat{\lambda}_{\text{scale}}(x^{(k)}, c) = \frac{\text{DetectionCount}(x^{(k)}, c)}{\sum_{j=1}^{h \cdot w} \Phi_j(x^{(k)}, c)},
\end{equation}
and $\text{DetectionCount}(x^{(k)}, c)$ denotes the number of objects of class $c$ detected in image $x^{(k)}$ by a pre-trained detection model.

Importantly, the dynamic scaling factor (both the detection count and the density summation) is computed as a score and is not involved in backpropagation. This formulation allows the detection-derived count to act as a fixed target, enabling gradients to flow through the density map while aligning the soft, potentially inaccurate count signal with a detection-based objective at each iteration. Intuitively, as illustrated in Fig.~\ref{fig:potential}, when the density is spread broadly (e.g., right, bottles), the detection model decreases the scaling factor compared to more concentrated distributions (e.g., left, oranges).

\paragraph{CLIP Matching Penalty.}  
Optimizing for object count alone may distort image semantics. To preserve class identity, we add a CLIP-based semantic loss:
\begin{equation}
\mathcal{L}_{\text{semantic}}(x, c) = \operatorname{CLIP}(x, \text{``A photo of } c \text{''}),
\end{equation}
where $\operatorname{CLIP}$ measures image-text similarity. The full objective is:
\begin{equation}
\mathcal{L}(x, c, N) = \mathcal{L}_{\text{counting}}(x, c, N) + \lambda \mathcal{L}_{\text{semantic}}(x, c),
\end{equation}
with $\lambda$ balancing count accuracy and semantic fidelity. While modern diffusion models often preserve semantics, we include this term for robustness, especially with older models (see supplementary for examples).

\subsubsection{Token Reuse.}
\label{sec:method_reuse}
\begin{figure*}[t]
  \centering
  \setlength{\tabcolsep}{2pt}
  \renewcommand{\arraystretch}{1.1}
  \resizebox{\textwidth}{!}{%
    \begin{tabular}{c ccc|ccc|ccc}
      &
      \multicolumn{3}{c|}{\textbf{\shortstack{Token: 10 oranges}}} &
      \multicolumn{3}{c|}{\textbf{\shortstack{Token: 15 pigeons}}} &
      \multicolumn{3}{c}{\textbf{\shortstack{Token: 4 bottles}}} \\

      &
      \textbf{Trained} & \multicolumn{2}{c|}{\textbf{Reused}} &
      \textbf{Trained} & \multicolumn{2}{c|}{\textbf{Reused}} &
      \textbf{Trained} & \multicolumn{2}{c}{\textbf{Reused}} \\

      &
      \textbf{10 oranges} & \textbf{…and a dog} & \textbf{strawberries} &
      \textbf{15 pigeons} & \textbf{painting} & \textbf{crows} &
      \textbf{4 bottles} & \textbf{mice} & \textbf{cards in ocean} \\

      \rotatebox{90}{\textbf{SD}} &
      \includegraphics[width=0.13\textwidth]{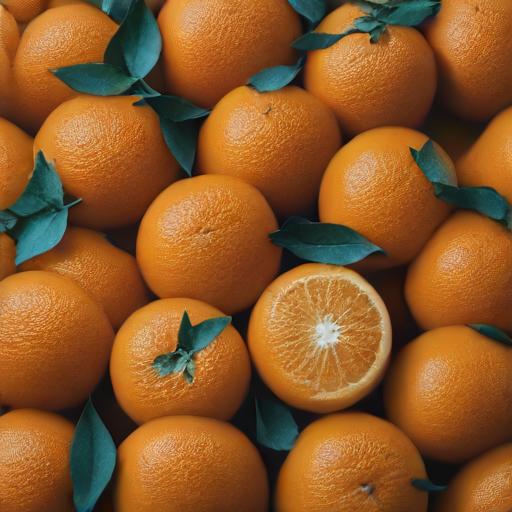} &
      \includegraphics[width=0.13\textwidth]{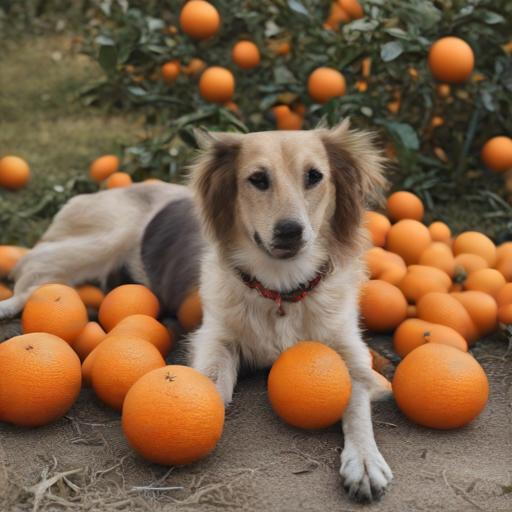} &
      \includegraphics[width=0.13\textwidth]{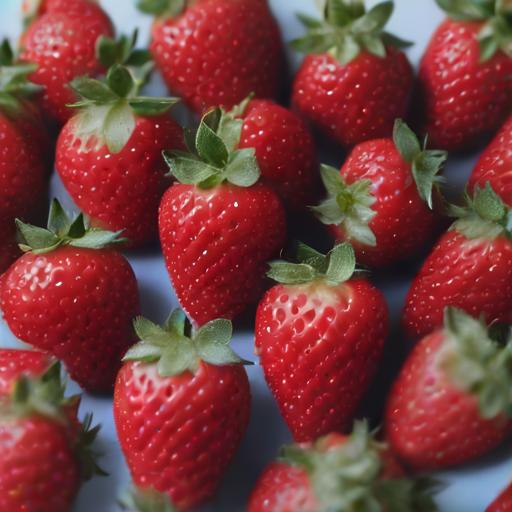} &
      \includegraphics[width=0.13\textwidth]{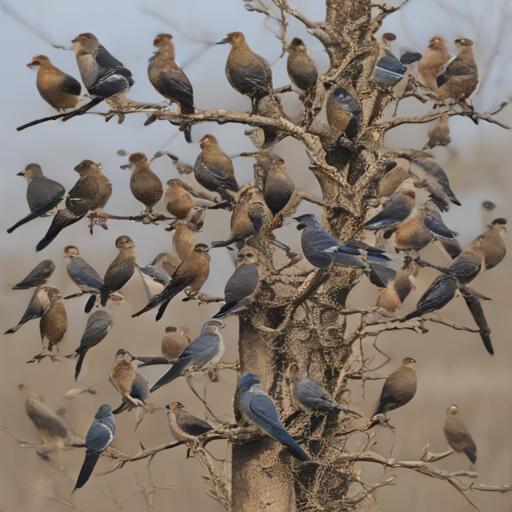} &
      \includegraphics[width=0.13\textwidth]{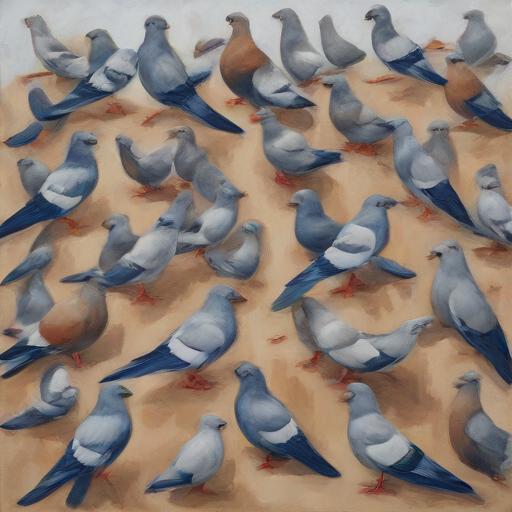} &
      \includegraphics[width=0.13\textwidth]{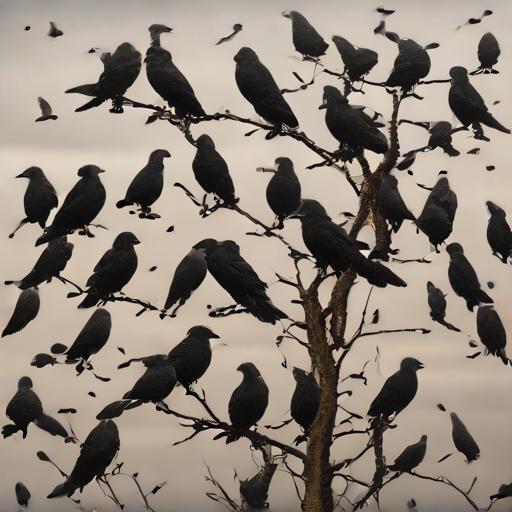} &
      \includegraphics[width=0.13\textwidth]{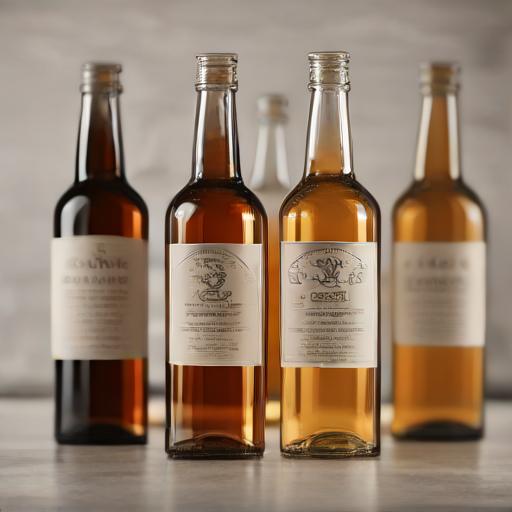} &
      \includegraphics[width=0.13\textwidth]{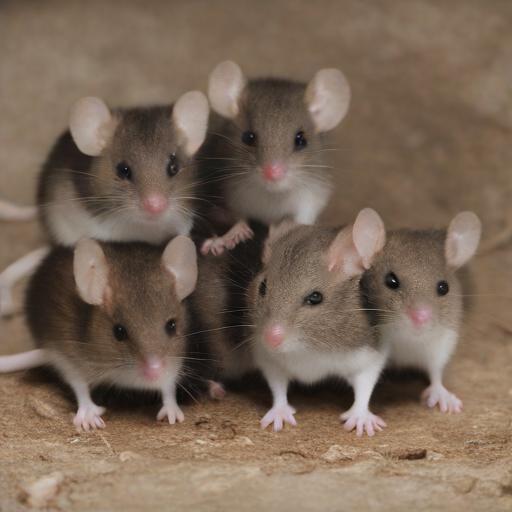} &
      \includegraphics[width=0.13\textwidth]{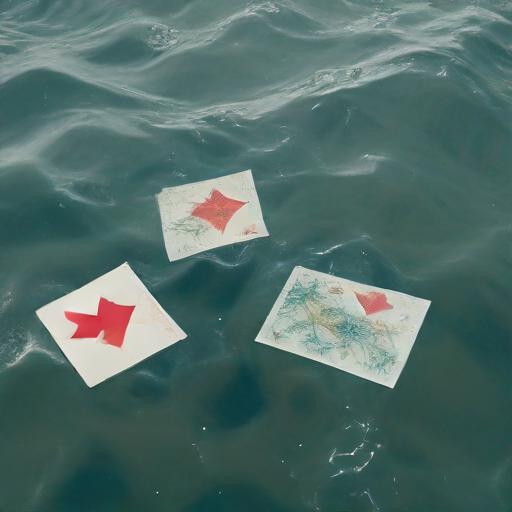} \\

      \rotatebox{90}{\textbf{Ours}} &
      \includegraphics[width=0.13\textwidth]{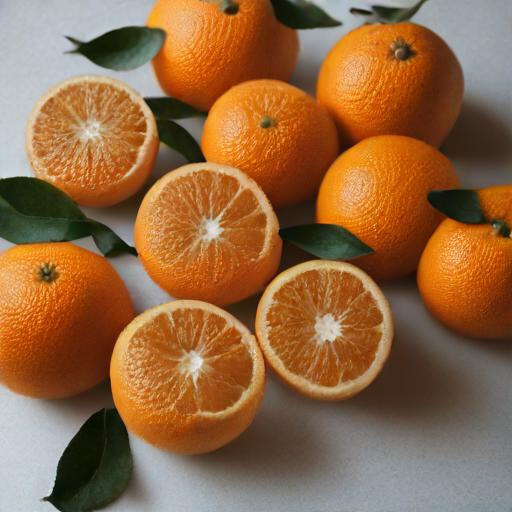} &
      \includegraphics[width=0.13\textwidth]{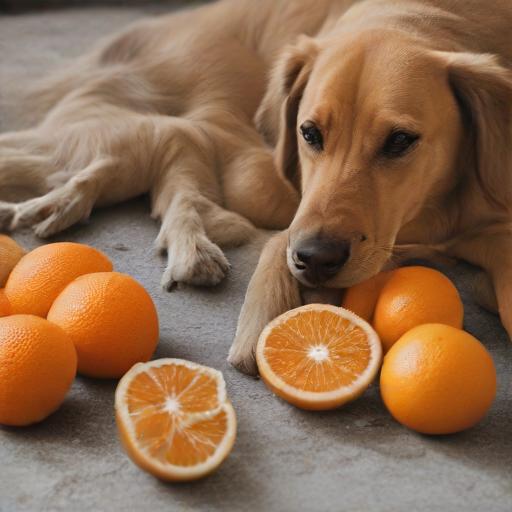} &
      \includegraphics[width=0.13\textwidth]{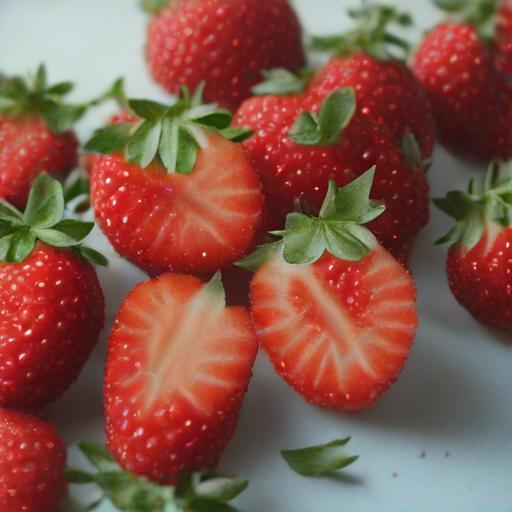} &
      \includegraphics[width=0.13\textwidth]{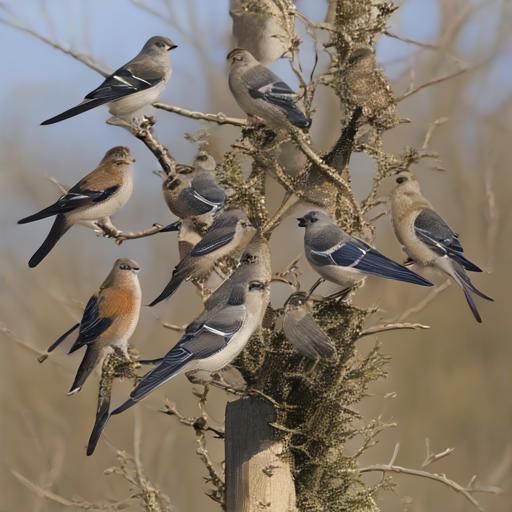} &
      \includegraphics[width=0.13\textwidth]{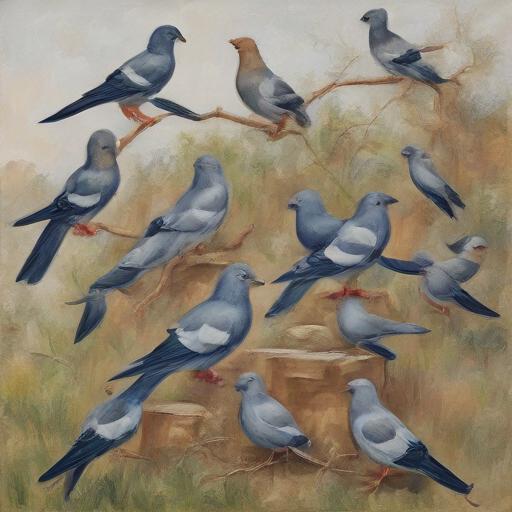} &
      \includegraphics[width=0.13\textwidth]{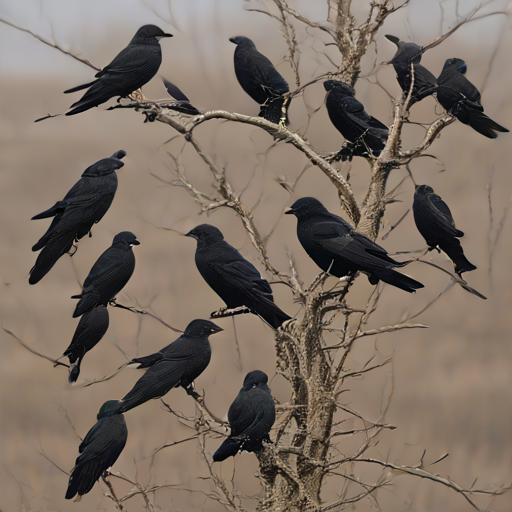} &
      \includegraphics[width=0.13\textwidth]{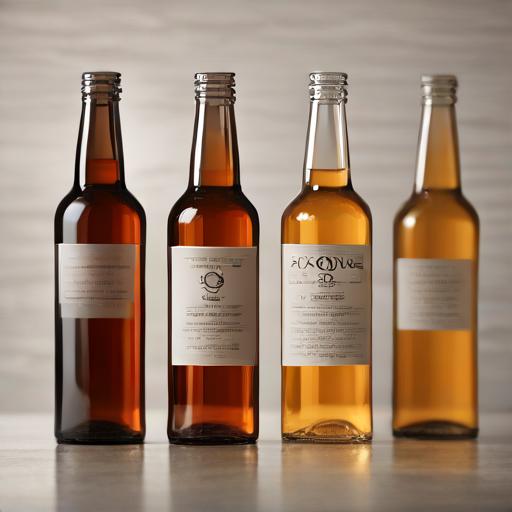} &
      \includegraphics[width=0.13\textwidth]{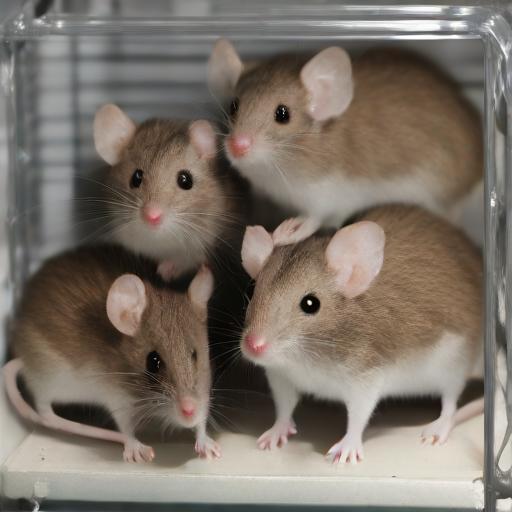} &
      \includegraphics[width=0.13\textwidth]{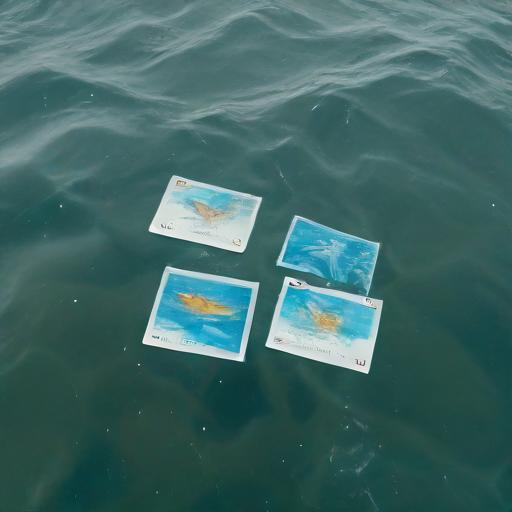}
    \end{tabular}
  }
  \caption{Token reuse across classes and prompts. A token trained on 10 oranges is reused for prompts with a dog and 10 strawberries. A 15-pigeon token generalizes to paintings and 15 crows. We also show out-of-domain reuse: a token trained on four bottles is applied to scenes with four mice and four cards in the ocean. Our method improves object counts across all cases without requiring inference-time optimization. SD-Turbo often fails to generate the correct number of objects.
}
  \Description{A grid of 9 columns. Top row: prompt labels. Row 2: SD-generated images. Row 3: our optimized results. Columns include oranges, dogs, strawberries, pigeons, crows, and three new cases.}
  \label{fig:token_reuse_final}
\end{figure*}

While optimization is performed with a simple prompt to stabilize the training and focus on object counting, the resulting token embedding generalizes well. At inference, without any further optimization, it improves counting in novel prompts and even unseen classes, suggesting it captures quantity semantics beyond the optimized category (see Fig.~\ref{fig:token_reuse_final}).

\section{Results}

\paragraph{Benchmark.}  We have designed a benchmark to test various classes and quantities. We employ the same classes as the FSC-147 dataset, a common dataset for object counting comprising 147 categories, including common objects, vehicles, and animals~\cite{ranjan2021learning}. We generated 25 examples for each number and each class, ranging from one to 25, resulting in 3,674 samples.

\paragraph{Baselines.}
We evaluate \textit{CountGen}~\cite{binyamin2024make} and \textit{Counting-Guidance}~\cite{kang2025counting}. Since the baselines use different backbones, we also report results per backbone.

\paragraph{Metrics.} We report exact-match accuracy using a combined YOLO and DETR verifier (one agreement), allowing a margin of one count for quantities over two. Human verification showed a high correlation with automated accuracy, confirming the metric’s reliability. We also report MAE and RMSE, using YOLO, DETR, and \clipcount. Since YOLO supports a limited set of object classes, we apply it only to the 20 classes in our dataset that overlap with FSC-147.  Semantic alignment is measured by the normalized CLIP score, \(\operatorname{CLIP\text{-}S}(p,x) = w \cdot \max(\cos(p, x), 0)\), where \( p \) is the prompt and \( x \) is the generated image~\cite{hessel2021clipscore}.

\begin{table*}[t!]
  \centering
  \caption{Evaluation results for object-count accuracy using three counting mechanisms: (i) \clipcount{} in evaluation mode, (ii) detection-based counting with YOLO, and (iii) detection-based counting with DETR. We also report the CLIP-S score for semantic correspondence. Arrows indicate the direction of improvement: $\downarrow$ lower is better (MAE/RMSE), $\uparrow$ higher is better ($\Delta$MAE) for difference from the backbone.}
  \begin{adjustbox}{width=0.9\textwidth}
    \begin{tabular}{llcccccccccc}
      \toprule
      \textbf{Model} & \textbf{Backbone} &
        \multicolumn{2}{c}{\textbf{\clipcount{}}} &
        \multicolumn{3}{c}{\textbf{YOLO}} &
        \multicolumn{3}{c}{\textbf{DETR}} &
        \textbf{CLIP\textendash S} & \textbf{Accuracy (\%)} \\
      \cmidrule(r){3-4}\cmidrule(r){5-7}\cmidrule(r){8-10}
      & &
        \thead{MAE $\downarrow$} & \thead{RMSE $\downarrow$} &
        \thead{MAE $\downarrow$} & \thead{RMSE $\downarrow$} & \thead{$\Delta$MAE $\uparrow$} &
        \thead{MAE $\downarrow$} & \thead{RMSE $\downarrow$} & \thead{$\Delta$MAE $\uparrow$} & & \\
      \midrule
      \multicolumn{12}{c}{\textbf{Up to 9 objects benchmark}} \\
      \midrule
      SD-XL                      & SD-XL    & 5.34 & 6.70  & 5.90  & 8.95  & 0.00 & 3.61 & 5.58 & 0.00 & 68.21 & 28\% \\
      SD-Turbo                   & SD-Turbo & 7.62 & 19.10 & 10.32 & 17.64 & 0.00 & 6.99 & 11.29 & 0.00 & 70.54 &  26.9\%\\
      \midrule
      CountGen                   & SD-XL    & 5.96 & 7.55  & 5.49  & 10.75 & 0.41 & 3.53 & 7.62 & 0.08 & 69.27 & 31\% \\
      Ours + Static Scale        & SD-Turbo & 5.32 & 9.37  & 6.92  & 11.03 & 3.40 & 4.97 & 7.95 & 2.02 & 70.72 &  23.97\%\\
      Ours + Dynamic Scale (YOLO)& SD-Turbo & \textbf{4.54} & \textbf{6.39}  & \textbf{3.04}  & \textbf{5.61}  & \textbf{7.28} & 3.42 & 6.15 & 3.57 & 70.12 & 38.01\% \\
      Ours + Dynamic Scale (DETR)& SD-Turbo & 5.00 & 8.65  & 4.95  & 9.64  & 5.37 & \textbf{2.51} & \textbf{4.84} & \textbf{4.48} & \textbf{70.41} & \textbf{45.61\%} \\
      \midrule
      \multicolumn{12}{c}{\textbf{Full benchmark}} \\
      \midrule
      SD v1.4                    & SD v1.4  & 9.72 & 14.54 & 10.53 & 13.62 & 0.00 & 9.73 & 12.68 & 0.00 & 70.54 & 11.15\% \\
      SD-Turbo                   & SD-Turbo & 11.62 & 19.69 & 17.10 & 23.01 & 0.00 & 13.67 & 18.26 & 0.00 & \textbf{70.66} & 12.63\% \\
      \midrule
      Counting-Guidance          & SD v1.4  & 9.60 & 15.71 & 10.38 & 13.90 & 0.15 & 8.84 & 11.52 & 0.91 & 69.90 & 9.47\% \\
      Ours + Static Scale        & SD-Turbo & \textbf{4.92} & \textbf{8.41}  & 11.89 & 16.85 & 5.21 & 8.14 & 11.68 & 5.53 & 70.73 & 13.37\% \\
      Ours + Dynamic Scale (YOLO)& SD-Turbo & 7.55 & 13.99 & \textbf{7.41}  & \textbf{11.75} & \textbf{9.69} & 7.92 & 11.71 & 5.75 & 70.59 & 18.73\% \\
      Ours + Dynamic Scale (DETR)& SD-Turbo & 6.88 & 13.02 & 10.48 & 16.17 & 6.62 & \textbf{5.44} & \textbf{8.45} & \textbf{8.23} & 70.61 & \textbf{24\%} \\
      \bottomrule
    \end{tabular}
  \end{adjustbox}
  \label{tab:results}
\end{table*}

\subsection{Qualitative results}

We begin with qualitative results, which provide intuitive insight into the model’s behavior. In Fig.~\ref{fig:results_basic}, from left to right, we first show that our method can make fine-grained adjustments, such as removing a single object. For example, it reduces four birds to three or six bowls to five. Next, we illustrate that Stable Diffusion often overestimates object counts. A prompt requesting five chairs may result in more than twenty. Similarly, prompting for 25 objects can produce a significantly larger number. In contrast, our method brings the count close to the target. In some cases, our model also causes a zoom in to support more accurate counting.

Fig.~\ref{fig:optimization_results} illustrates how the generated image and its density map change over iterations, revealing the dynamics of the optimization process. For categories like cotton balls, the density map often appears distorted, an issue our dynamic scaling effectively corrects. The method progressively reduces the object count at each step. \textit{Notably, changes are minimal}: visual attributes remain consistent, making the approach well-suited for fine-grained editing of generated content (see Fig.~\ref{fig:comparison} for more examples).

Fig.~\ref{fig:examples_vs_basline} shows qualitative comparisons between baseline methods and different variants of our approach. Our method effectively corrects object counts with minimal changes to the image content. In contrast, Counting-Guidance often fails to adjust the count, leaving the image largely unchanged, while CountGen frequently alters the overall visual composition, as it aggressively intervenes in the denoising process by adding and removing objects.

\begin{figure*}[t!]
  \centering
    \begin{subfigure}[t]{0.32\textwidth}
    \includegraphics[width=\textwidth]{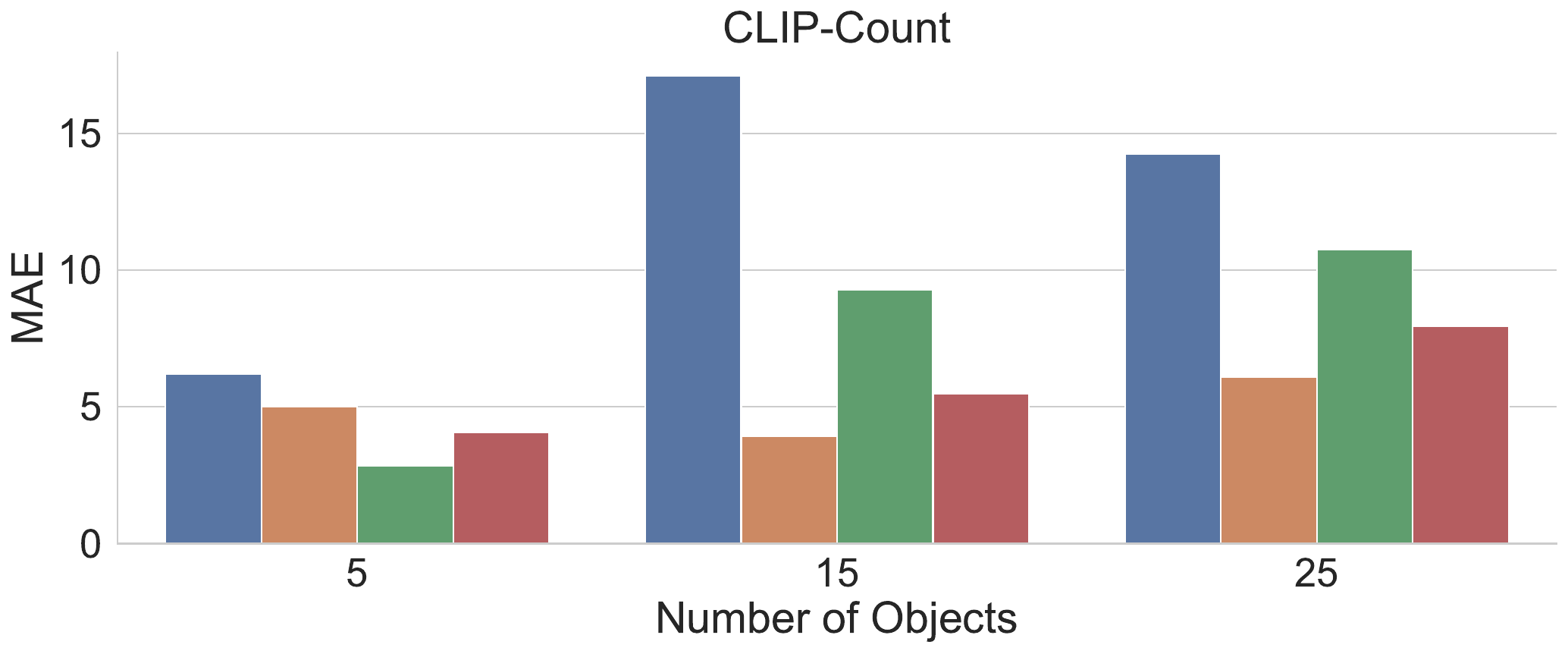}
    \label{fig:mae_clipcount}
  \end{subfigure}\hfill
  \begin{subfigure}[t]{0.32\textwidth}
    \includegraphics[width=\textwidth]{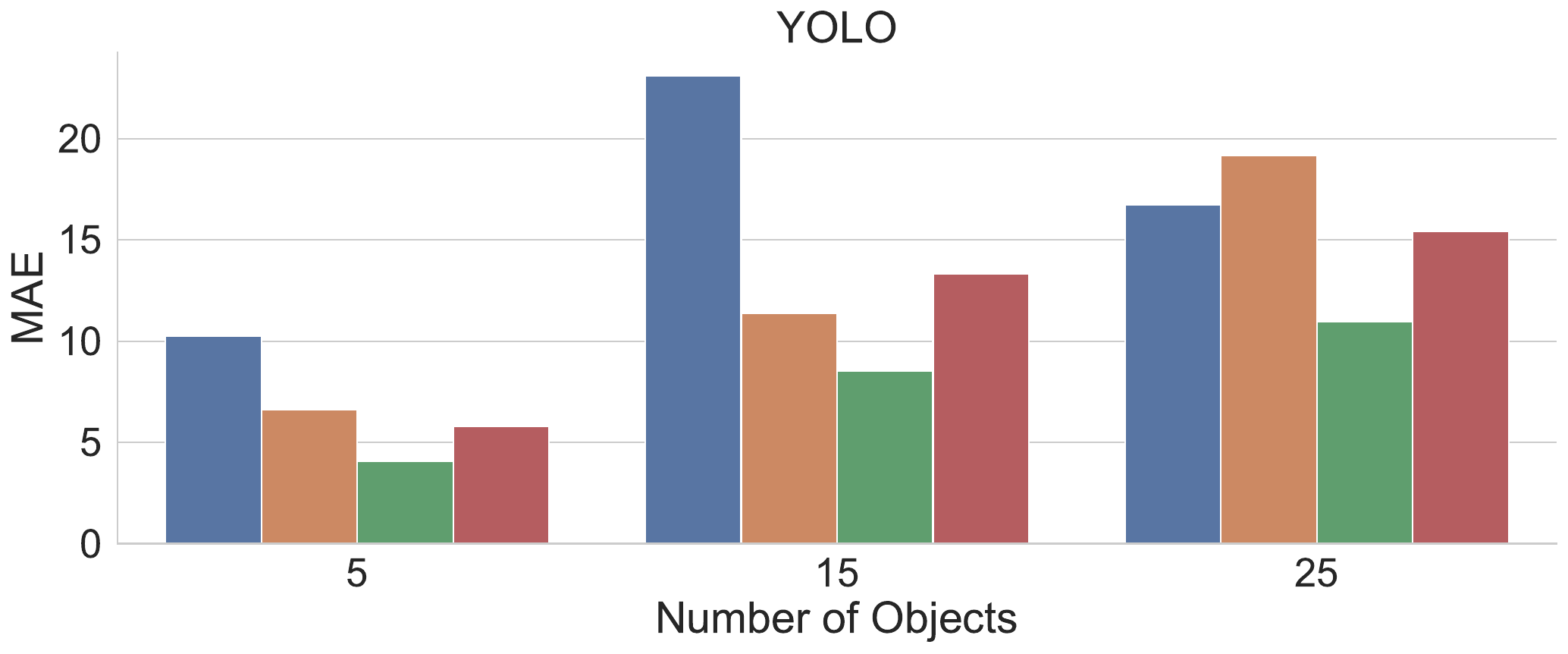}
    \label{fig:mae_yolo}
  \end{subfigure}\hfill
  \begin{subfigure}[t]{0.32\textwidth}
    \includegraphics[width=\textwidth]{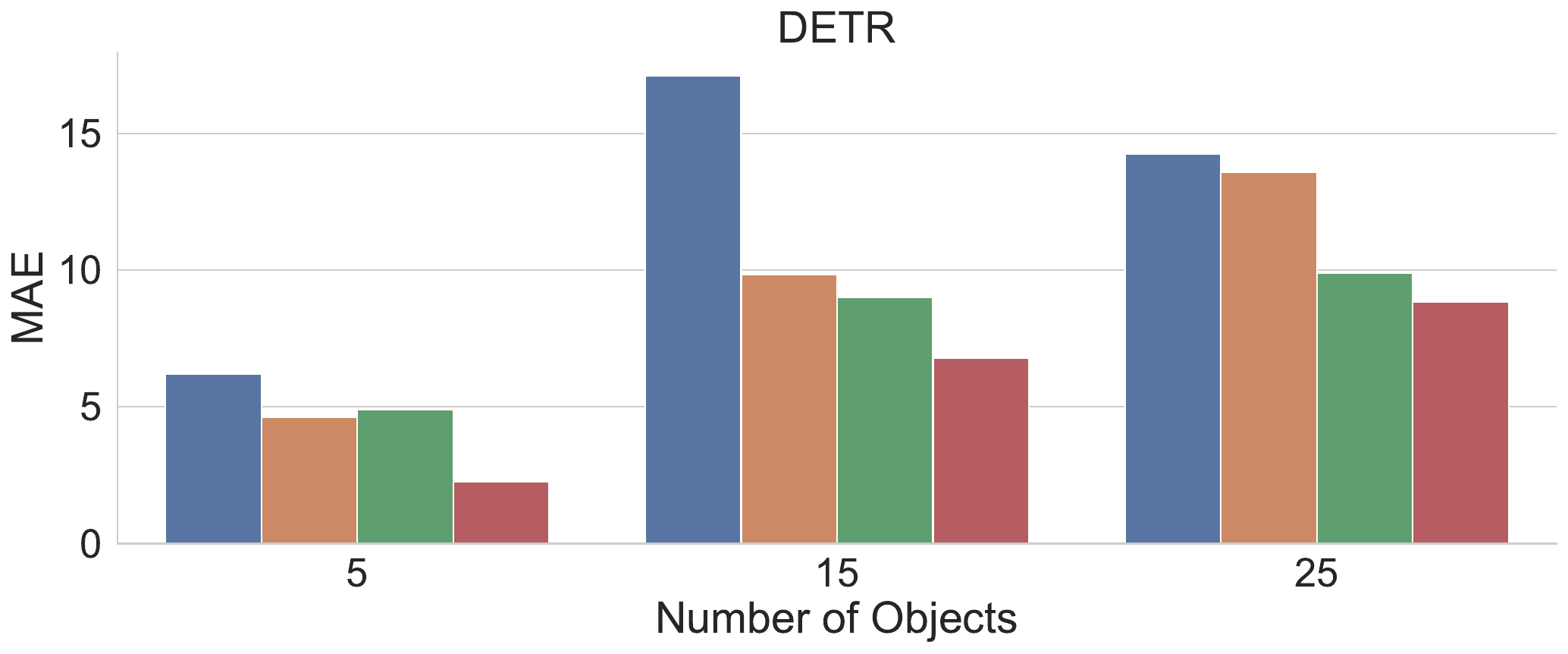}
    \label{fig:mae_detr}
  \end{subfigure}
  \vspace{-2em} %
\caption{Error for counting sets of 5, 15, and 25 objects using \clipcount{}, YOLO, and DETR. \textcolor{blue}{Blue} is Stable Diffusion (SD), \textcolor{orange}{orange} is our static scaling, \textcolor{green!60!black}{green} is our dynamic scaling with YOLO, and \textcolor{red}{red} is dynamic scaling with DETR.} \vspace{-7pt}

  \label{fig:different_count}
\end{figure*}

\begin{figure}[t!]
    \centering
    \includegraphics[width=0.80\linewidth]{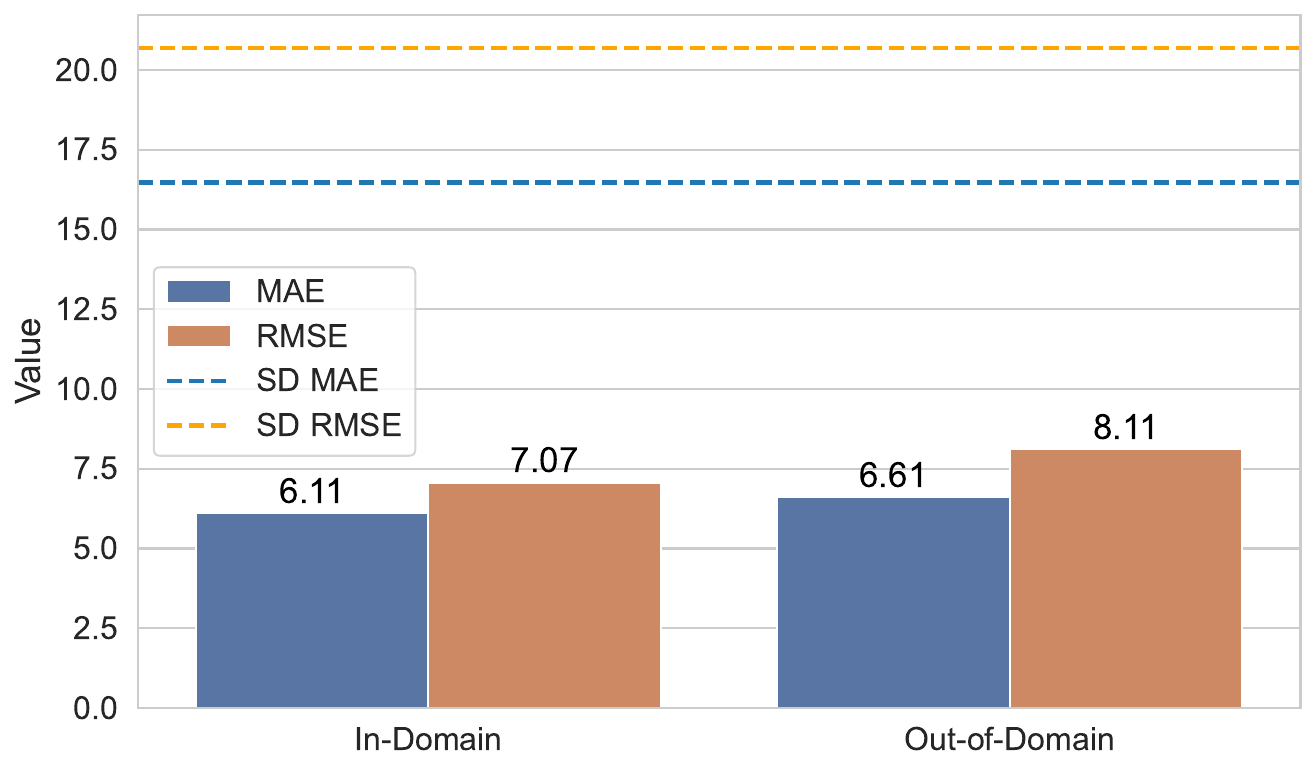}
    \caption{Token reuse evaluation. The ability of trained tokens to improve generation for other classes is divided into in- and out-of-domain classes.}
    \Description{A bar chart showing token reuse performance. Bars are grouped into in-domain and out-of-domain classes, illustrating how trained tokens affect generation across categories.}
    \label{fig:token_reuse_experiment}
    \vspace{-10pt}
\end{figure}

\begin{figure}[t!]
    \centering
    \begin{subfigure}[t]{0.49\linewidth}
        \centering
        \includegraphics[width=\linewidth]{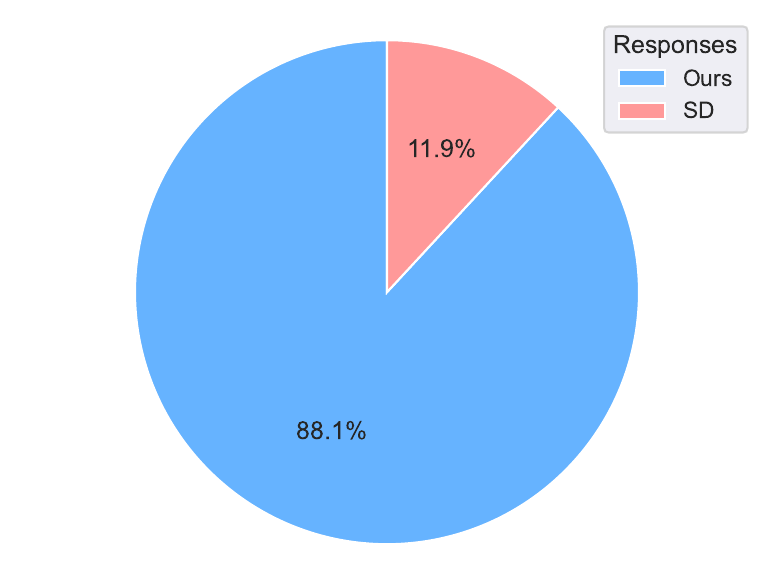}
        \caption{Object count accuracy.}
        \Description{Bar chart comparing the accuracy of object count estimations in generated images between different methods.}
        \label{fig:human_study_q1}
        \vspace{-1em}
    \end{subfigure}
    \hfill
    \begin{subfigure}[t]{0.49\linewidth}
        \centering
        \includegraphics[width=\linewidth]{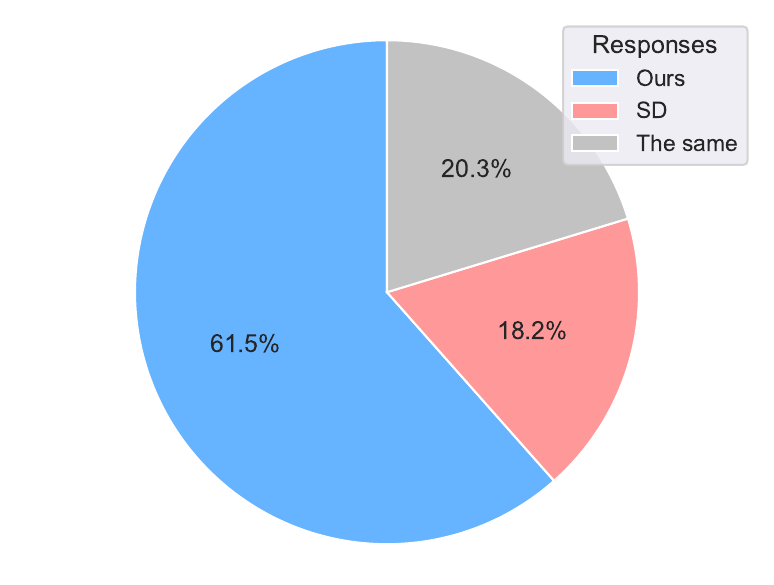}
        \caption{Image naturalness.}
        \Description{Bar chart comparing how natural the generated images appear to human raters for two different methods.}
        \label{fig:human_study_q2}
    \end{subfigure}
    \caption{Human study comparisons between our method and SD.}
    \Description{Two subfigures showing results of a human study. One measures object count accuracy and the other rates image naturalness, comparing our method with Stable Diffusion.}
    \label{fig:human_study}
    \vspace{-17pt}
\end{figure}

In Fig.~\ref{fig:token_reuse_final}, we demonstrate the ability to reuse trained tokens. For example, a token trained on the prompt ``A photo of 10 oranges'' was successfully applied to generate ``A photo of 10 strawberries,'' and similarly for prompts involving pigeons and crows. Moreover, the token generalizes to more complex prompts, such as ``10 oranges and a dog'' or ``A painting of 15 pigeons.'' This capability may reduce the need for repeated inference-time optimization to obtain counting tokens for specific quantities.

The above qualitative studies demonstrate that our method improves object counts with minimal changes and, through token reuse, generalizes to complex prompts without requiring additional inference-time optimization. We now quantitatively validate these results against existing methods.

\subsection{Quantitative Results}

In Tab.~\ref{tab:results}, we compare our method against the recently released CountGen. Since CountGen is trained only on prompts containing up to ten objects, we adjusted our evaluation datasets accordingly. While the average object count difference for CountGen is relatively small (0.41 according to YOLO and 0.08 according to DETR), it still yields a noticeable improvement in accuracy (from 28\% to 31\%). Our static variant reduces the error on both YOLO and DETR metrics but does not improve accuracy. In contrast, our dynamic variant significantly boosts accuracy.

We also demonstrate robustness across detection frameworks. Using YOLO supervision still improves DETR-based metrics compared to the backbone (6.99 vs.\ 3.42), and DETR supervision improves YOLO metrics (10.32 vs.\ 4.95). We also find that stronger detectors (e.g., DETR vs.\ YOLO) lead to better accuracy, highlighting a promising direction: improving discriminative models can enhance generative performance via our plug-and-play method.

In the full benchmark, we observe a much higher challenge indicated by the low accuracy; despite this, our method doubled the current backbone accuracy (24 vs.\ 12.63).

In terms of semantic alignment, all variants of our method achieve similar CLIP-S scores (around 71). In contrast, CountGen and Counting-Guidance show lower scores (69.27 and 69.90, respectively), likely due to training collapse (discussed in the appendix), and substantial image variations in the case of CountGen.

Fig.~\ref{fig:different_count} shows results by count size (5, 15, 25) to highlight behavioral differences. Smaller counts require fine-grained edits, while larger counts often demand substantial image changes. Our method consistently outperforms the SD-Turbo baseline, though higher counts remain more challenging.

In Fig.~\ref{fig:token_reuse_experiment}, We evaluate the reuse of trained counting tokens by defining three groups of classes: (i) fruits (apples, strawberries, tomatoes), (ii) birds (crows, pigeons, seagulls), and (iii) domestic mammals (zebras, horses, cows). For each class, we generate images containing ten objects and assess token reuse both in-domain (within the same group) and out-of-domain (across groups). Reused tokens improve counting accuracy over SD, with in-domain reuse performing slightly better than out-of-domain (6.11 vs. 6.61). This demonstrates that our method can resolve counting without inference-time optimization, even in out-of-domain cases. We stress that this capability has significant implications for the object counting objective, as it shows that generative images can improve counting by pre-training dedicated counting tokens for different quantities.

In Fig.~\ref{fig:human_study}, we evaluate our method versus the SD-Turbo backbone with a human study, addressing inaccuracies in automatic metrics. Fifty participants were shown 20 image pairs generated from the same prompts. They selected (i) the image with the correct number of objects and (ii) the one that looked more natural and semantically aligned, with a ``no difference'' option for the latter. Our method was preferred for count accuracy in 88\% of cases and for naturalness in 79.7\%, with 18\% indicating no difference. These results confirm a significant improvement over the backbone.

\section{Discussions}

\paragraph{Limitations.} Our approach is limited by the performance of the pre-trained models it relies on, particularly the quality of the underlying counting mechanism. For example, using a YOLO-based metric, we observe higher error on classes unseen by \clipcount~(MAE 8.2 vs.\ 7.72). 
On the other hand, the plug-and-play nature of our method, adjustable via scaling, allows easy integration of improved detectors or counting models. For instance, switching to DETR-based scaling improves accuracy. This highlights a promising direction: making image generation more accurate by leveraging guidance from non-differentiable discriminative methods.

\paragraph{Conclusions.}  
This work shows that state-of-the-art text-to-image methods often fail to depict exact object counts. We propose a novel method that improves accuracy by operating on fully generated images, computing a soft count loss, and scaling it using a detection-driven counting mechanism. To address the inefficiency of inference-time optimization, we instead optimize a newly introduced counting token embedding, which can be reused across novel prompts and object classes. We hope our method promotes broader adoption of supervision from external models.

\bibliographystyle{ACM-Reference-Format}
\bibliography{aaai25}

\clearpage
\appendix

\begin{figure*}[t!]
  \centering
  \setlength{\tabcolsep}{2pt}
  \renewcommand{\arraystretch}{1.1}

  \resizebox{0.9\linewidth}{!}{%
  \begin{tabular}{lccccc}
    & \textbf{Ours (DETR)} & \textbf{Ours (YOLO)} & \textbf{Ours (Static)} & \textbf{Counting-Guidance} & \textbf{CountGen} \\ \midrule

    \textbf{3 sheep} &
      \begin{tabular}{@{}c@{}}
        \includegraphics[width=.08\linewidth]{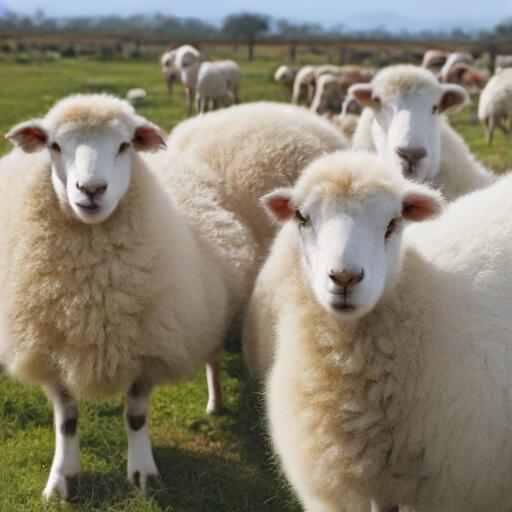}\\[-2pt]
        \includegraphics[width=.08\linewidth]{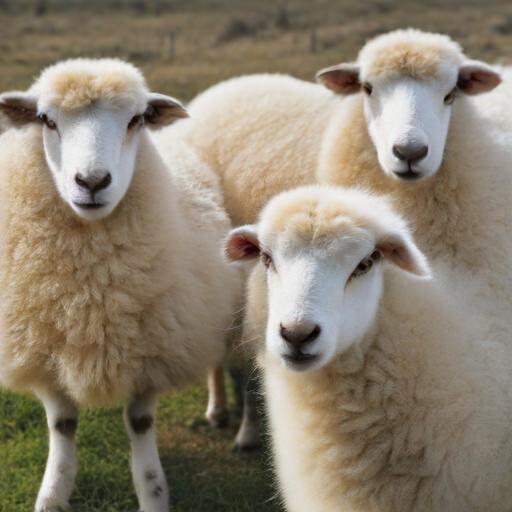}
      \end{tabular} &
      \begin{tabular}{@{}c@{}}
        \includegraphics[width=.08\linewidth]{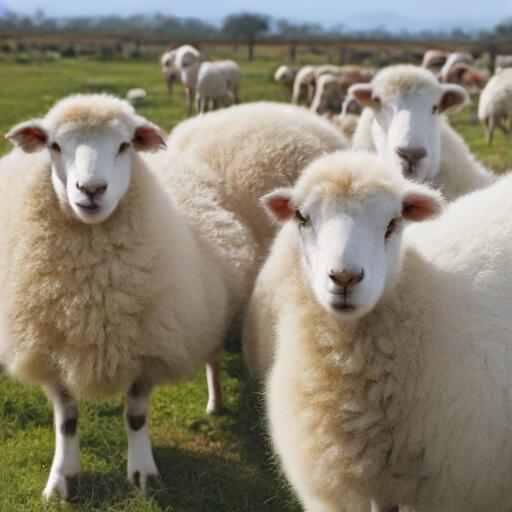}\\[-2pt]
        \includegraphics[width=.08\linewidth]{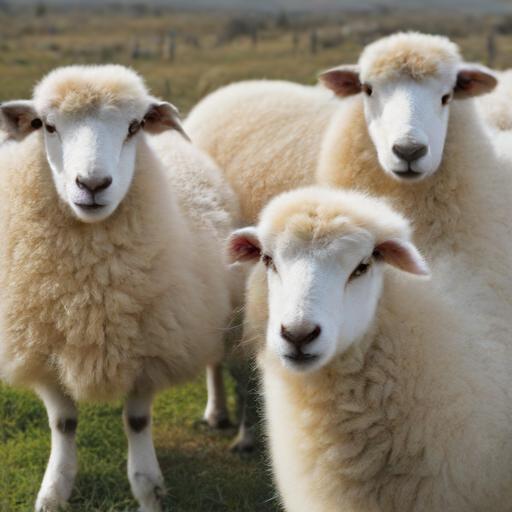}
      \end{tabular} &
      \begin{tabular}{@{}c@{}}
        \includegraphics[width=.08\linewidth]{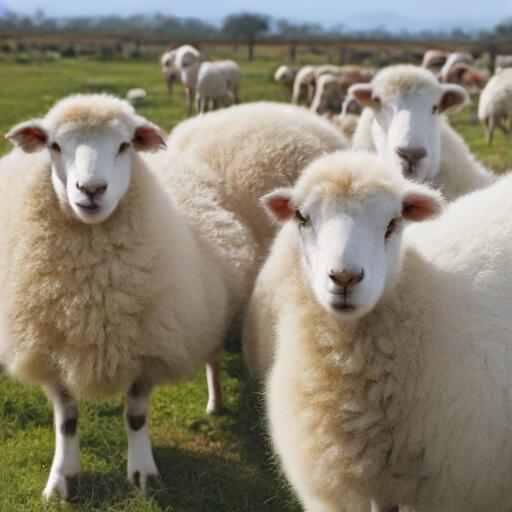}\\[-2pt]
        \includegraphics[width=.08\linewidth]{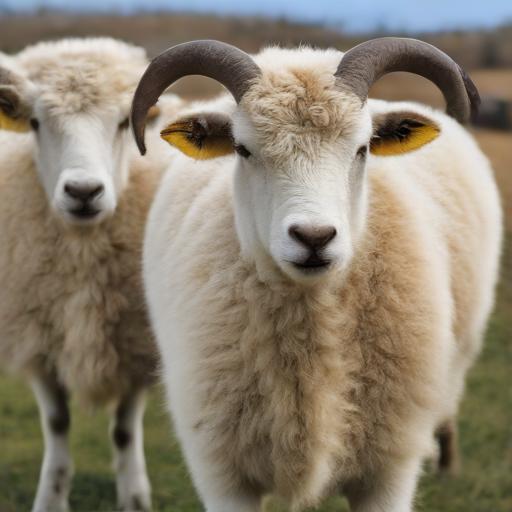}
      \end{tabular} &
      \begin{tabular}{@{}c@{}}
        \includegraphics[width=.08\linewidth]{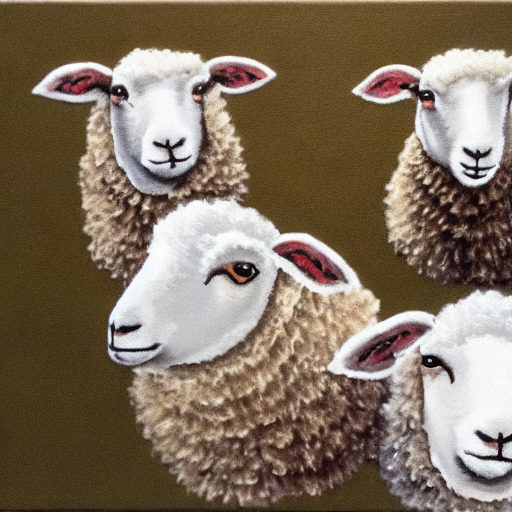}\\[-2pt]
        \includegraphics[width=.08\linewidth]{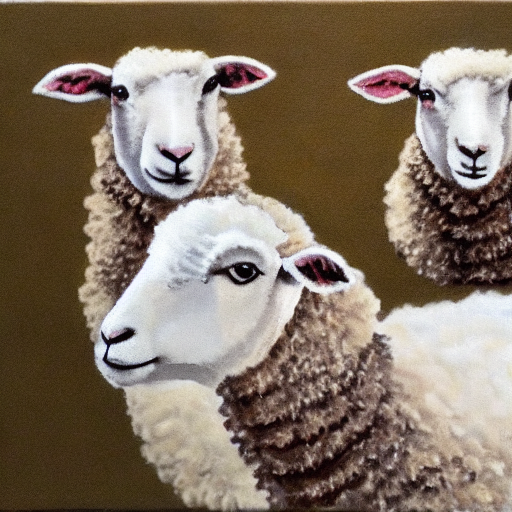}
      \end{tabular} &
      \begin{tabular}{@{}c@{}}
        \includegraphics[width=.08\linewidth]{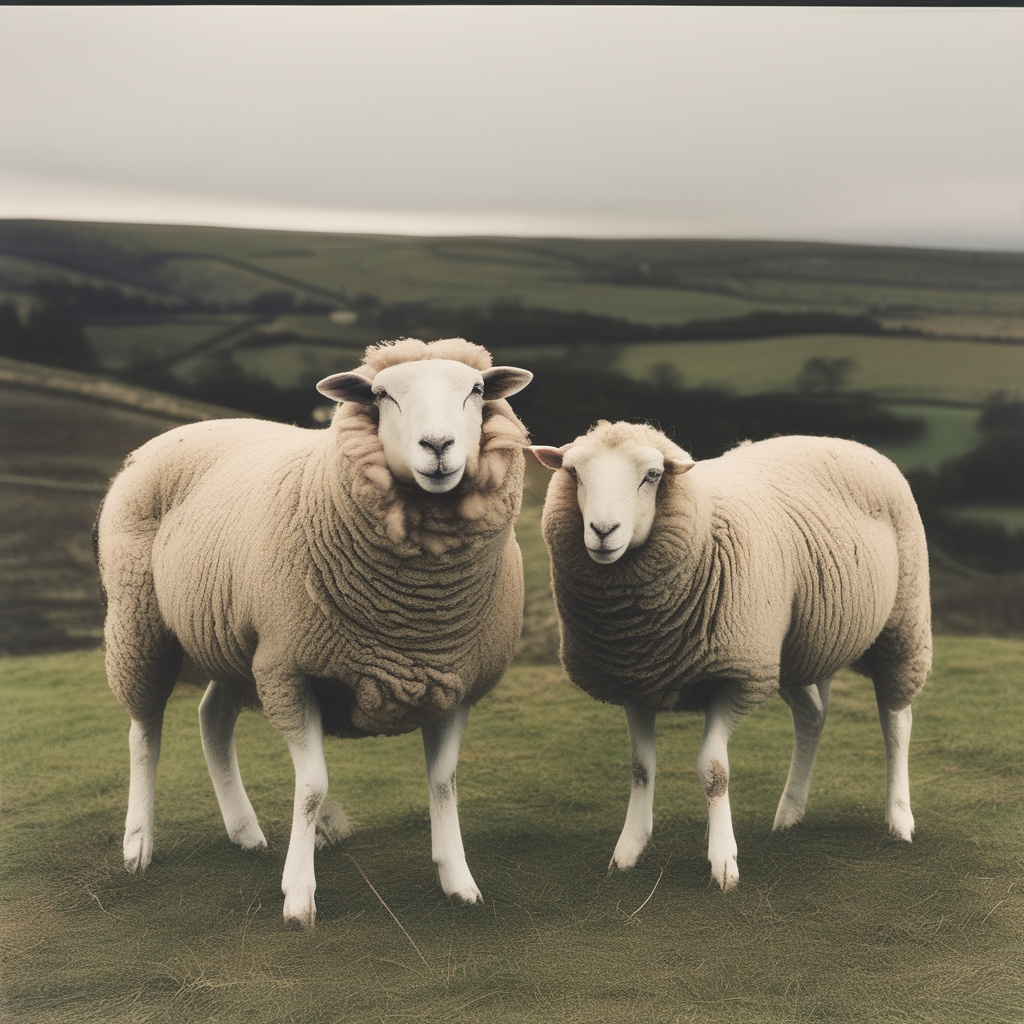}\\[-2pt]
        \includegraphics[width=.08\linewidth]{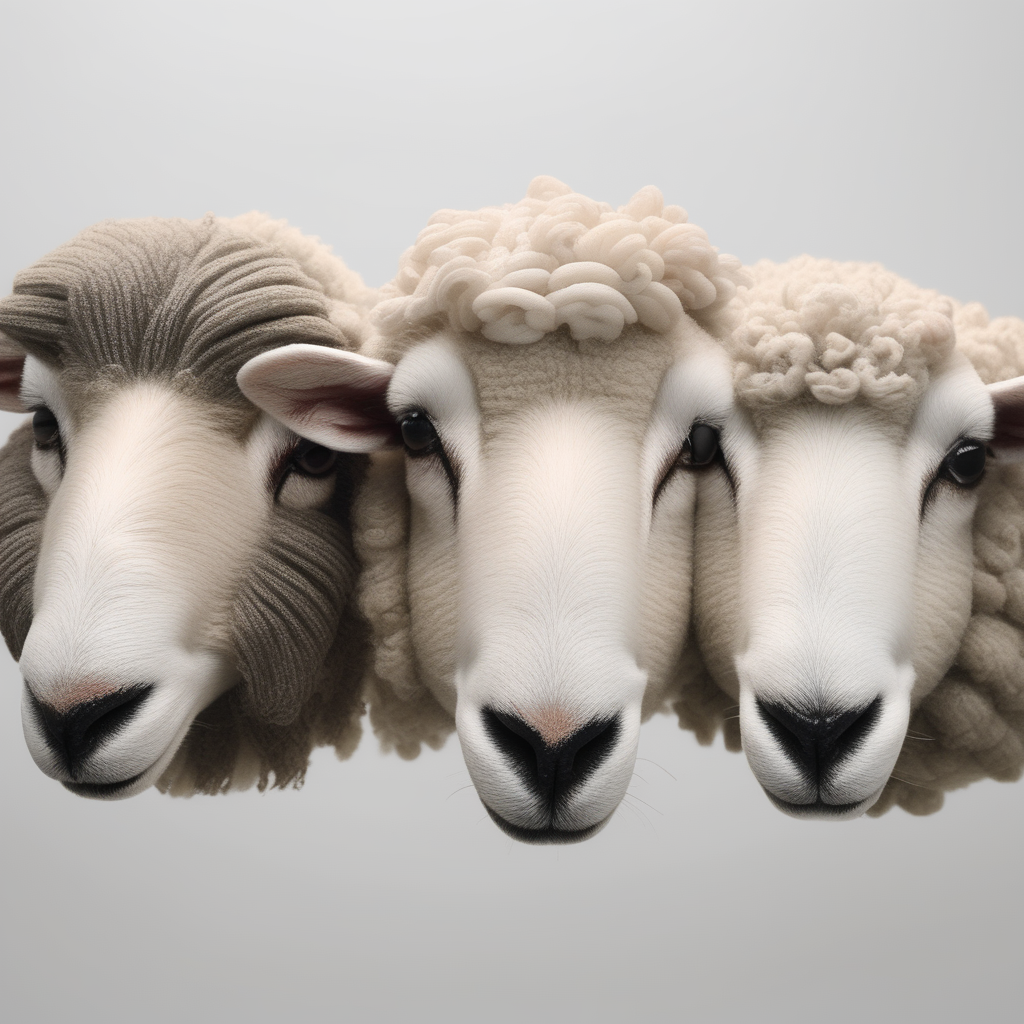}
      \end{tabular} \\

    \textbf{6 carrots} &
      \begin{tabular}{@{}c@{}}
        \includegraphics[width=.08\linewidth]{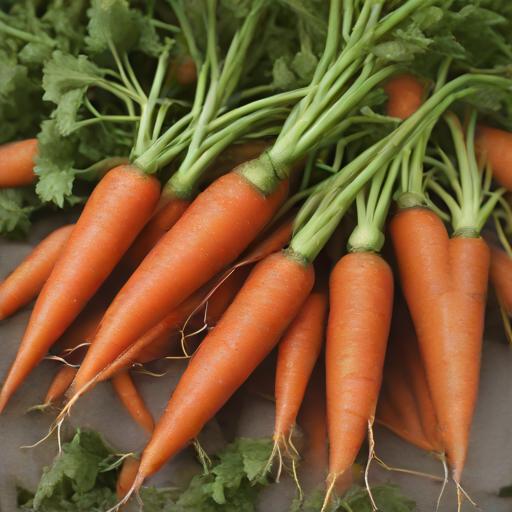}\\[-2pt]
        \includegraphics[width=.08\linewidth]{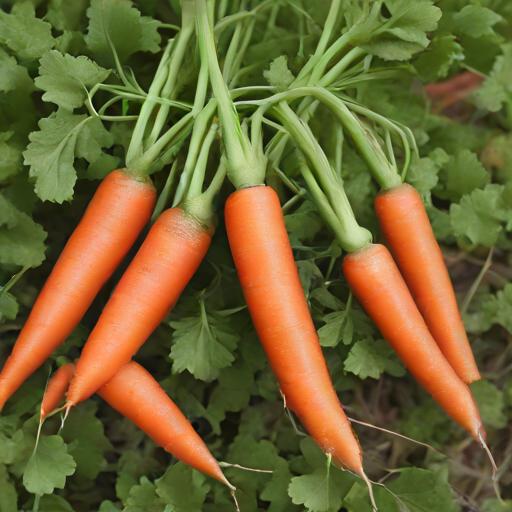}
      \end{tabular} &
      \begin{tabular}{@{}c@{}}
        \includegraphics[width=.08\linewidth]{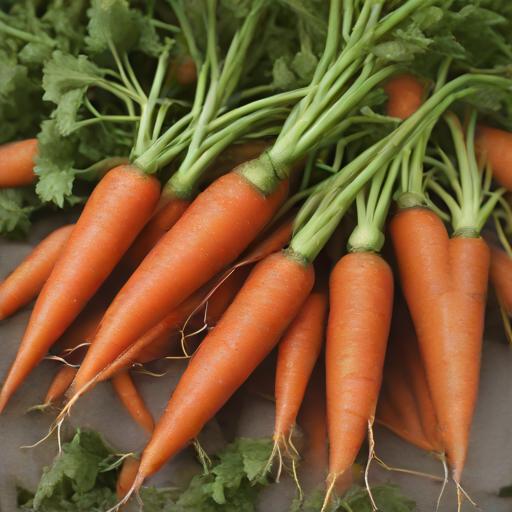}\\[-2pt]
        \includegraphics[width=.08\linewidth]{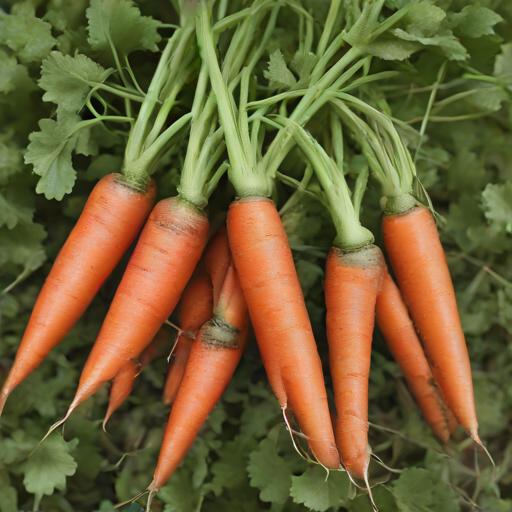}
      \end{tabular} &
      \begin{tabular}{@{}c@{}}
        \includegraphics[width=.08\linewidth]{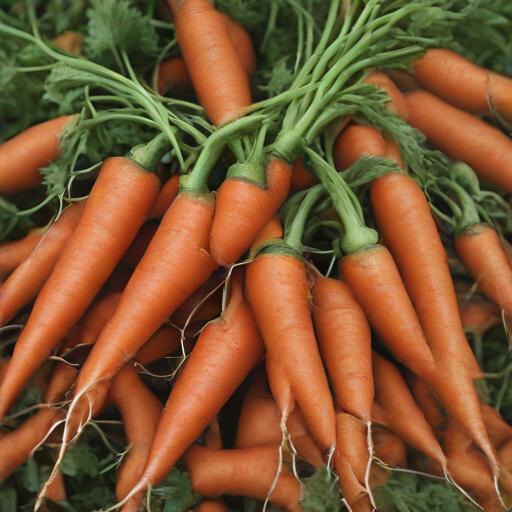}\\[-2pt]
        \includegraphics[width=.08\linewidth]{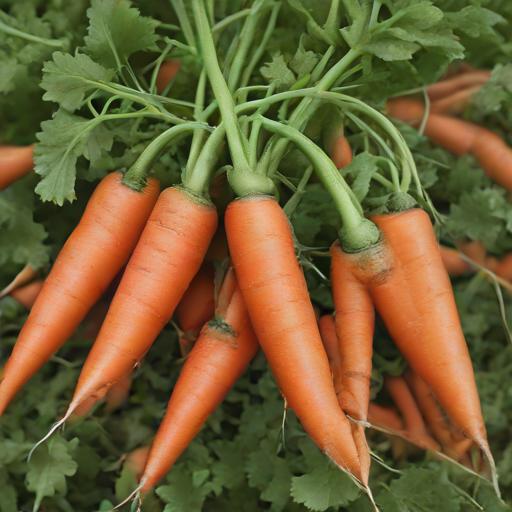}
      \end{tabular} &
      \begin{tabular}{@{}c@{}}
        \includegraphics[width=.08\linewidth]{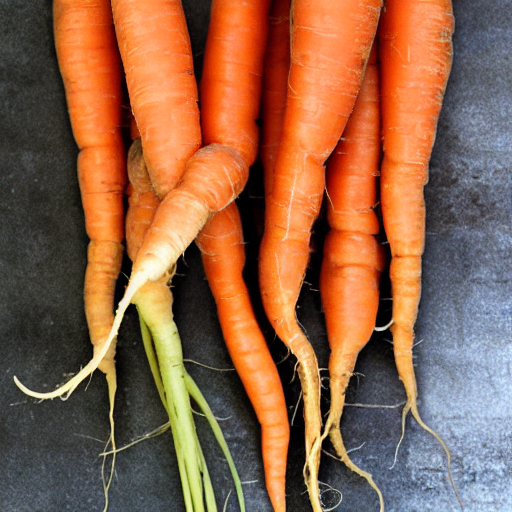}\\[-2pt]
        \includegraphics[width=.08\linewidth]{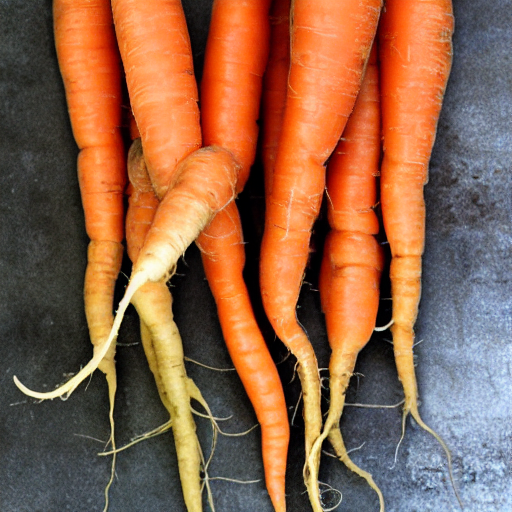}
      \end{tabular} &
      \begin{tabular}{@{}c@{}}
        \includegraphics[width=.08\linewidth]{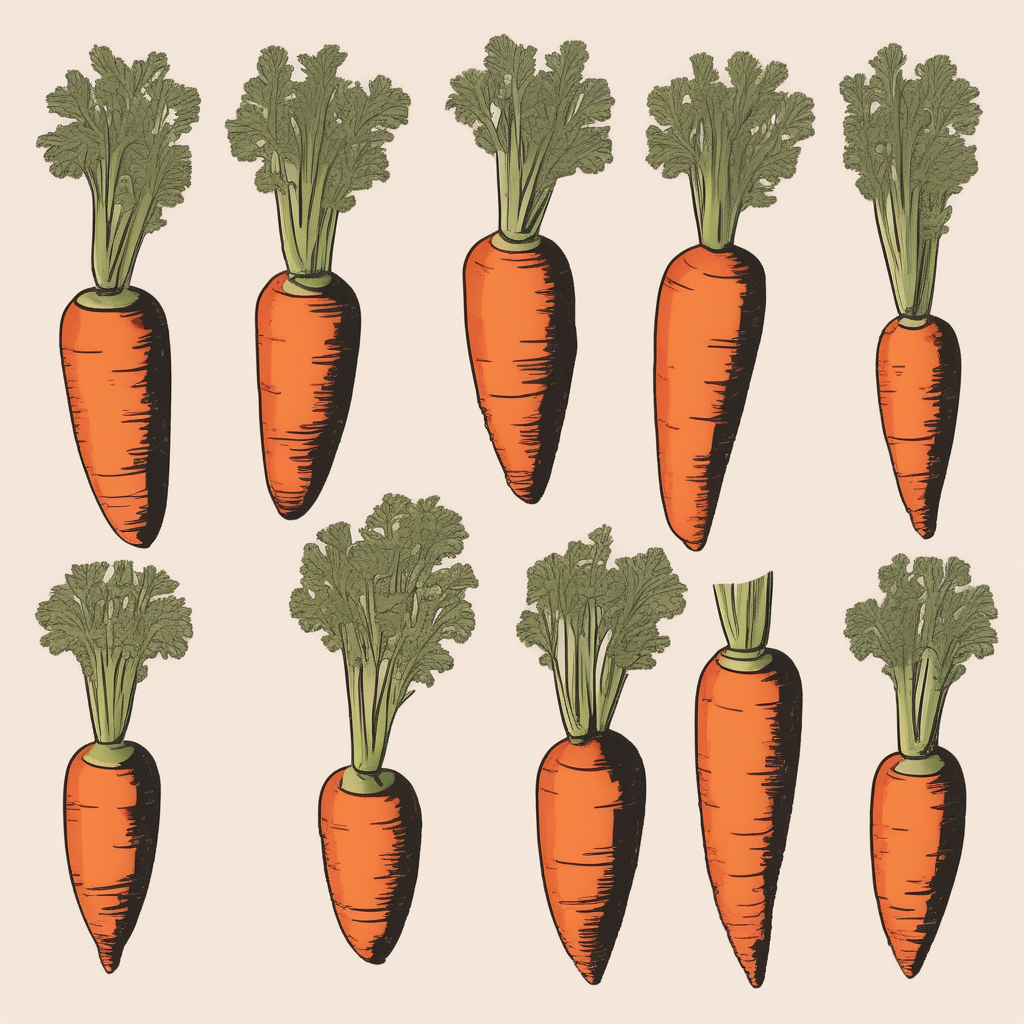}\\[-2pt]
        \includegraphics[width=.08\linewidth]{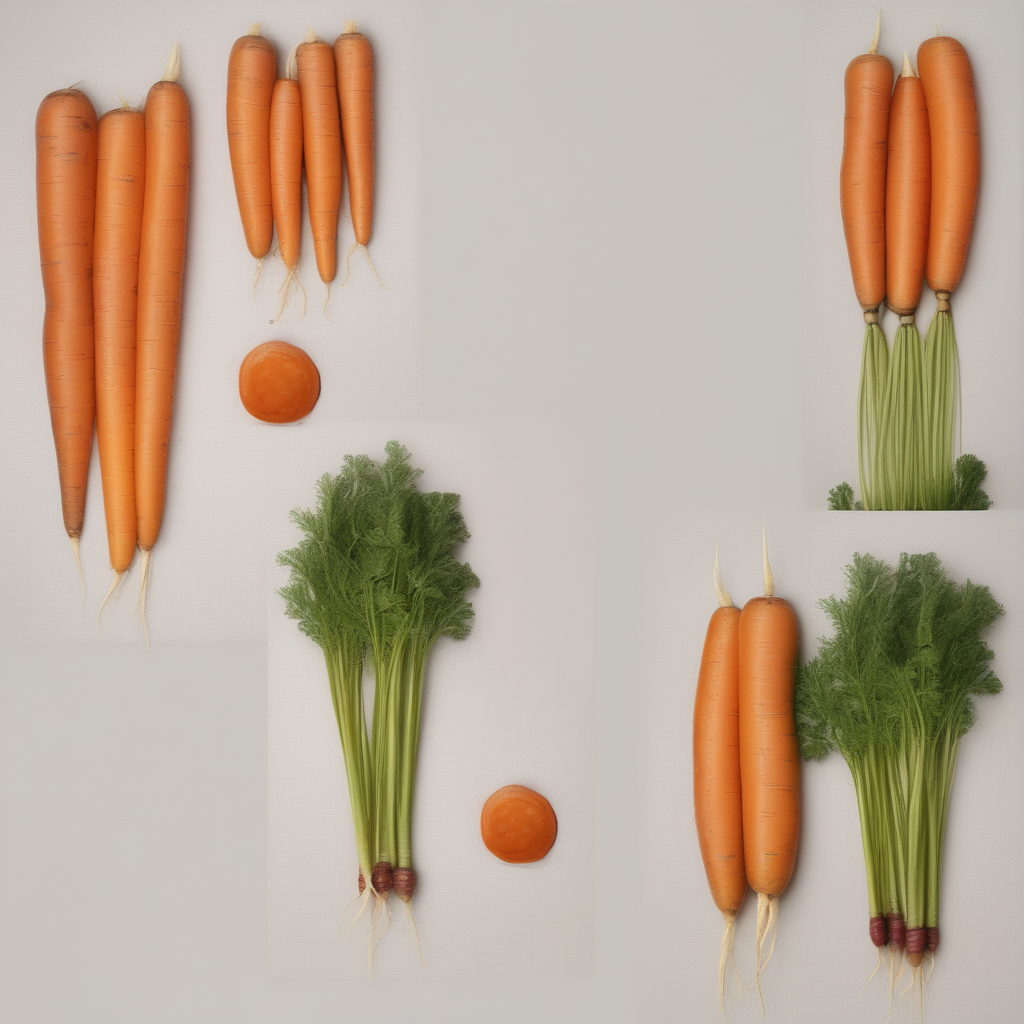}
      \end{tabular} \\

    \textbf{6 bottles} &
      \begin{tabular}{@{}c@{}}
        \includegraphics[width=.08\linewidth]{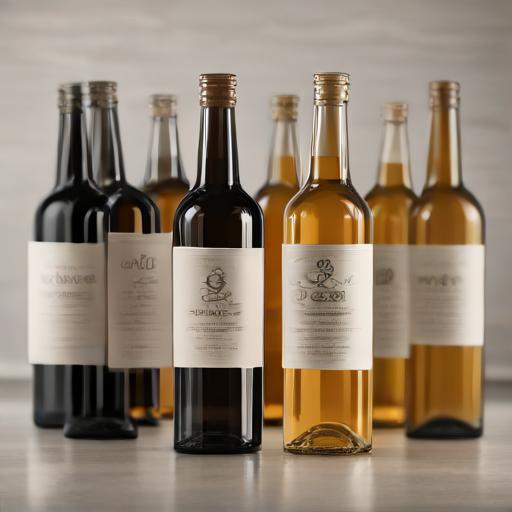}\\[-2pt]
        \includegraphics[width=.08\linewidth]{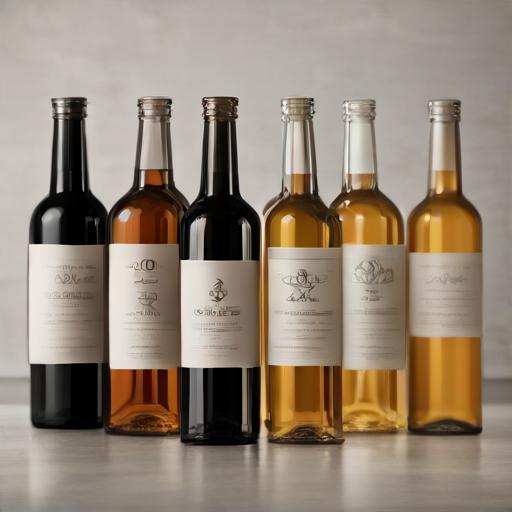}
      \end{tabular} &
      \begin{tabular}{@{}c@{}}
        \includegraphics[width=.08\linewidth]{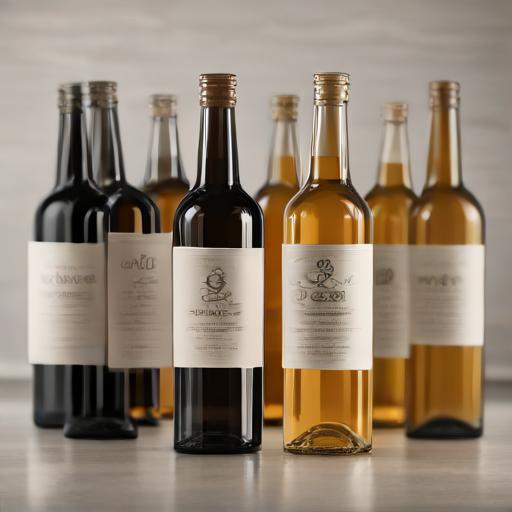}\\[-2pt]
        \includegraphics[width=.08\linewidth]{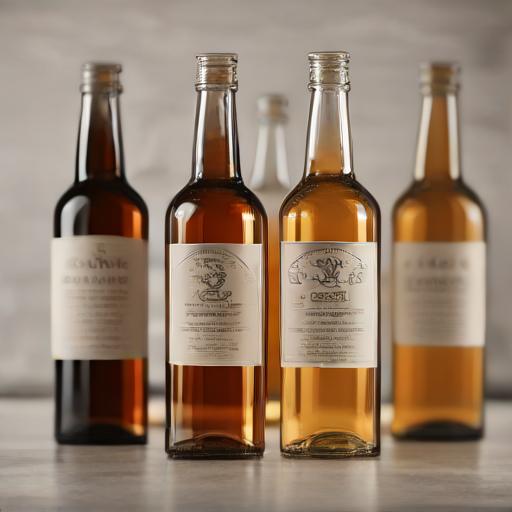}
      \end{tabular} &
      \begin{tabular}{@{}c@{}}
        \includegraphics[width=.08\linewidth]{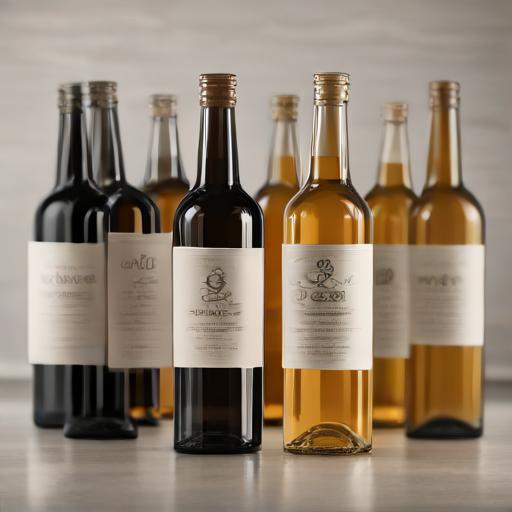}\\[-2pt]
        \includegraphics[width=.08\linewidth]{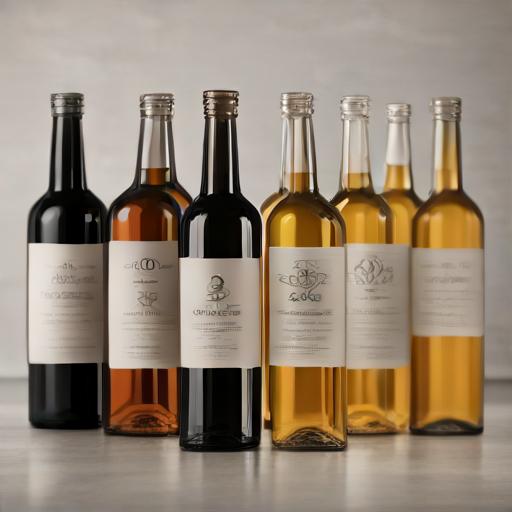}
      \end{tabular} &
      \begin{tabular}{@{}c@{}}
        \includegraphics[width=.08\linewidth]{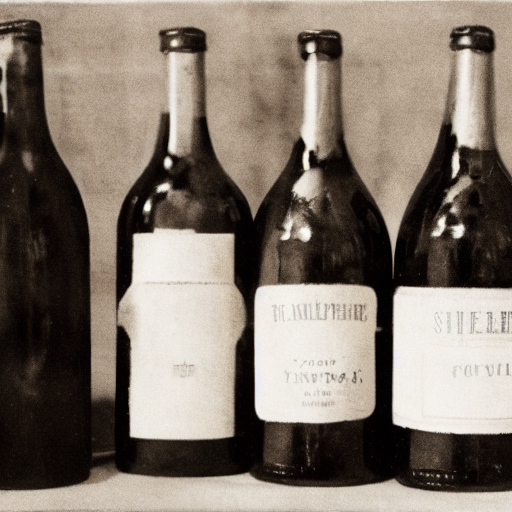}\\[-2pt]
        \includegraphics[width=.08\linewidth]{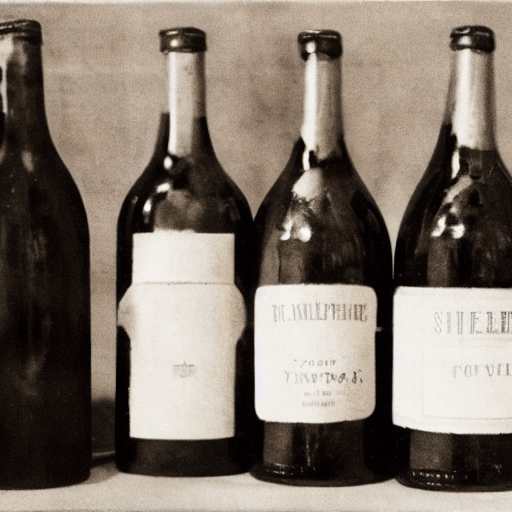}
      \end{tabular} &
      \begin{tabular}{@{}c@{}}
        \includegraphics[width=.08\linewidth]{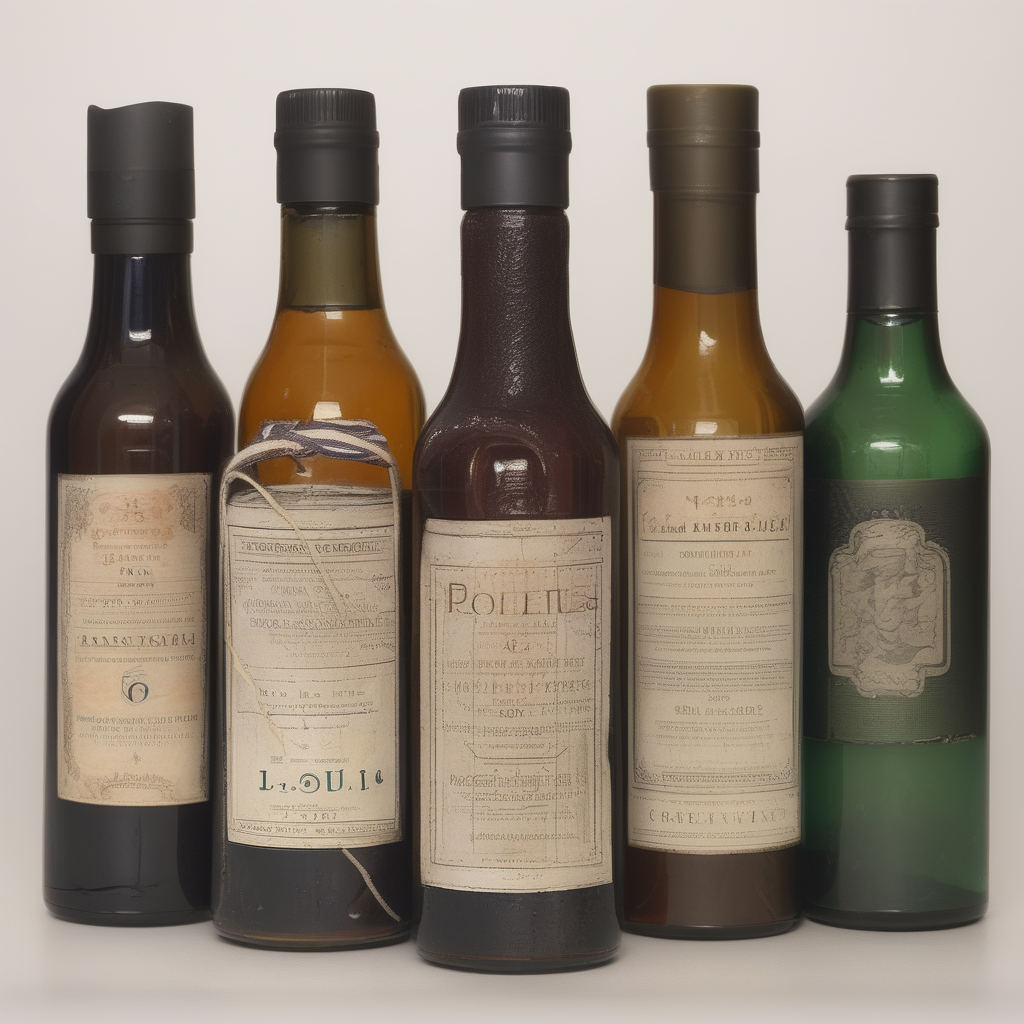}\\[-2pt]
        \includegraphics[width=.08\linewidth]{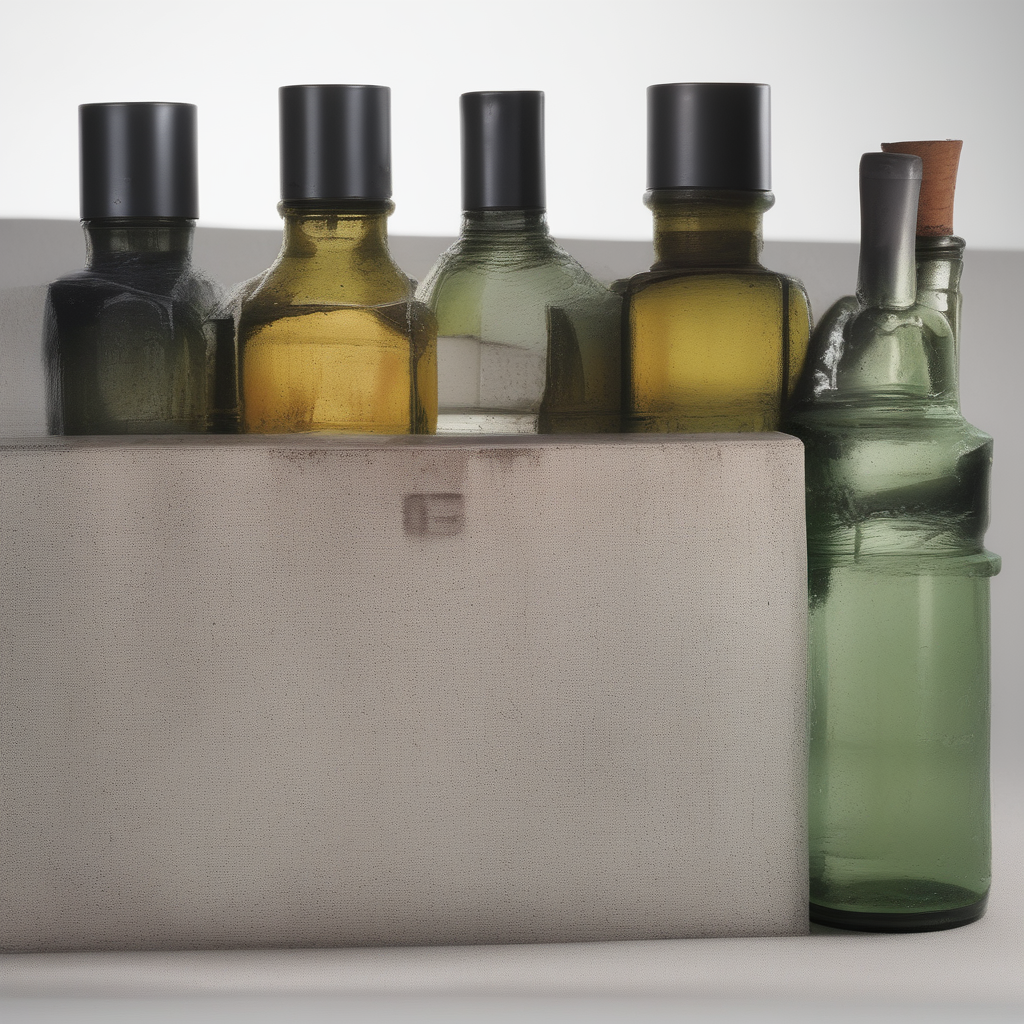}
      \end{tabular} \\

    \textbf{9 suitcases} &
      \begin{tabular}{@{}c@{}}
        \includegraphics[width=.08\linewidth]{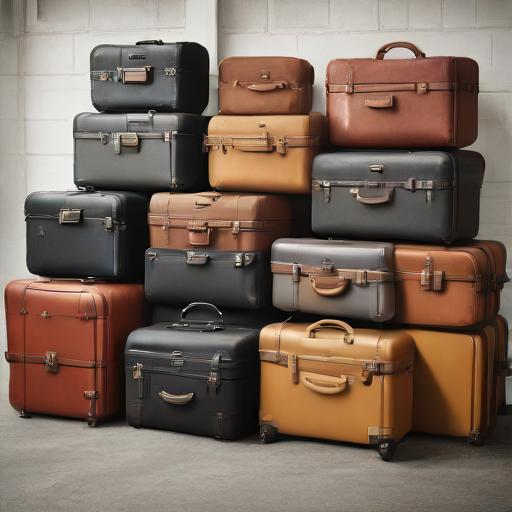}\\[-2pt]
        \includegraphics[width=.08\linewidth]{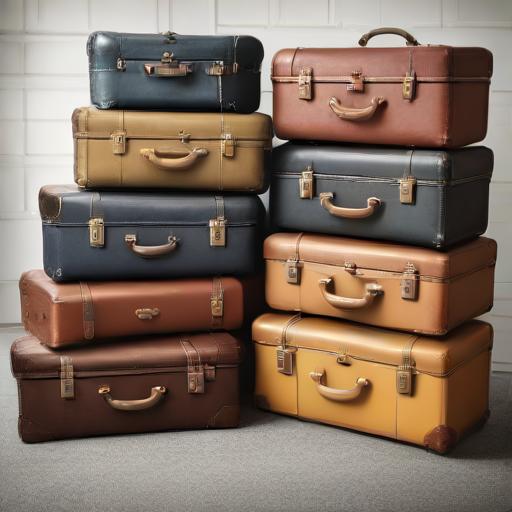}
      \end{tabular} &
      \begin{tabular}{@{}c@{}}
        \includegraphics[width=.08\linewidth]{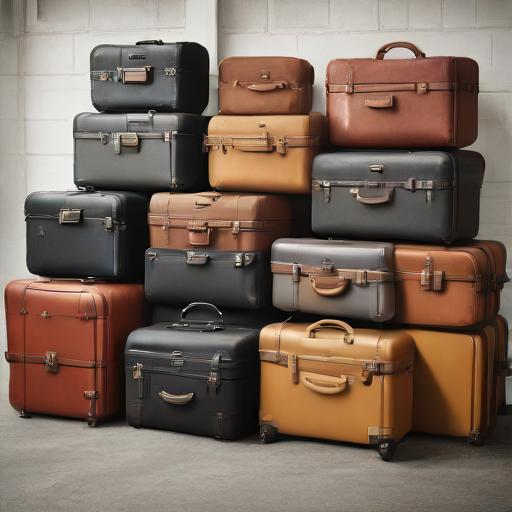}\\[-2pt]
        \includegraphics[width=.08\linewidth]{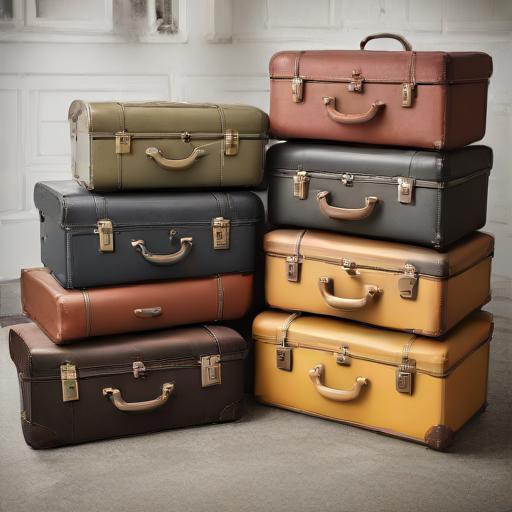}
      \end{tabular} &
      \begin{tabular}{@{}c@{}}
        \includegraphics[width=.08\linewidth]{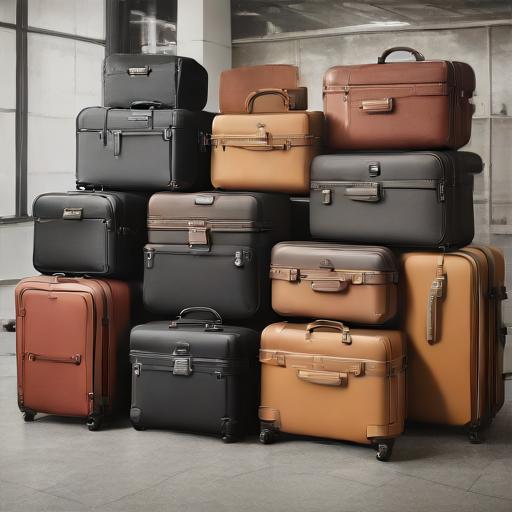}\\[-2pt]
        \includegraphics[width=.08\linewidth]{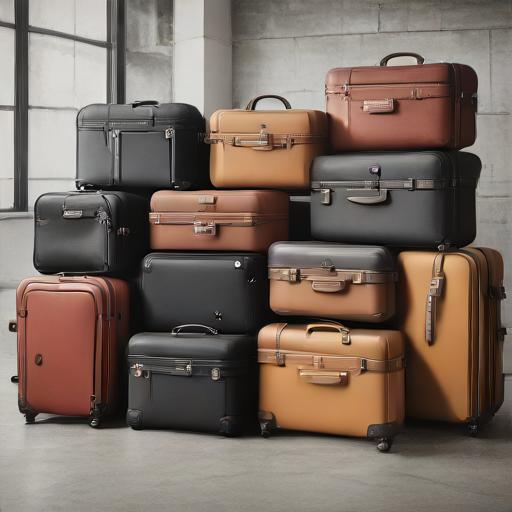}
      \end{tabular} &
      \begin{tabular}{@{}c@{}}
        \includegraphics[width=.08\linewidth]{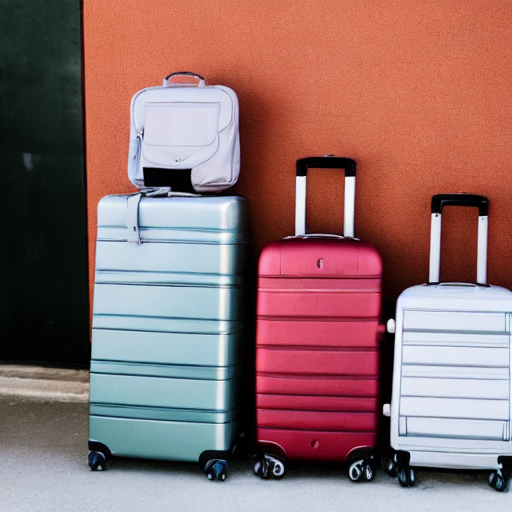}\\[-2pt]
        \includegraphics[width=.08\linewidth]{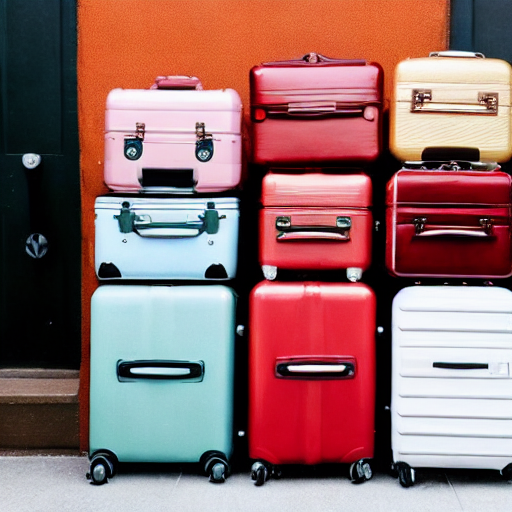}
      \end{tabular} &
      \begin{tabular}{@{}c@{}}
        \includegraphics[width=.08\linewidth]{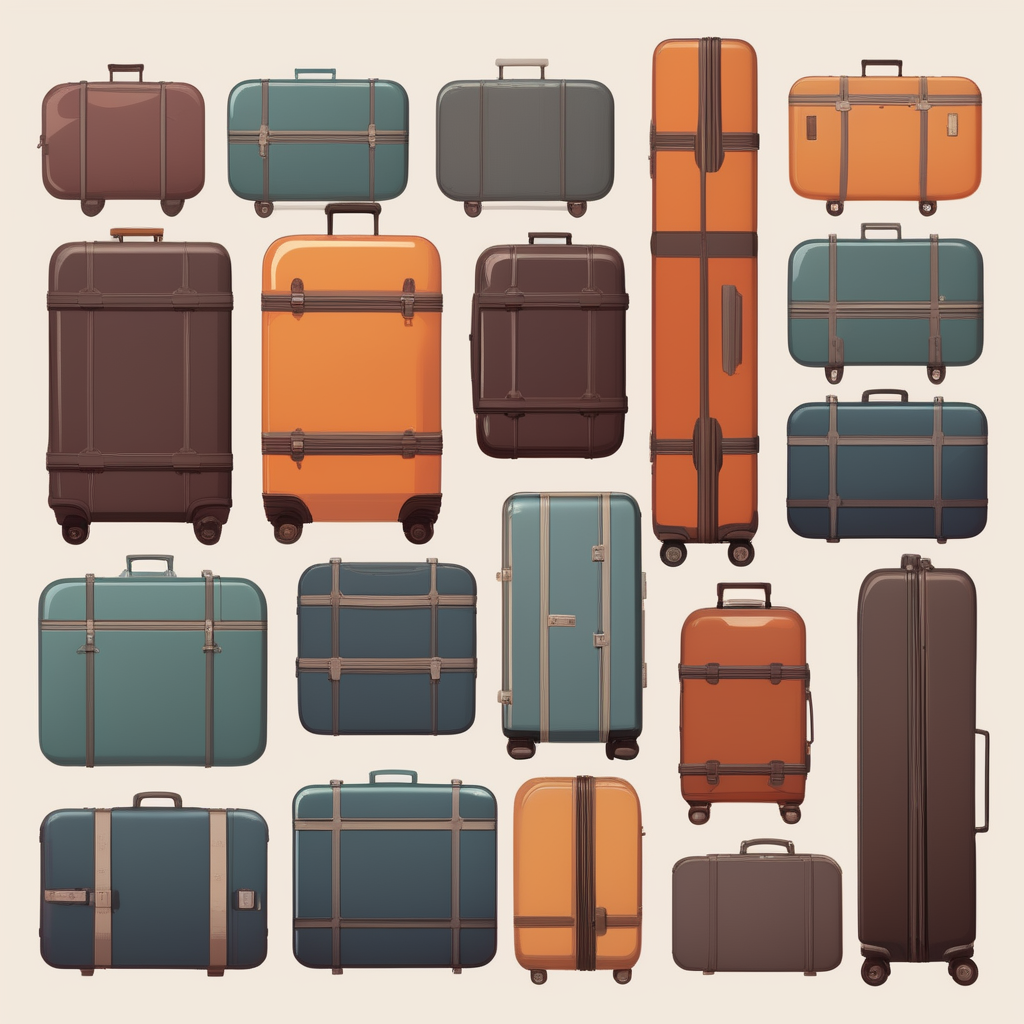}\\[-2pt]
        \includegraphics[width=.08\linewidth]{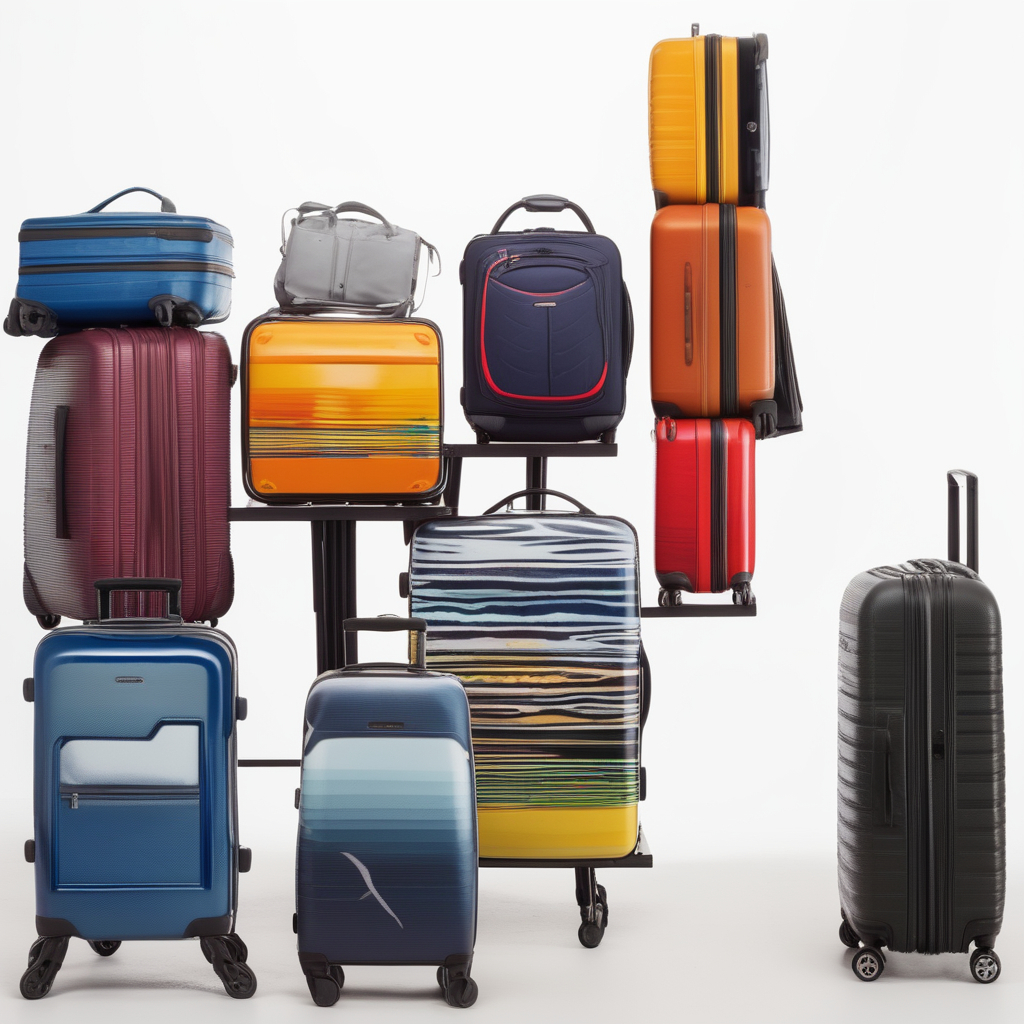}
      \end{tabular} \\

    \textbf{5 handbags} &
      \begin{tabular}{@{}c@{}}
        \includegraphics[width=.08\linewidth]{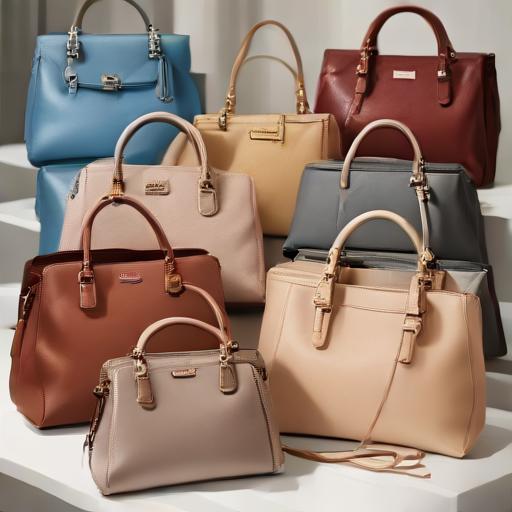}\\[-2pt]
        \includegraphics[width=.08\linewidth]{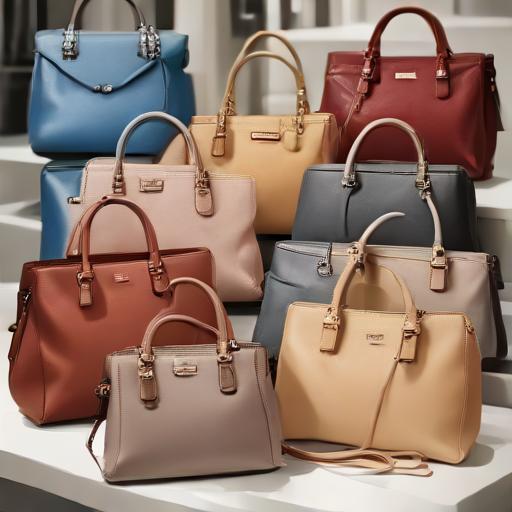}
      \end{tabular} &
      \begin{tabular}{@{}c@{}}
        \includegraphics[width=.08\linewidth]{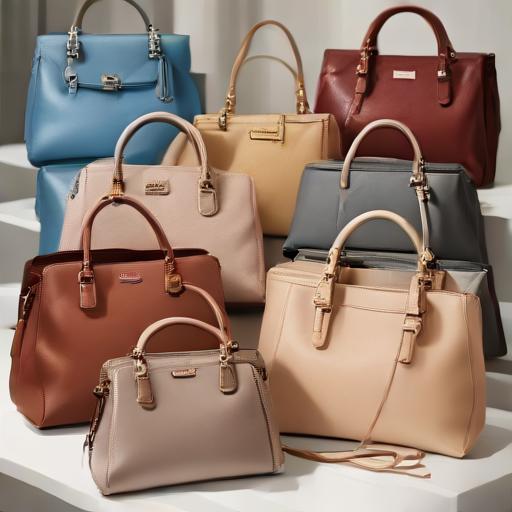}\\[-2pt]
        \includegraphics[width=.08\linewidth]{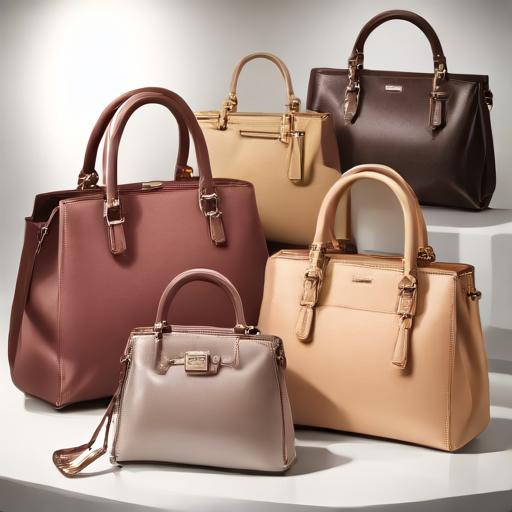}
      \end{tabular} &
      \begin{tabular}{@{}c@{}}
        \includegraphics[width=.08\linewidth]{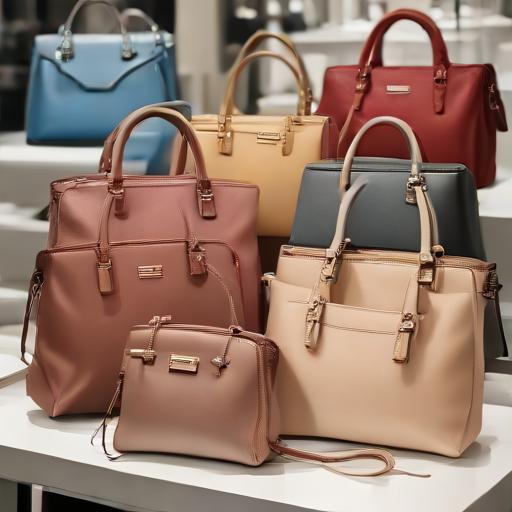}\\[-2pt]
        \includegraphics[width=.08\linewidth]{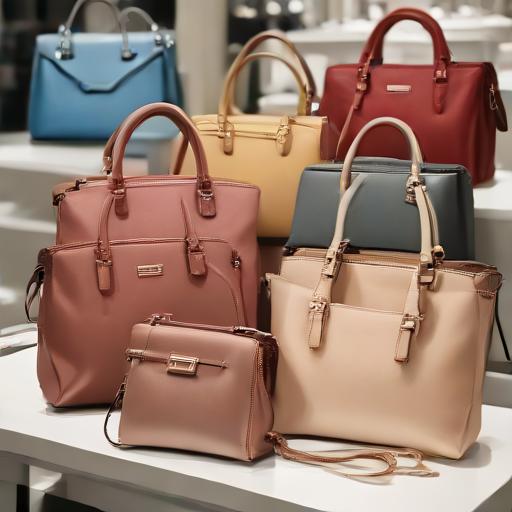}
      \end{tabular} &
      \begin{tabular}{@{}c@{}}
        \includegraphics[width=.08\linewidth]{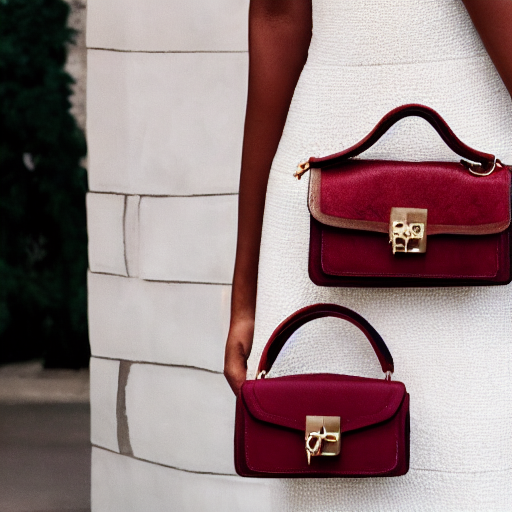}\\[-2pt]
        \includegraphics[width=.08\linewidth]{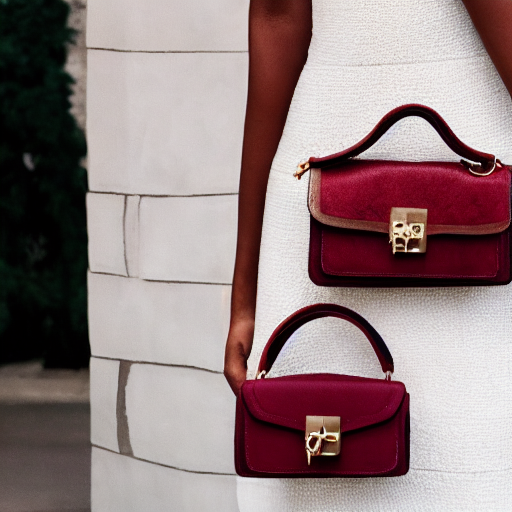}
      \end{tabular} &
      \begin{tabular}{@{}c@{}}
        \includegraphics[width=.08\linewidth]{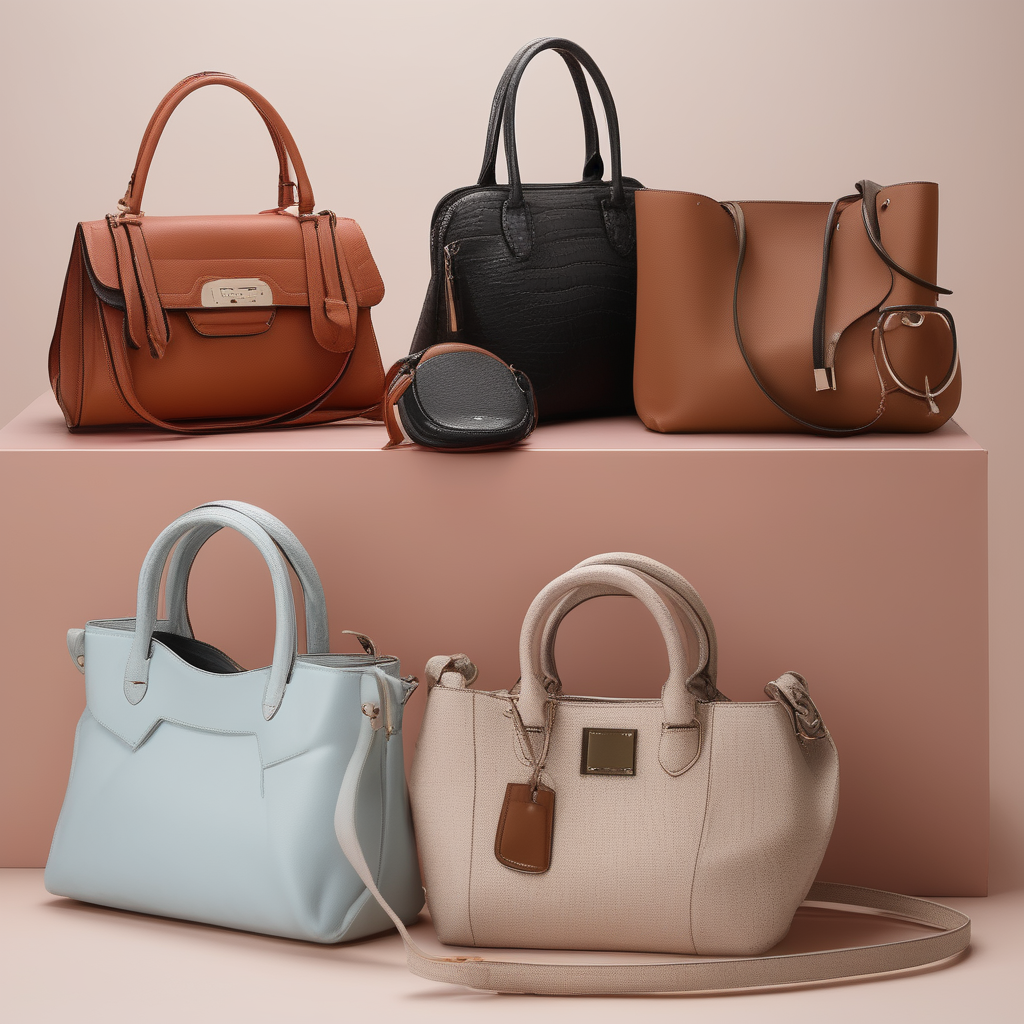}\\[-2pt]
        \includegraphics[width=.08\linewidth]{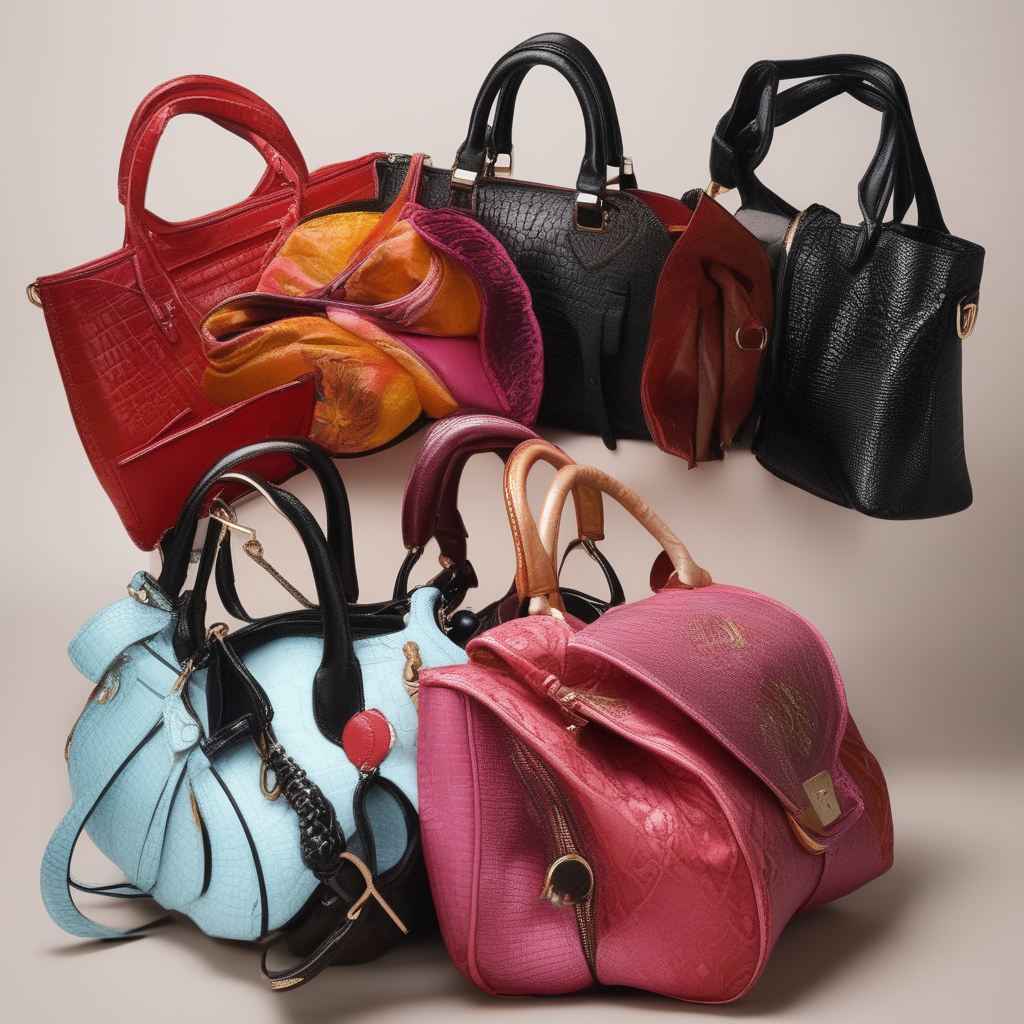}
      \end{tabular} \\
  \end{tabular}}

  \caption{Comparison to baselines of our method with different scaling variants, as well as Counting-Guidance and CountGen. For each example, we show the original image generated by the backbone model (top) and the optimized version with corrected object count (bottom). Prompts used for generation are shown on the left.}
  \label{fig:examples_vs_basline}
\end{figure*}

\begin{figure*}[t!]
    \centering
    \begin{subfigure}{0.48\textwidth}
        \includegraphics[width=\textwidth]{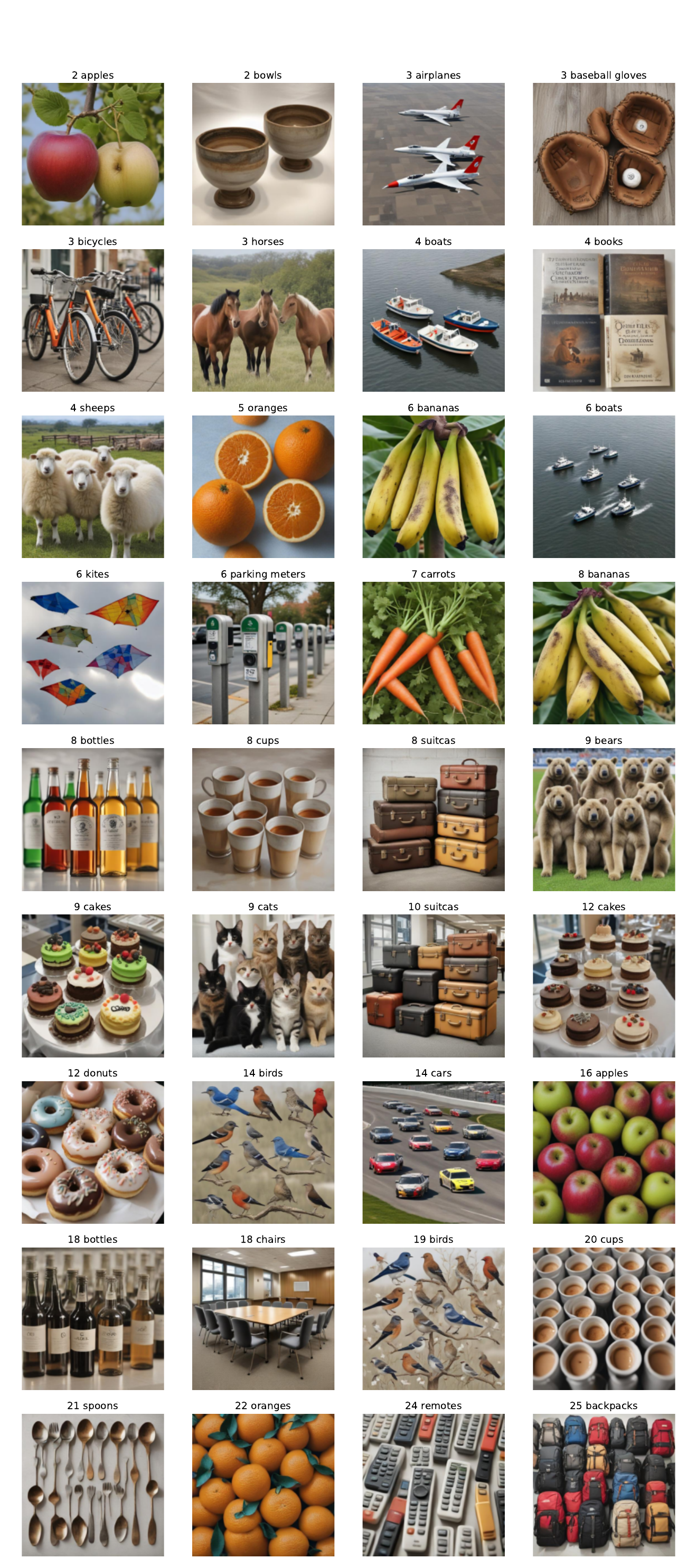}
        \caption{Ours}
    \end{subfigure}
    \hfill
    \begin{subfigure}{0.48\textwidth}
        \includegraphics[width=\textwidth]{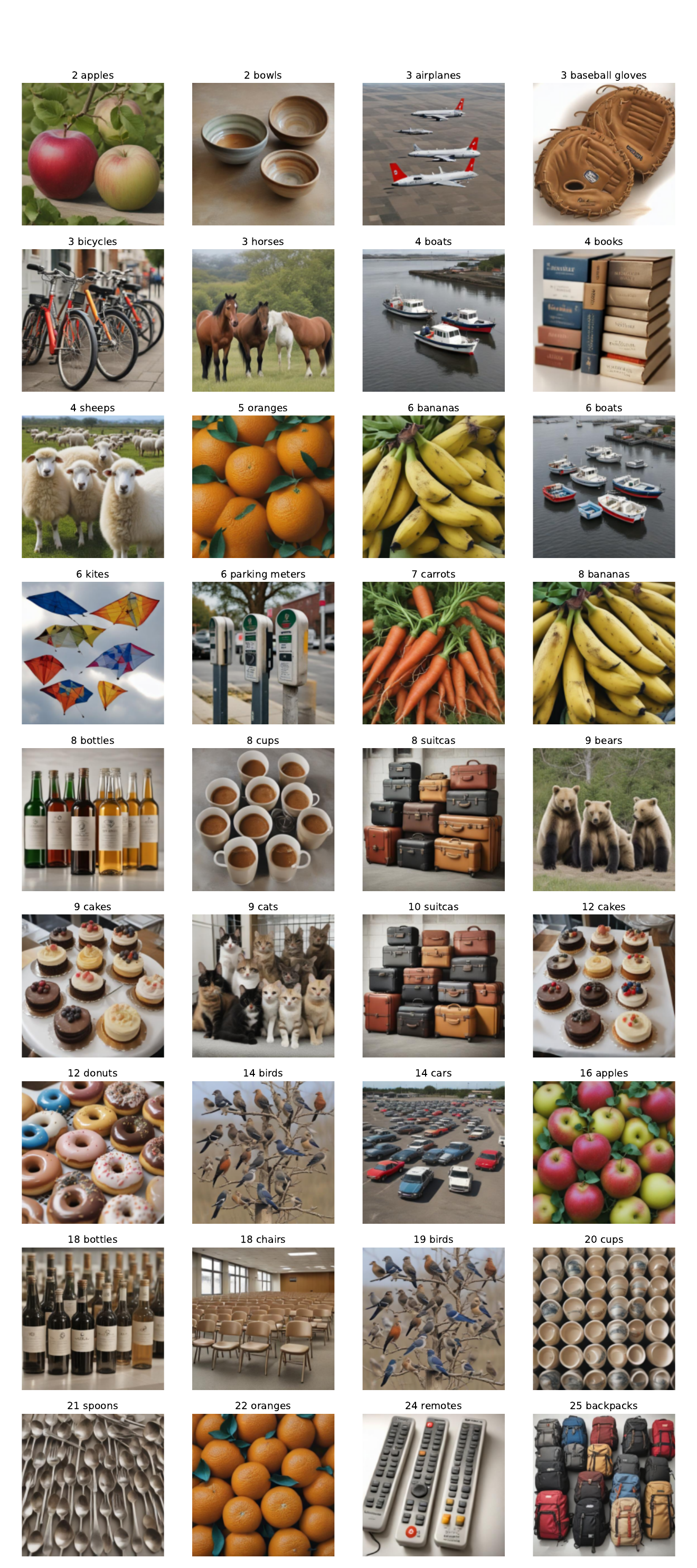}
        \caption{SD}
    \end{subfigure}
    \hfill
   \caption{We show additional results of our method using DETR-based scaling, compared to the backbone. Samples shown in increasing order. In many cases, SD-Turbo depicts partial objects; for example, multiple heads or broken chairs. Our method often corrects these issues, enhancing naturalism and counting accuracy. We find that for large quantities, the number of objects often does not meaningfully differ; for instance, outputs for 14 and 19 birds are almost identical.}
  \label{fig:comparison}
\end{figure*}

\clearpage

In this supplementary material, we present additional experimental results. The supplementary comprises the following subsections:

\begin{enumerate}
    \item Sec.~\ref{sec:experiment}, experimental setup
    \item Sec.~\ref{sec:collapse}, collapse of inference-time optimization 
    \item Sec.~\ref{detailed_evaluation_results}, detailed quantitative analysis
    \item Sec.~\ref{clip_ablation}, CLIP loss coefficient ablation study
    \item Sec.~\ref{learning_rate_ablation}, learning rate ablation study
    \item Sec.~\ref{threshold_ablation}, YOLO confidence threshold ablation study
\end{enumerate}

\section{Experimental setup} 
\label{sec:experiment}
We employed a single NVIDIA GPU L40. Training each token takes about one minute for approximately 30 iterations using SDXL-Turbo~\cite {sauer2023adversarial}.  The optimized counting token can be reused without extra inference time. We set $\lambda=5$,  the learning rate to 0.05 for the optimization. the \clipcount~scaling hyperparameter  $\lambda_{\text{scale}}$ is set to 60 for the static scale. 

\section{Stability and Failure Modes of Inference-Time Optimization}
\label{sec:collapse}

Inference-time optimization can suffer from training collapse, where the image either fails to change or becomes distorted. We observed these issues across the baselines. These issues may arise because their method does not incorporate any penalty, such as our CLIP penalty, to ensure semantic accuracy.

\begin{figure}[t]
  \centering
  \setlength{\tabcolsep}{0pt}
  \resizebox{0.9\linewidth}{!}{%
    \begin{tabular}{c@{}ccc}
      & \scriptsize \textbf{25 apples} & \scriptsize \textbf{21 boats} & \scriptsize \textbf{17 chairs} \\
      \hspace{-0.5cm} \rotatebox{90}{\scriptsize \textbf{SDv1.4}} &
      \includegraphics[width=0.15\linewidth]{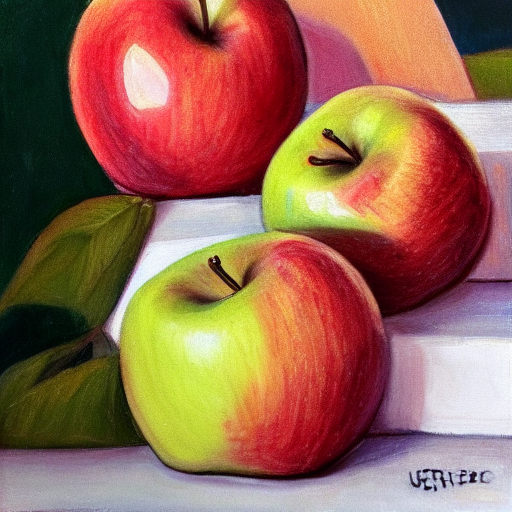} &
      \includegraphics[width=0.15\linewidth]{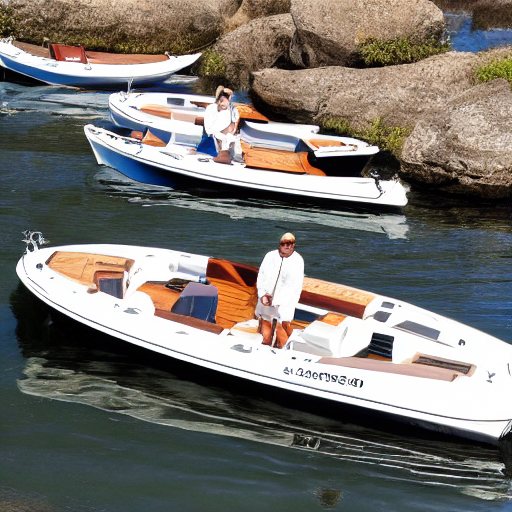} &
      \includegraphics[width=0.15\linewidth]{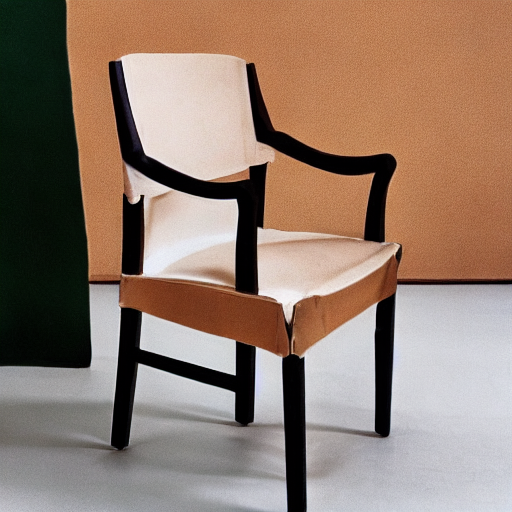} \\
      \hspace{-0.5cm} \rotatebox{90}{\scriptsize \textbf{Counting-Guidance}} &
      \includegraphics[width=0.15\linewidth]{figures/Counting_Guidance/fails/apple_25/Counting-Guidance_counting_best_apple_25_actual.png} &
      \includegraphics[width=0.15\linewidth]{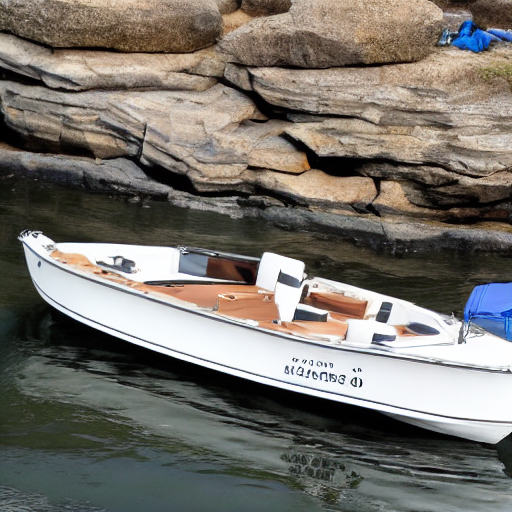} &
      \includegraphics[width=0.15\linewidth]{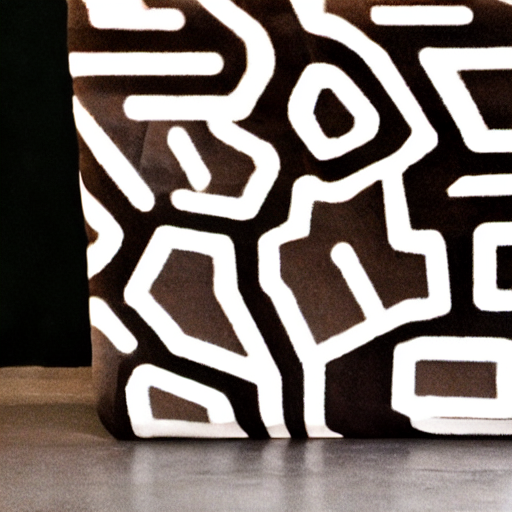}
    \end{tabular}
  }
  \caption{Failure cases of Counting-Guidance. The left and middle show little to no correction for large counts; the rightmost image collapses into a noisy pattern.}
  \label{fig:counting_guidance_fail_results}
\end{figure}

Counting-Guidance~\cite{binyamin2024make} often struggles with large object counts. As shown in Fig.~\ref{fig:counting_guidance_fail_results}, the method frequently applies minimal correction (left and middle) or collapses into noisy, indistinct patterns (right).

\begin{figure}[t]
  \centering
  \setlength{\tabcolsep}{0pt}
  \resizebox{0.9\linewidth}{!}{%
    \begin{tabular}{c@{}ccc}
      & \scriptsize \textbf{4 books} & \scriptsize \textbf{3 cups} & \scriptsize \textbf{3 elephants} \\
      \hspace{-0.5cm} \rotatebox{90}{\scriptsize \textbf{SDXL}} &
      \includegraphics[width=0.15\linewidth]{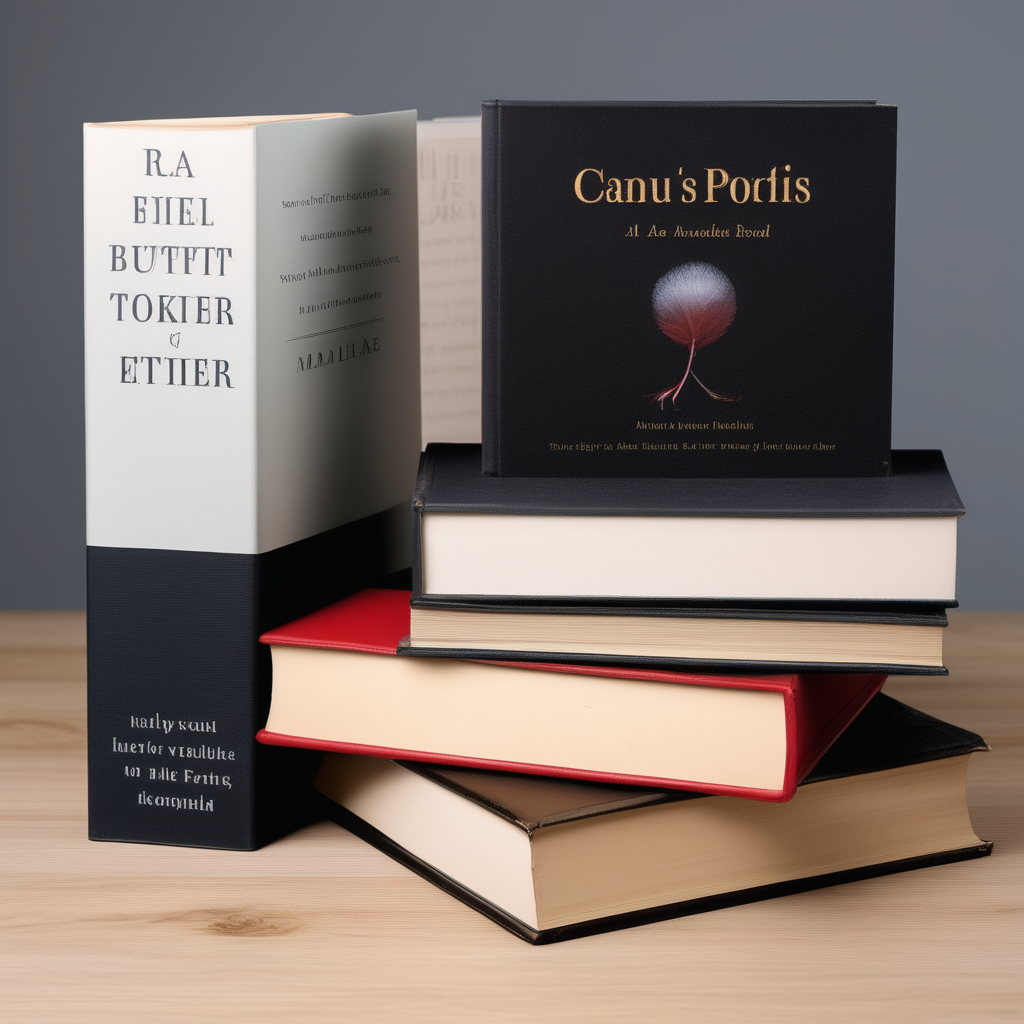} &
      \includegraphics[width=0.15\linewidth]{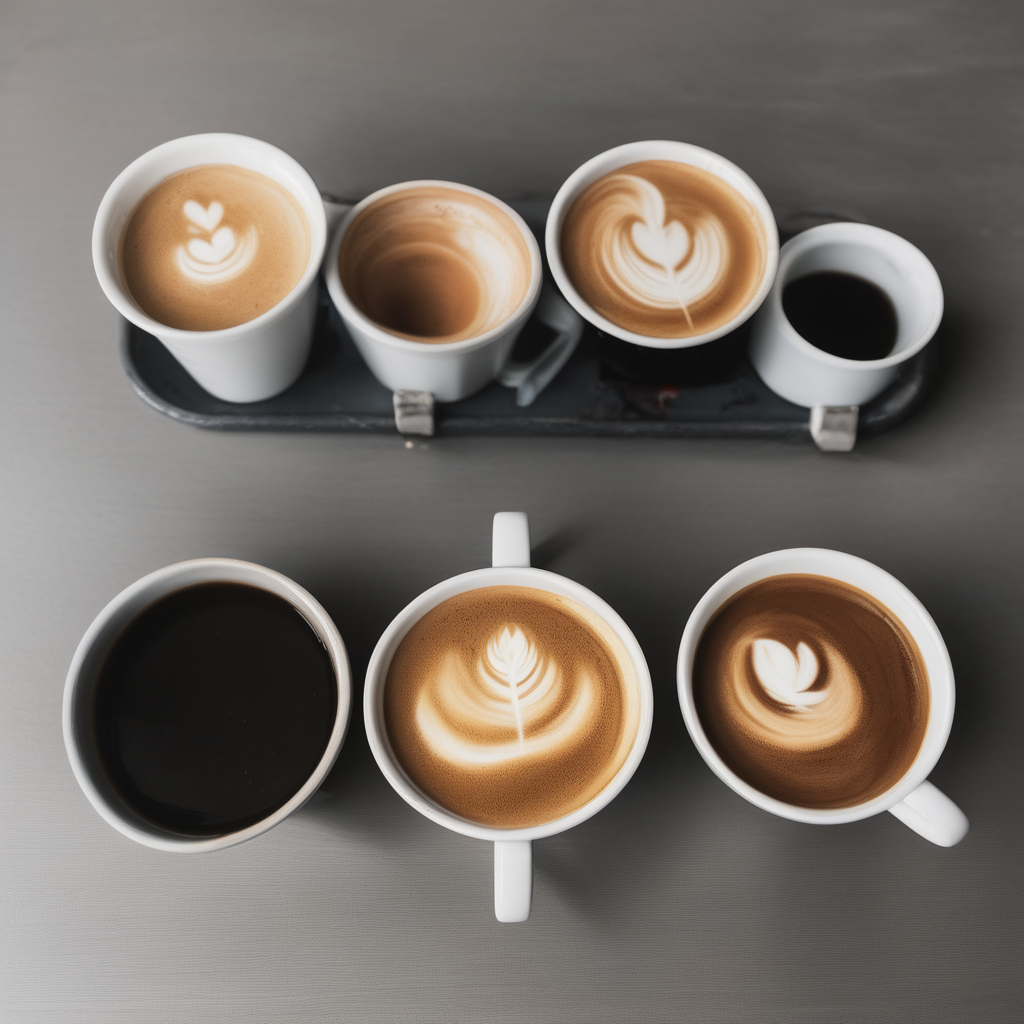} &
      \includegraphics[width=0.15\linewidth]{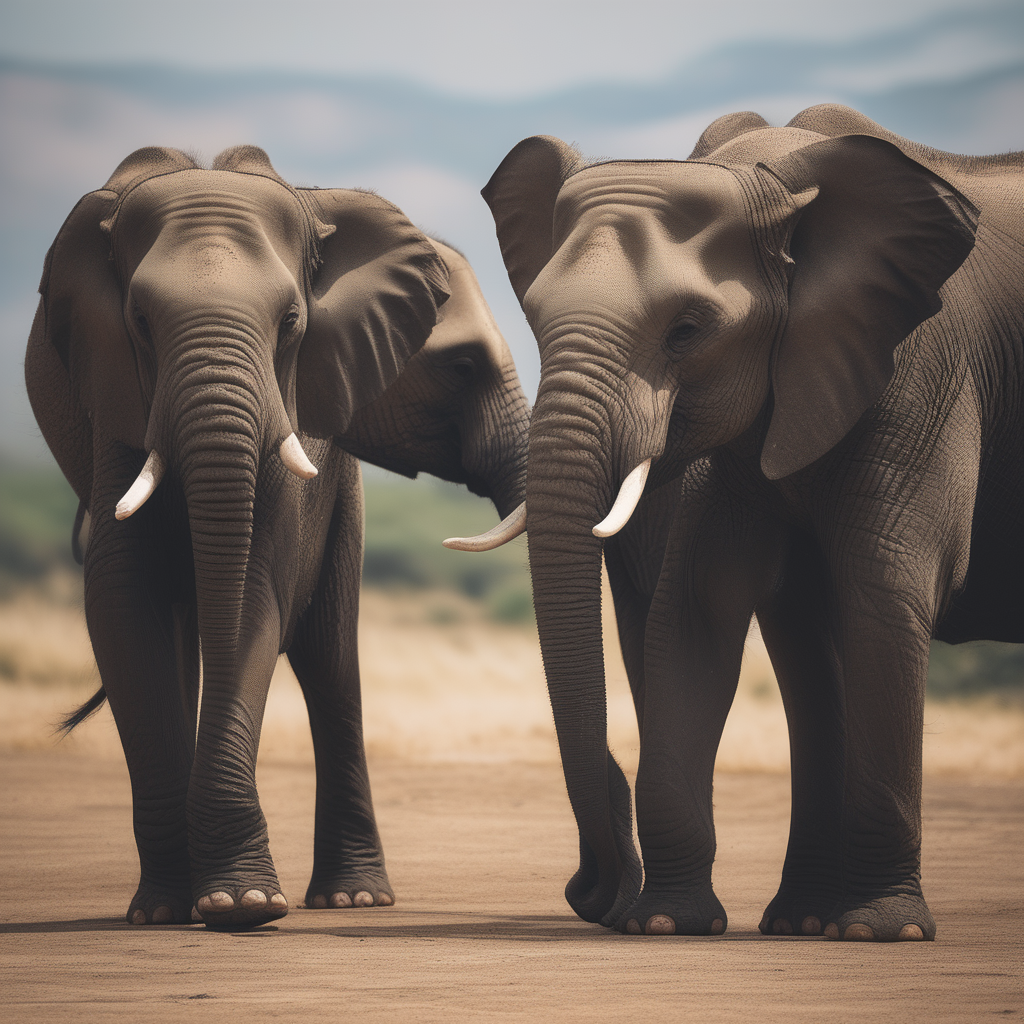} \\
      \hspace{-0.5cm} \rotatebox{90}{\scriptsize \textbf{CountGen}} &
      \includegraphics[width=0.15\linewidth]{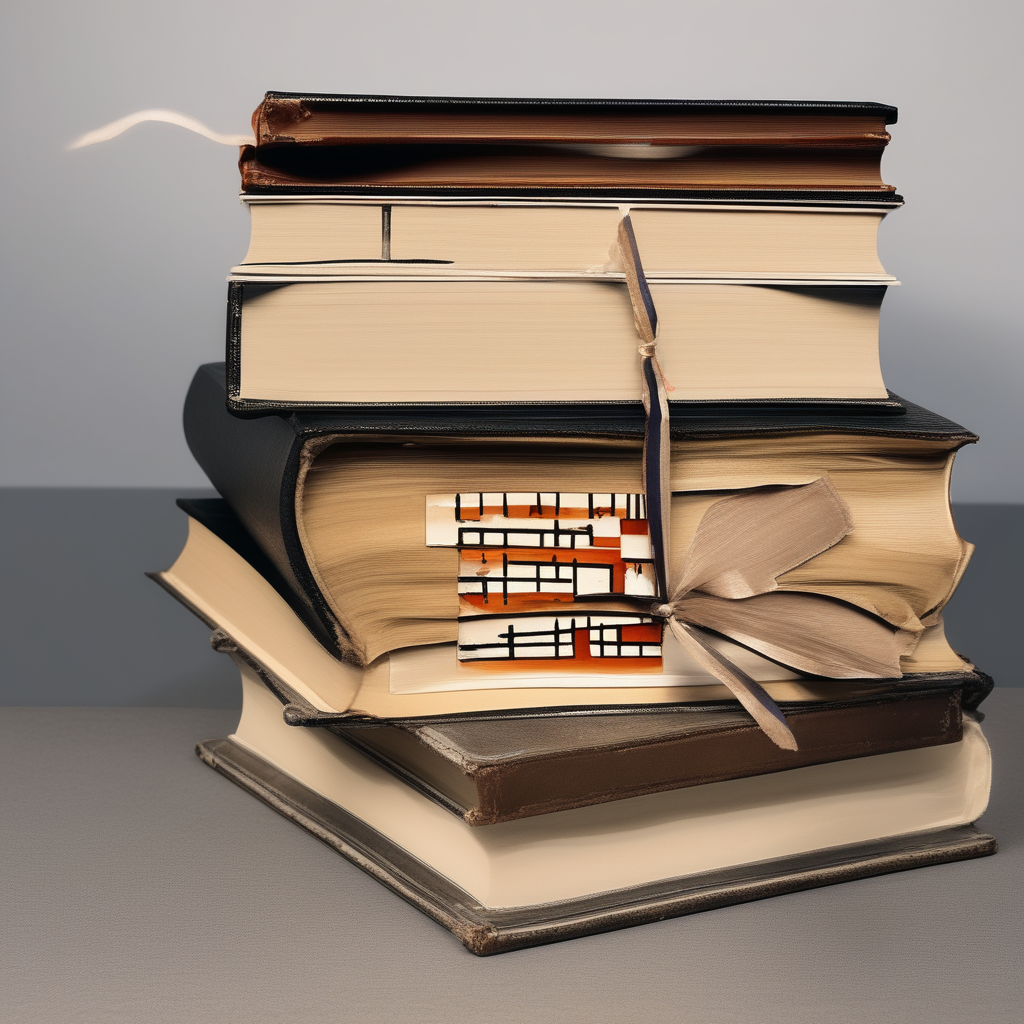} &
      \includegraphics[width=0.15\linewidth]{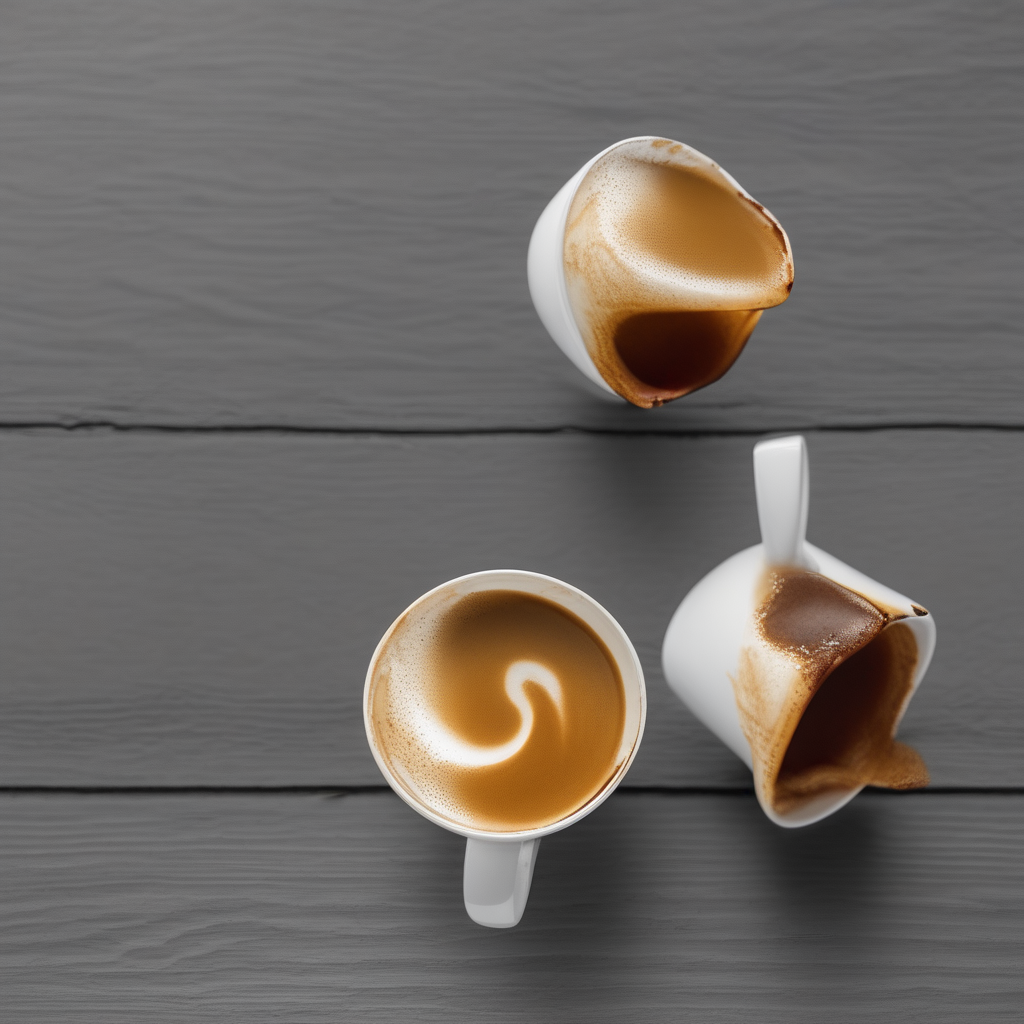} &
      \includegraphics[width=0.15\linewidth]{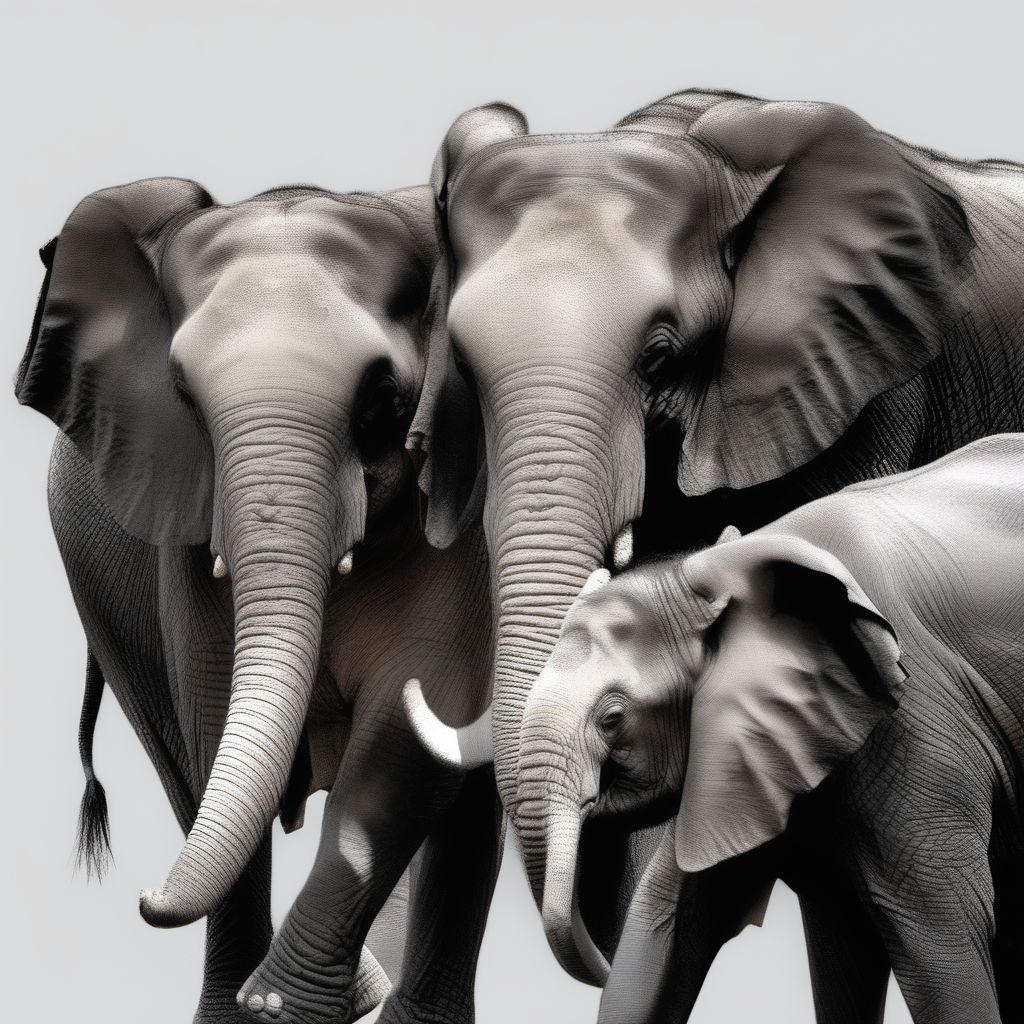}
    \end{tabular}
  }
  \caption{Failure cases of CountGen. Left: incorrect count. Middle: unrealistic appearance. Right: accurate count and realism, but altered scene.}
  \label{fig:make_it_count_results}
\end{figure}

CountGen~\cite{kang2025counting} exhibits a different failure mode (Fig.~\ref{fig:make_it_count_results}). It often enforces the correct count, but at the cost of visual realism. The middle image shows unrealistic object shapes, while the rightmost alters the scene semantics, despite counting success.

\begin{figure}[h]
  \centering
  \setlength{\tabcolsep}{0pt}
  \resizebox{0.8\linewidth}{!}{%
    \begin{tabular}{c@{}cc}
      \scriptsize \textbf{3 frogs} & \scriptsize \textbf{15 oranges} \\
      \includegraphics[width=0.15\linewidth]{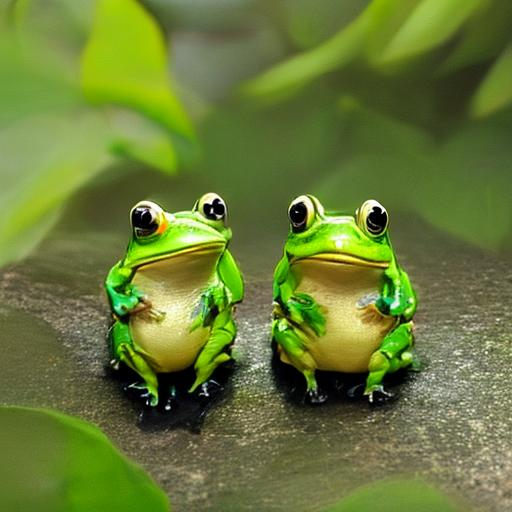} &
      \includegraphics[width=0.15\linewidth]{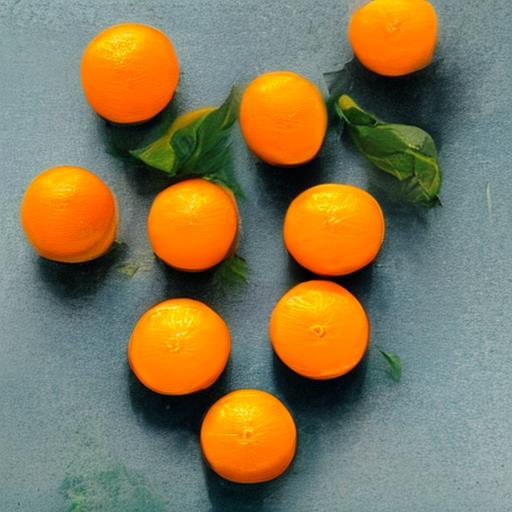} \\
      \includegraphics[width=0.15\linewidth]{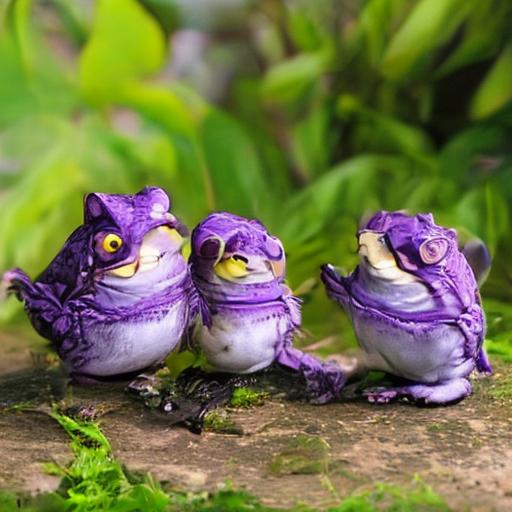} &
      \includegraphics[width=0.15\linewidth]{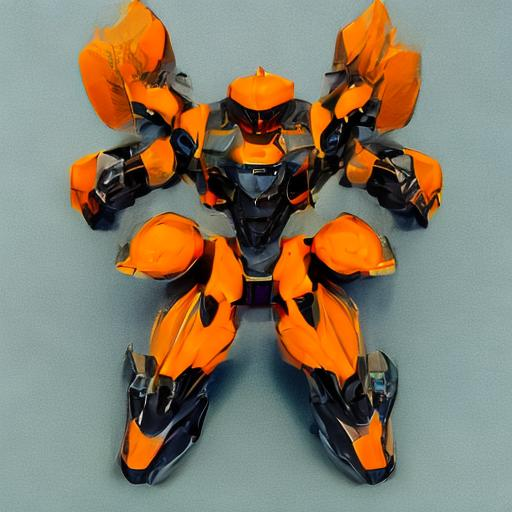}
    \end{tabular}
  }
  \caption{Collapse in early experiments with InstaFlow~\cite{liu2023instaflow} without CLIP regularization. Top: original generation. Bottom: optimized but adversarial.}
  \label{fig:failure}
\end{figure}

We encountered similar collapses in our early experiments using \textbf{InstaFlow}~\cite{liu2023instaflow}. As Fig.~\ref{fig:failure} shows, removing the CLIP alignment penalty resulted in adversarial changes—e.g., color distortions or content shifts—despite correct counts. Adding a CLIP penalty preserved prompt alignment. Notably, such failures did not occur with \textbf{SD-Turbo,} which remained stable even without CLIP guidance.

\section{Detailed quantitative analysis}
\label{detailed_evaluation_results}

\begin{figure*}[t!]
    \centering
    \begin{subfigure}{0.49\textwidth}
        \includegraphics[width=\textwidth]{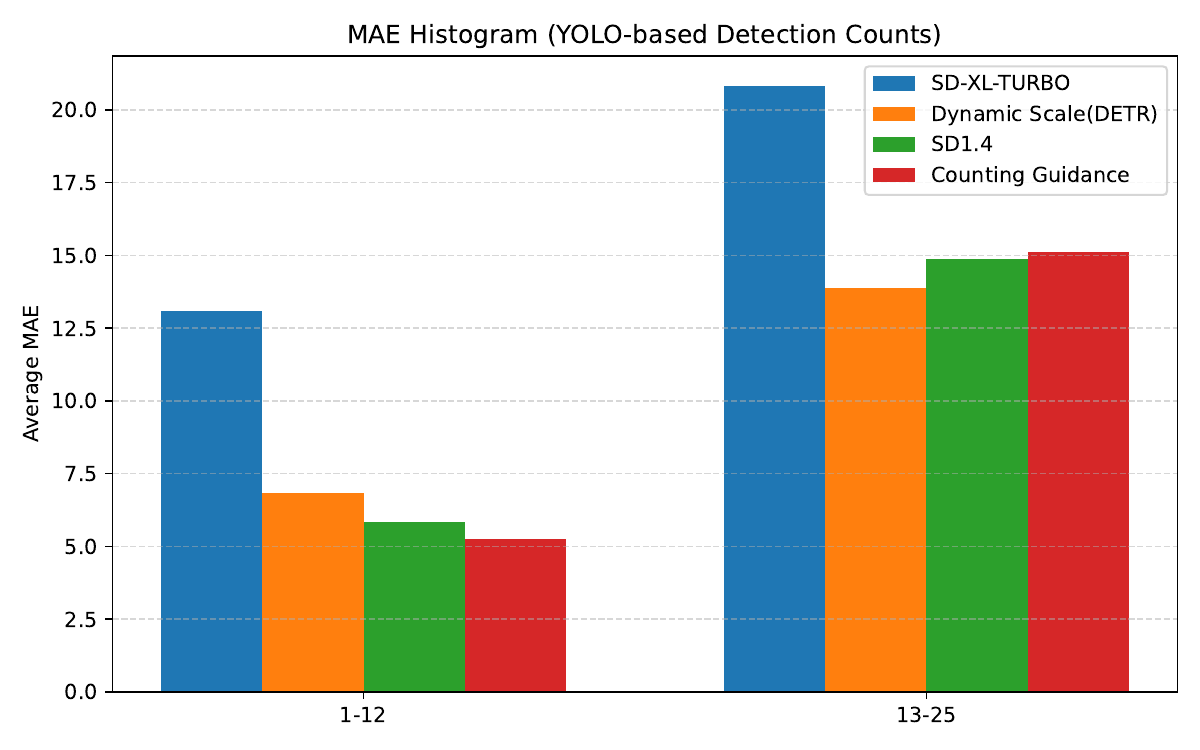}
    \end{subfigure}
    \hfill
    \begin{subfigure}{0.49\textwidth}
        \includegraphics[width=\textwidth] {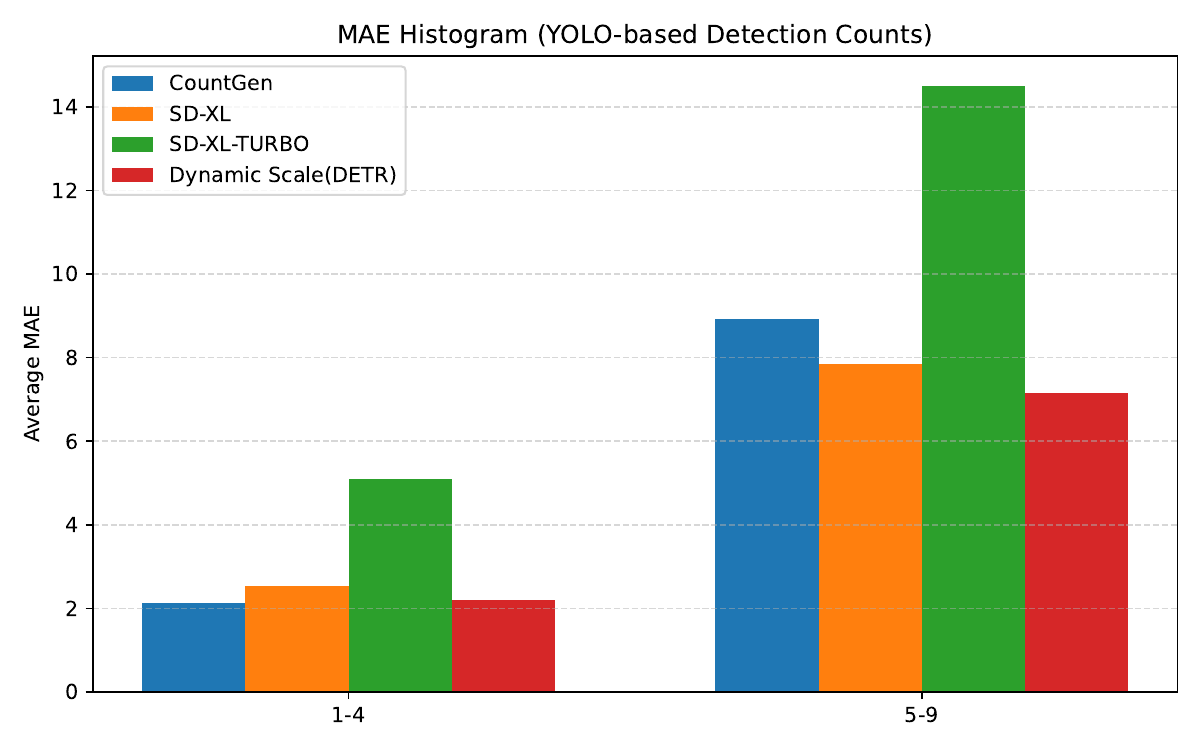}
    \end{subfigure}\vspace{-15pt}
    \hfill
    \caption{Comparison of different methods on sets of varying quantities. Since CountGen supports only up to nine objects, we limit the evaluation to 1-4 for small quantities and 5-9 for large quantities. We use YOLO to count the number of objects and report the mean absolute error (MAE).}
    \Description{Comparison with baselines}
  \label{fig:baseline_histogram}
\end{figure*}

In Fig.~\ref{fig:baseline_histogram}, we show the YOLO-based counting error for our method compared to the baselines. Although the comparison is not entirely fair since each baseline uses a different backbone, our method achieves the highest accuracy for large object counts despite relying on SD Turbo, the weakest backbone. For smaller quantities, our performance is comparable to other methods, even though they benefit from stronger backbones.

\begin{figure*}[t!]
    \centering
    \includegraphics[width=0.7\textwidth]{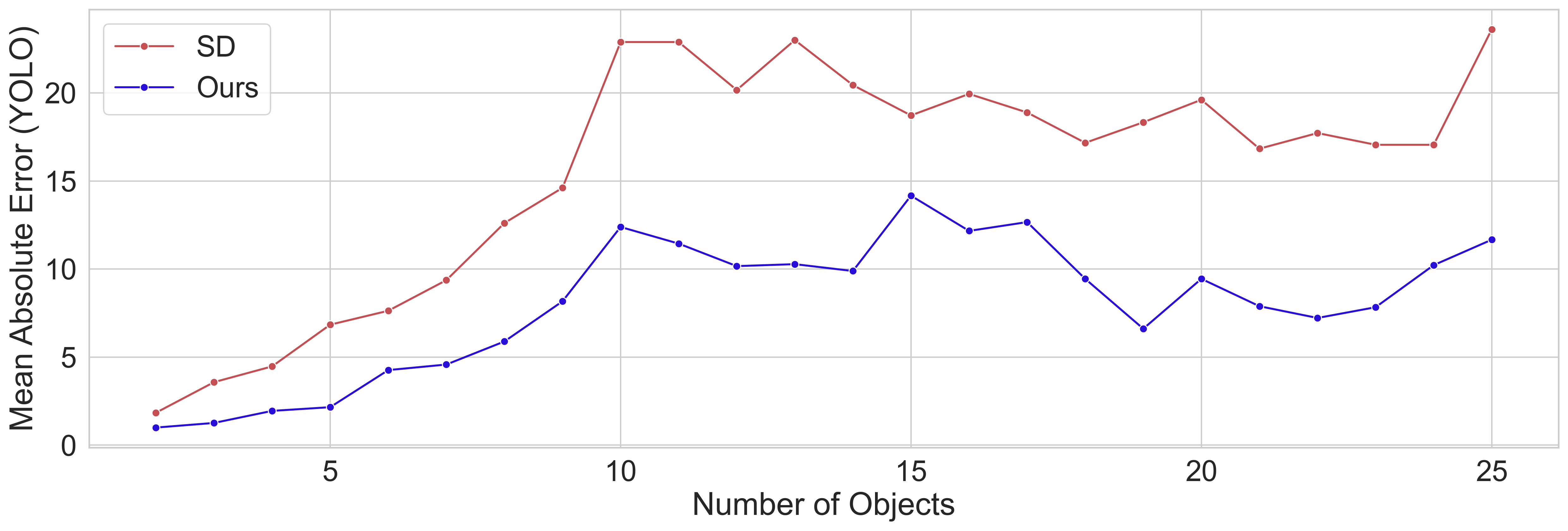}
    \caption{MAE scores as a function of the number of objects ranging from 2 to 25.}
    \label{fig:mae_per_amount}
\end{figure*}

In Fig.~\ref{fig:mae_per_amount}, we show detailed scores of our benchmark for each number of objects from two to 25. Our method consistently outperforms others in accuracy. The error grows linearly as the number of objects increases from two to 10. Interestingly, the error remains relatively constant beyond ten objects. This could be due to the challenges the YOLO-based counting method faces with larger numbers, affecting the metric’s accuracy.

\begin{figure*}[t!]
    \centering
    \includegraphics[width=0.7\textwidth]{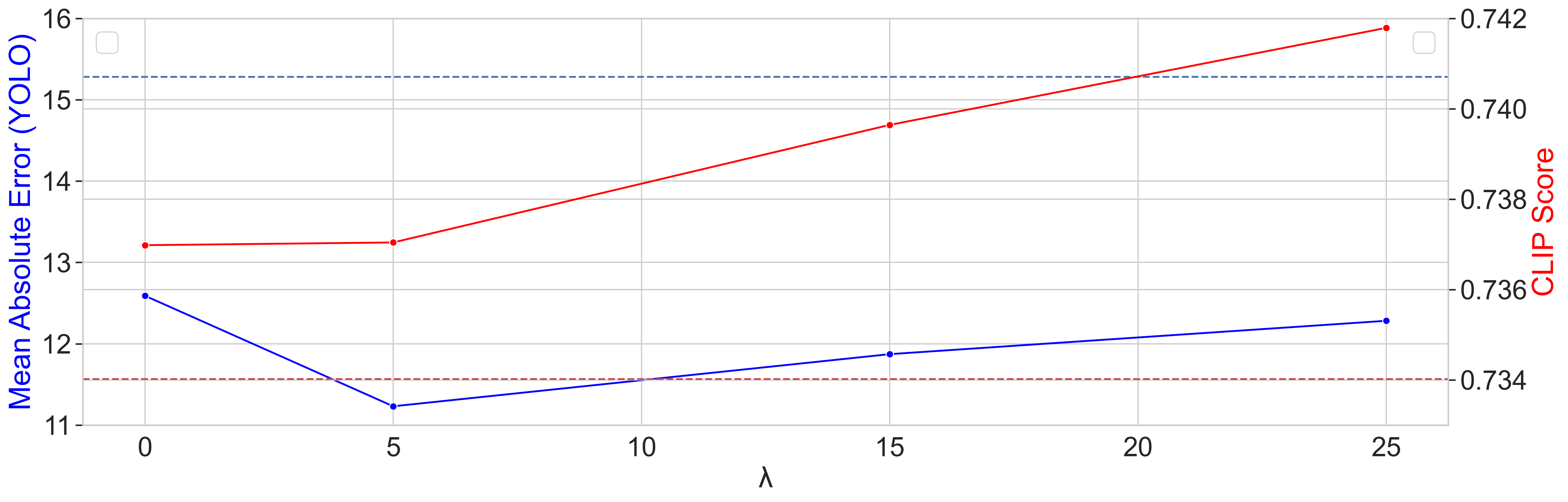}
    \caption{MAE (blue) and CLIP (red) scores as a function of $\lambda$. We show the SD-Turbo result baseline in a dotted line.}
    \label{fig:lambda_ablation}
\end{figure*}

\begin{figure*}[t!]
    \centering
    \includegraphics[width=0.7\textwidth]{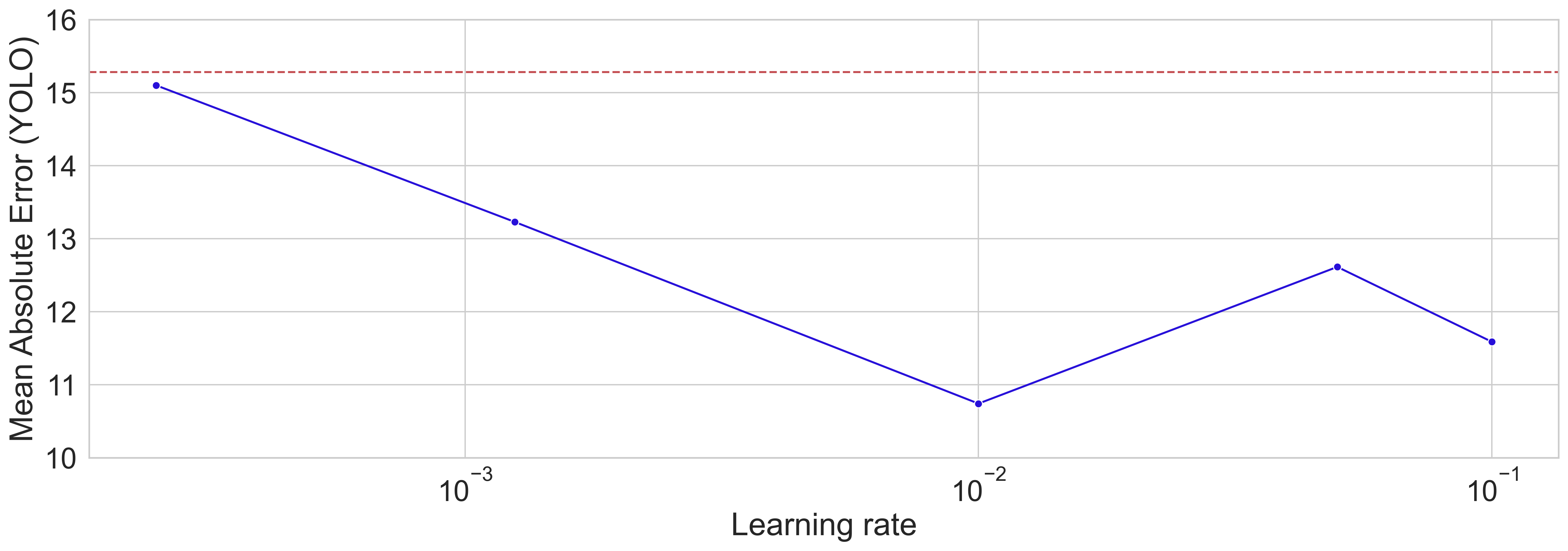}
    \caption{MAE score as a function of the learning rate. The MAE of SD-Turbo is displayed in red.}
    \label{fig:lr_ablation}
\end{figure*}

\begin{figure*}[t!]
    \centering
    \includegraphics[width=0.7\textwidth]{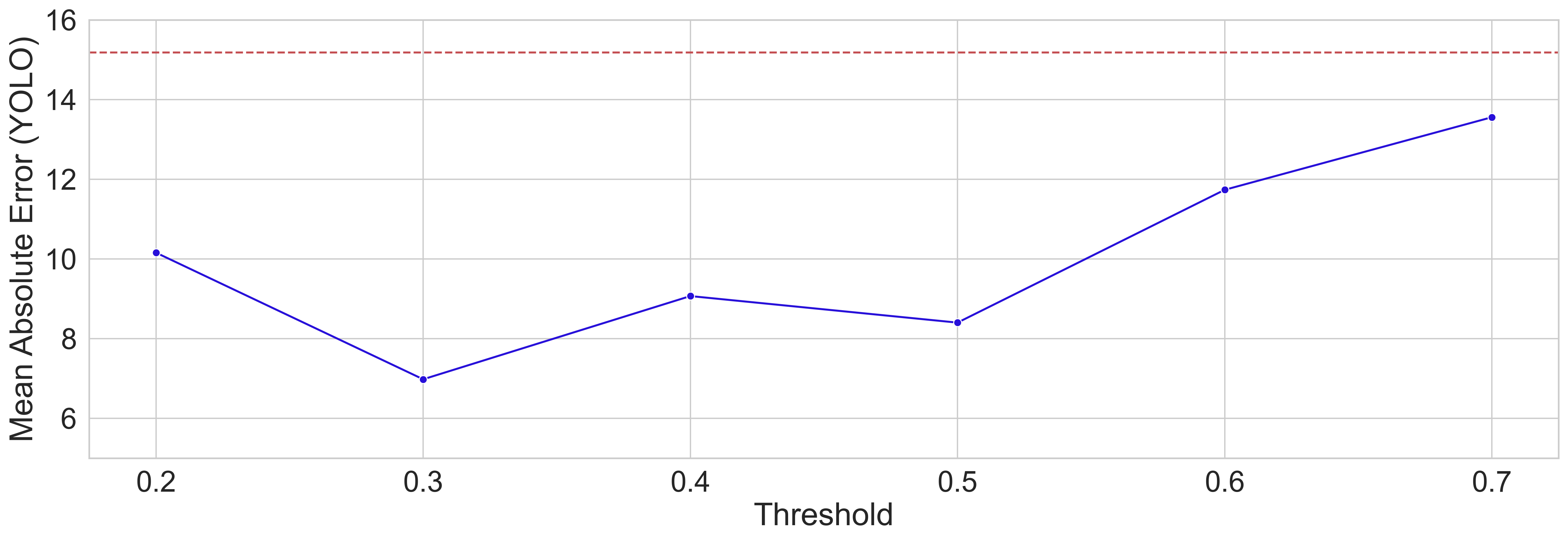}
    \caption{MAE score as a function of YOLO's confidence threshold. The MAE of SD-Turbo is displayed in red as a baseline.}
    \label{fig:yolo_threshold_ablation}
\end{figure*}
\begin{figure*}[t!]
    \centering
    \includegraphics[width=0.7\textwidth]{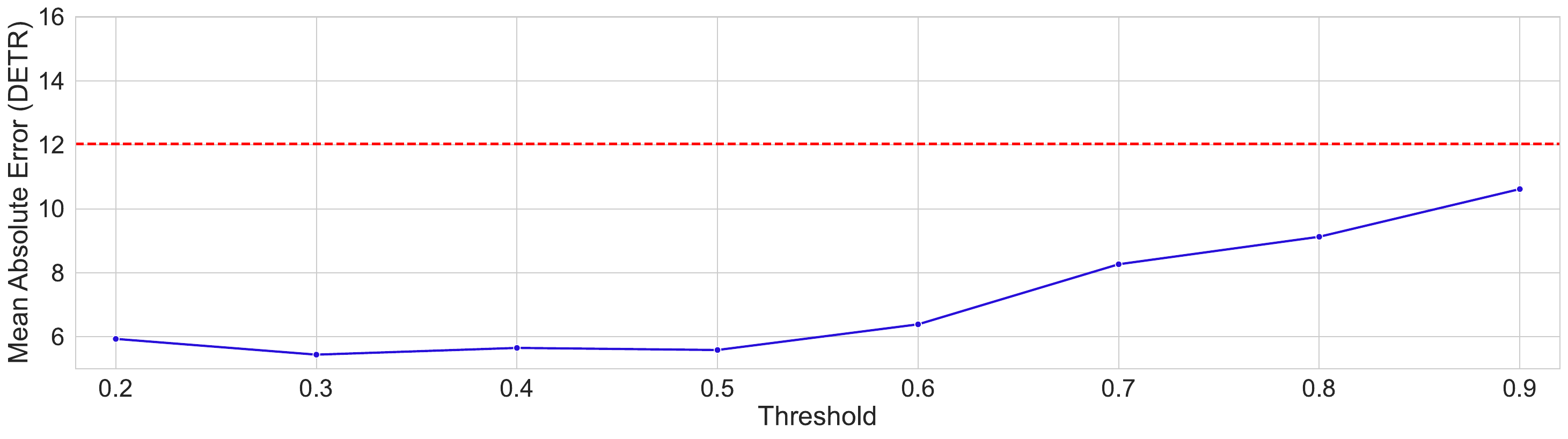}
    \caption{MAE score as a function of DETR's confidence threshold. The MAE of SD-Turbo is displayed in red as a baseline.}
    \label{fig:detr_threshold_ablation}
\end{figure*}

\section{Ablation study: CLIP penalty}
\label{clip_ablation}

Next, in Fig.~\ref{fig:lambda_ablation}, we assess the influence of the CLIP penalty term. We find that setting it to five improves the MAE score. Beyond five, while the CLIP score improves, there is a trade-off with the MAE.

\section{Ablation study: Learning rate}
\label{learning_rate_ablation}

In Fig.~\ref{fig:lr_ablation}, we experimented with our method using learning rates of 0.00025, 0.00125, 0.01, 0.05, and 0.1. The learning rate of 0.01 yielded the best results, suggesting that our method benefits from a higher learning rate.

\section{Ablation study: detection confidence threshold}
\label{threshold_ablation}

Our dynamic scale factor method employs the detection mechanism to accurately scale the object’s density map. Detectors defines a threshold for object detection. 

In Fig.~\ref{fig:yolo_threshold_ablation} and Fig.~\ref{fig:detr_threshold_ablation}, we ablate this parameter and find that a threshold of 0.3 yields the best results for our detectors.

The detection models struggle to be confident in images with many objects. Since our benchmark also consists of a relatively high number of objects, a lower threshold helps achieve more accurate results. 

We hypothesize that dynamically changing this threshold for different counting requirements (e.g., higher confidence for smaller quantities below five and lower confidence for higher quantities beyond twenty) could lead to further improvements.

\end{document}